\documentclass[letterpaper, 10 pt, conference]{ieeeconf}  

\IEEEoverridecommandlockouts                              

\usepackage{algorithm}
\usepackage{algorithmic}
\usepackage{booktabs}
\usepackage{hyperref}
\hypersetup{hypertex=true,
colorlinks=true,
linkcolor=blue,
anchorcolor=blue,
citecolor=blue}

\usepackage{caption}
\usepackage{graphicx, subfig}
\usepackage{amsmath}
\usepackage{amssymb}

\title{\LARGE \bf
MagicStyle: Portrait Stylization Based on Reference Image
}

\author {
    \normalsize
     Zhaoli Deng\textsuperscript{\rm 1},
     Kaibin Zhou\textsuperscript{\rm 1,2},
     Fanyi Wang\textsuperscript{\rm 1*},
     Zhenpeng Mi\textsuperscript{\rm 1}
     \thanks{* Corresponding author,\textsuperscript{\rm 1}Honor Device Co., Ltd,\textsuperscript{\rm 2}Tongji University}\\
     {dengzhaoli, wangfanyi, mizhenpeng}@honor.com, kb824999404@tongji.edu.cn\\
}

\begin{document}

\bibliographystyle{IEEEtran}

\maketitle
\thispagestyle{empty}
\pagestyle{empty}

\begin{abstract}

The development of diffusion models has significantly advanced the research on image stylization, particularly in the area of stylizing a content image based on a given style image, which has attracted many scholars. The main challenge in this reference image stylization task lies in how to maintain the details of the content image while incorporating the color and texture features of the style image. This challenge becomes even more pronounced when the content image is a portrait which has complex textural details. To address this challenge, we propose a diffusion model-based reference image stylization method specifically for portraits, called MagicStyle. MagicStyle consists of two phases: Content and Style DDIM Inversion (CSDI) and Feature Fusion Forward (FFF). The CSDI phase involves a reverse denoising process, where DDIM Inversion is performed separately on the content image and the style image, storing the self-attention query, key and value features of both images during the inversion process. The FFF phase executes forward denoising, harmoniously integrating the texture and color information from the pre-stored feature queries, keys and values into the diffusion generation process based on our Well-designed Feature Fusion Attention (FFA). We conducted comprehensive comparative and ablation experiments to validate the effectiveness of our proposed MagicStyle and FFA.

\end{abstract}

\section{INTRODUCTION}

With the rapid advancement of deep learning technologies, diffusion models \cite{ho2020denoising,rombach2022high} have emerged as significant tools in the fields of image generation and stylization \cite{hertz2024style,chung2024style,xing2024csgo,gao2024styleshot,zhang2023inversion}. In recent years, the application of diffusion models in image stylization research has garnered widespread attention, particularly in tasks that involve stylizing a content image based on a given style image. This task not only holds substantial theoretical significance but also demonstrates immense potential in practical applications such as artistic creation \cite{wang2024diffusion,yi2021exploring} and advertising design \cite{mansour2023intelligent}.

However, the task of stylizing reference images presents numerous challenges. The primary difficulty lies in how to maintain the details and structural integrity of the content image while incorporating the texture features of the style image. This complexity is significantly heightened when the content image is a portrait. Portrait images \cite{wang2024instantid,li2024photomaker,papantoniou2024arc2face} typically contain rich details and subtle features, and any improper stylization can lead to distortion, adversely affecting the final output.

To address this issue, we propose a novel reference image stylization method based on diffusion models, termed MagicStyle. The design philosophy of MagicStyle revolves around two key phases—Content and Style DDIM \cite{song2020denoising} Inversion (CSDI) and Feature Fusion Forward (FFF) to effectively merge content and style information. In the CSDI phase, we perform DDIM Inversion on both the content and style images through a reverse denoising process, extracting and storing self-attention features in the process. The FFF phase then utilizes our designed Feature Fusion Attention (FFA) \cite{vaswani2017attention} to harmoniously integrate the pre-stored feature information into the diffusion generation process, achieving high-quality stylization results. We validate the effectiveness of MagicStyle and FFA through comprehensive comparative and ablation experiments. The results demonstrate that MagicStyle successfully introduces the texture features of the style image while preserving the details of the content, providing a new solution for the stylization of portrait images. Our contributions are as following:
\begin{itemize}
\item We propose a reference image stylization method called MagicStyle based on diffusion models and DDIM inversion sampling characteristics. 
\item We have carefully designed a Feature Fusion Attention (FFA) mechanism that balances the preservation of character detail features from content image and the injection of styles from the style image.
\end{itemize}

\section{RELATED WORKS}

\subsection{Style Transfer}


Style transfer \cite{gatys2016image,jing2019neural} techniques have gained significant attention in recent years, particularly with the advancements in deep learning and generative models. Adaptive Instance Normalization (AdaIN) \cite{huang2017arbitrary} has become a representative method by effectively separating content and style features from deep representations. This approach has inspired a series of techniques based on statistical mean and variance, driving the development of style transfer research.

In the context of diffusion models, DDIM Inversion \cite{song2020denoising} serves as an efficient image recovery technique that enables the mapping from latent space to data space through an implicit denoising process. DDIM Inversion effectively extracts latent features while preserving image details in both image generation and style transfer tasks. InST \cite{zhang2023inversion} further leverages the advantages of DDIM Inversion by directly learning artistic style from a single painting, avoiding complex textual descriptions, thus achieving efficient and accurate style transfer.

To address the issue of style consistency, StyleAligned \cite{hertz2024style} introduces a technique for establishing style alignment among a series of generated images. By employing minimal "attention sharing" during the diffusion process, this method maintains style consistency across images in text-to-image (T2I) models. StyleAligned allows for the creation of style-consistent images using reference styles through a straightforward inversion operation.

StyleID \cite{chung2024style} presents a novel artistic style transfer method based on pre-trained diffusion models without requiring any optimization. This method manipulates the features of self-attention layers, mimicking the behavior of cross-attention mechanisms, and replaces the keys and values of content images with those of style images during the generation process, effectively decoupling content and style features.

CSGO \cite{xing2024csgo} addresses the issue of data scarcity by proposing a data construction pipeline for generating content-style-stylized image triplets and automatically cleansing stylized data. This pipeline has led to the creation of IMAGStyle, a large-scale style transfer dataset containing 210,000 image triplets. CSGO enables image-driven style transfer, text-driven stylized synthesis, and text editing-driven stylized synthesis, further advancing the field of style transfer.

Despite the effectiveness of these methods, a significant challenge remains in maintaining the details of the content image while incorporating the texture features of the style image. This challenge becomes particularly pronounced when the content image is a portrait, as portraits often contain rich details and subtle features, making any improper stylization potentially detrimental to the final output.

\subsection{T2I Personalization}


In the realm of personalized image generation based on diffusion models \cite{zhang2023adding,ye2023ip,wang2024instantid,li2024photomaker,mou2024t2i,huang2023composer}, preserving the details of the content image is a critical challenge. Several emerging methods have demonstrated promising performance in addressing this issue.

ControlNet \cite{zhang2023adding} enhances the control capability of image generation by locking diffusion models and reusing their deep encoding layers to provide strong conditional control functionality. ControlNet employs a "zero convolution" design, ensuring that harmful noise can not affect the fine-tuning process, thereby achieving higher generation quality.

IP-Adapter \cite{ye2023ip} is an effective and lightweight adapter designed to enable image prompt capabilities for pre-trained text-to-image diffusion models. Its key design feature is a decoupled cross-attention mechanism that separates the cross-attention layers for text and image features. Despite its simplicity, IP-Adapter, with only 22 million parameters, achieves performance comparable to or even better than fully fine-tuned image prompt models. By freezing the pre-trained diffusion model, IP-Adapter can be generalized to other custom models and integrated with existing controllable tools for flexible generation.

InstantID \cite{wang2024instantid} offers a powerful diffusion model-based solution that can handle various styles of image personalization using just a single facial image while ensuring high fidelity. To achieve this, InstantID introduces a novel IdentityNet that imposes strong semantic and weak spatial conditions, integrating facial and landmark images with textual prompts to guide image generation. This method demonstrates exceptional performance and efficiency in applications where identity preservation is crucial.

PhotoMaker \cite{li2024photomaker} advances personalized generation by providing an efficient text-to-image generation method that encodes an arbitrary number of input ID images into a stacked ID embedding to preserve ID information. This unified ID representation not only encapsulates the characteristics of the same input ID comprehensively but also accommodates the features of different IDs for subsequent integration, paving the way for more intriguing and practical applications.

\begin{figure*}[!ht]
\begin{center}
\includegraphics[width=0.92\textwidth]{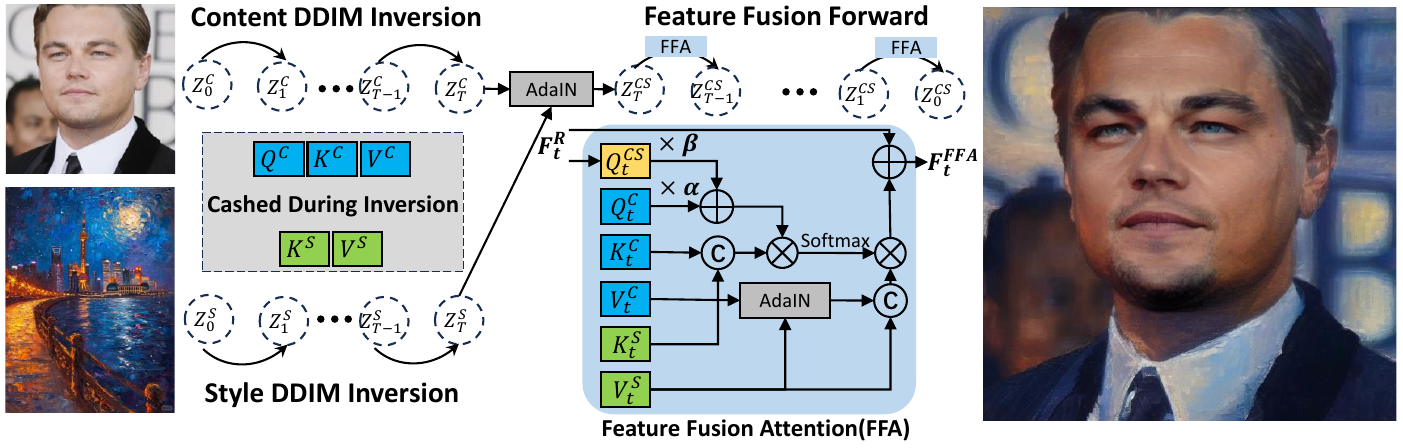}
\end{center}
\caption{Flowchart of MagicStyle. Left side illustrates the Content and Style DDIM Inversion (CSDI) process, and the self-attention features will be restored for the right side Feature Fusion Forward (FFF) process. The fusion operation is mainly conducted during Feature Fusion Attention (FFA).}
\label{fig:flowchart}
\end{figure*}

\section{METHODS}

In this section, we detail our proposed portrait stylization method, MagicStyle, which is based on diffusion models and consists of two main phases: Content and Style DDIM Inversion (CSDI) and Feature Fusion Forward (FFF). The Flowchart of MagicStyle is shown in Fig. \ref{fig:flowchart}.

\subsection{Preliminaries}
Diffusion models~\cite{ho2020denoising} show promising abilities for both image generation. In this work, we opt for a pretrained text-to-image model SD1.5 \cite{rombach2022high} as the base model, which adapts the denoising procedure in the latent space with lower computations. It initially employs VQ-VAE ~\cite{van2017neural} as encoder to transform an image ${x}_{0}$ into the latent space: ${{X}_{0}={\varepsilon}({x}_{0})}$. We can sample $X_t$ at any timestep ${t}$ from ${X}_{0}$ directly using a parameterization trick:
\begin{equation}
\label{FL_DROP}
{X}_t = \sqrt{\bar{\alpha_t}} {X}_{0} + \sqrt{1 - \bar{\alpha_t}}\epsilon, ~~ \epsilon\sim \mathcal{N}(\mathbf{0}, \mathbf{I}),
\end{equation}

\noindent where $\bar{\alpha_t}=\prod_{i=1}^t \alpha_i$, $\alpha_t= 1-\beta_t$. and $\beta_t \in (0,1)$ is a predefined noise schedule. The diffusion model uses a neural network $\epsilon_\theta$ to learn to predict the added noise $\epsilon$ by minimizing the mean square error of the predicted noise which writes:
\begin{equation} 
\label{eq:backward}
\min_\theta \mathbb{E}_{X,\epsilon \sim \mathcal{N}(\mathbf{0},\mathbf{I}),t}[\Vert{\epsilon- \epsilon_\theta(X_t,t,{c_t})}\Vert_2^2],
\end{equation}
\noindent where ${c_t}$ is semantic embedding encoded from text-prompt.

\subsection{Content and Style DDIM Inversion}


The Content and Style DDIM Inversion phase aims to extract features from the content image and style image through a reverse denoising process. Specifically, we perform DDIM \cite{song2020denoising} Inversion on both the content image $I_C$ and the style image $I_S$, resulting in the noisy latent representations $Z^C_T$ and $Z^S_T$ at timestep $T$. 

In the case of SD, $\epsilon_\theta$ is a U-Net architecture in which a block for each resolution comprises a residual block, self-attention \cite{vaswani2017attention} block (SA), and cross-attention
block (CA), sequentially. Following StyleID \cite{chung2024style}, we focus on the SA block to transfer. Specifically, given a feature $ \phi $ input into the SA block, it performs as follows:

\begin{equation} 
\begin{aligned}
Q = W_Q(\phi), K = W_K(\phi), V = W_V(\phi), \\
\text{Attn}(Q, K, V) = \text{softmax}\left(\frac{Q K^T}{\sqrt{d}}\right) \cdot V.
\end{aligned}
\end{equation}

During this process, we store the $\{query, key, value\}$, $\{key, value\}$ of self-attention features of content and style images separately, denoted as $\{Q^C,K^C,V^C\}$ and $\{K^S,V^S\}$ separately, which will be utilized in the subsequent Feature Fusion Forward phase.

\subsection{Feature Fusion Forward}

\begin{algorithm}[t]
\caption{Feature Fusion Attention}
\label{alg:FFA}
\textbf{Input}: Cashed query, key, value of content image 
 $Q^{C}_{t}$, $K^{C}_{t}$, $V^{C}_{t}$, cashed key, value of style image $K^{S}_{t}$,$V^{S}_{t}$, feature maps of resbolck $F^{R}_{t}$, timestep t.\\
\textbf{Parameter}: Fused query of content and style $Q^{CS}_{t}$. $\alpha$, $\beta$ are scaling factors of $Q^{C}_{t}$, $Q^{CS}_{t}$ separately. $Q, K, V$ are query, key, value of Feature Fusion Attention.\\
\textbf{Output}: $F^{FFA}_{t}$.
\begin{algorithmic}[1] 

\FOR{$t = T$ to $1$ }
    \STATE $Q^{CS}_{t} \gets MLP(F^{R}_{t})$
    \STATE $Q \gets \alpha*Q^{C}_{t}+\beta*Q^{CS}_{t}$
    \STATE $K \gets [K^{C}_{t},K^{S}_{t}]$
    \STATE $V \gets [AdaIN(V^{C}_{t},V^{S}_{t}),V^{S}_{t}]$
    \STATE $F^{FFA}_{t} \gets Softmax(Q,K,V)+F^{R}_{t}$
\ENDFOR
\end{algorithmic}
\end{algorithm}


In the Feature Fusion Forward (FFF) phase, the core objective is to harmoniously integrate the texture and color information from the pre-stored feature keys into the diffusion generation process. We employ Adaptive Instance Normalization (AdaIN) \cite{huang2017arbitrary} to fuse the content and style features, which can be expressed as:

\begin{equation}
\begin{aligned}
\label{AdaIN_Z_CS}
Z^{CS}_{T} &= \text{AdaIN}(Z^{C}_{T}, Z^{S}_{T}) \\
&= \sigma(Z^{S}_{T}) \cdot \left( \frac{Z^{C}_{T} - \mu(Z^{C}_{T})}{\sigma(Z^{C}_{T})} \right) + \mu(Z^{S}_{T}),
\end{aligned}
\end{equation}

where $Z^{CS}_{T}$ represents the fused feature representation.
Following this, we perform DDIM sampling on $Z^{CS}_{T}$. During this generation process, we update the query $Q$ to better combine the content and style information, which writes:

\begin{equation}
\begin{aligned}
\label{Q_function}
Q=\alpha *Q^{C}_{t}+\beta*Q^{CS}_{t}, \alpha+\beta=1,
\end{aligned}
\end{equation}
where $\alpha$ and $\beta$ are multiplication factors for the content and style queries, respectively. In the experimental section, we explore these two factors, demonstrating that they can influence the degree of stylization and the retention of content details in the generation results of MagicStyle.


The Feature Fusion Attention (FFA) algorithm is implemented as Algorithm \ref{alg:FFA}. For each timestep $t$ from $T$ to $1$, we update the feature maps of residual block $F^R_t$, using the cashed features $\{Q^C,K^C,V^C\}$ and $\{K^S,V^S\}$. The output feature maps $F^{FFA}_{t}$ then go througth cross-attention.
Through the synergistic operation of these two phases, MagicStyle effectively achieves portrait image stylization by combining the features of content and style images to produce high-quality images.

\section{EXPERIMENTS}

To validate the effectiveness of MagicStyle, we conduct visualization experiments for portrait stylization using different Content and Style images, comparing the results with other baseline models. Furthermore, to verify the effectiveness of Feature Fusion Attention, we performed ablation studies.

\subsection{Implement Details} 
MagicStyle employs RealisticVisionV6 (SD1.5) as the base model. During Content and Style DDIM Inversion stage, we employ 30-step DDIM sampling, with the classifier-free guidance \cite{ho2022classifier} scale set to 1. During Feature Fusion Forward stage, we employ 30-step DDIM sampling, with the classifier-free guidance scale set to 5, $\alpha$ and $\beta$ set to 0.8 and 0.2 separately. To compare the results of style transfer across different content and style images, we collected images from the Internet featuring various genders, ages, and styles for testing. In detail, 48 content images and 36 style images.


\subsection{Visualization Comparison Results}

\begin{figure*}[t]
\centering
\begin{minipage}[b]{0.105\textwidth}
    \includegraphics[width=1\textwidth]{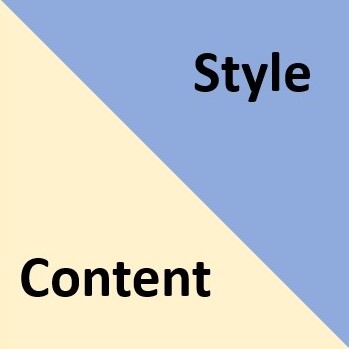}\\
    \includegraphics[width=1\textwidth]{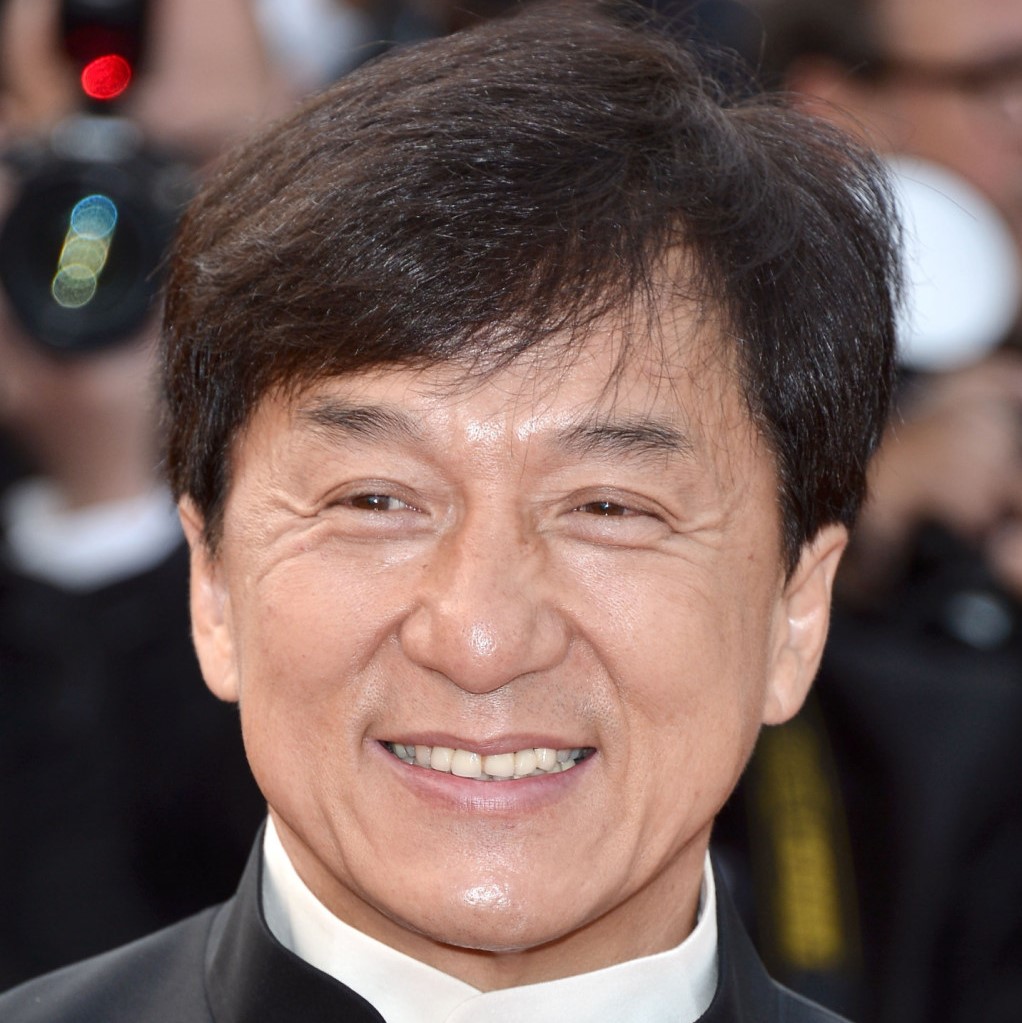}\\
    \includegraphics[width=1\textwidth]{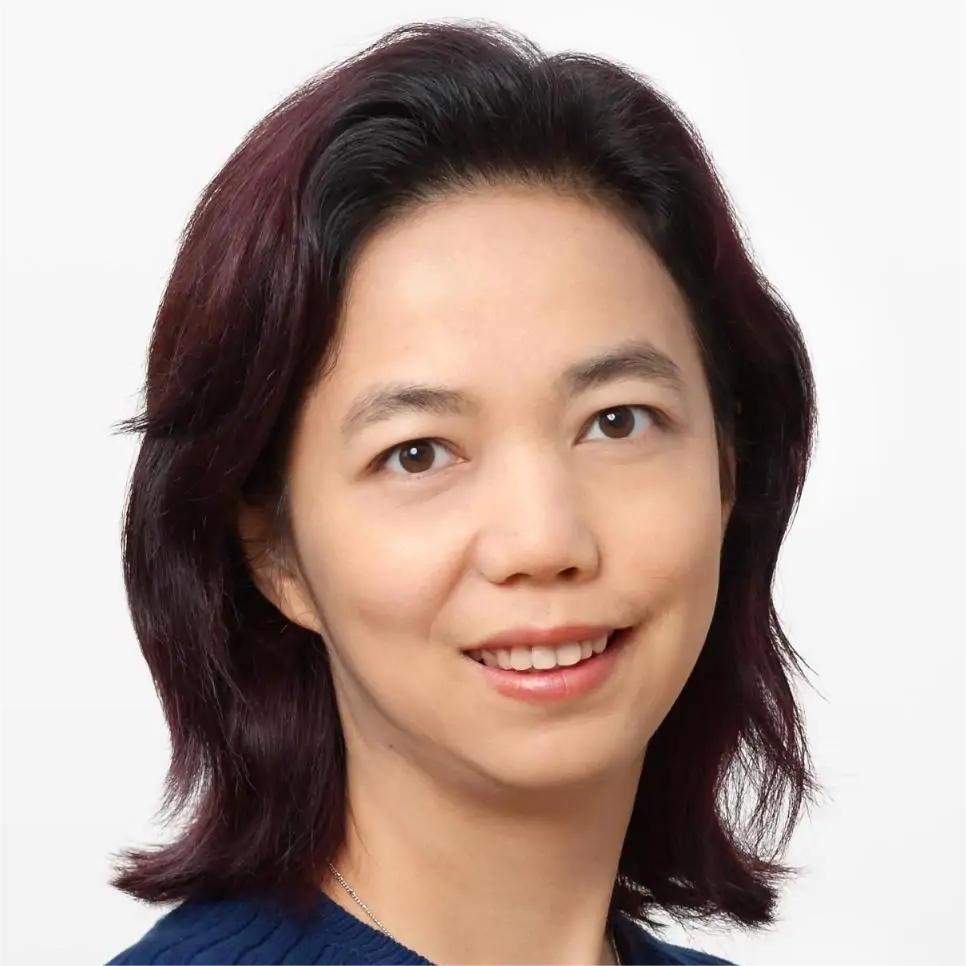}\\
    \includegraphics[width=1\textwidth]{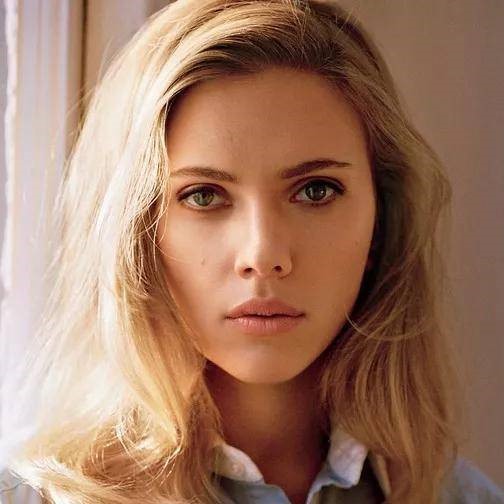}\\
    \includegraphics[width=1\textwidth]{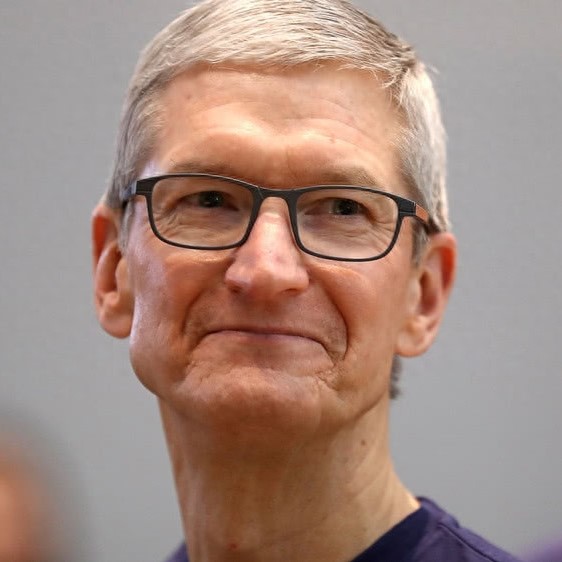}\\
    \includegraphics[width=1\textwidth]{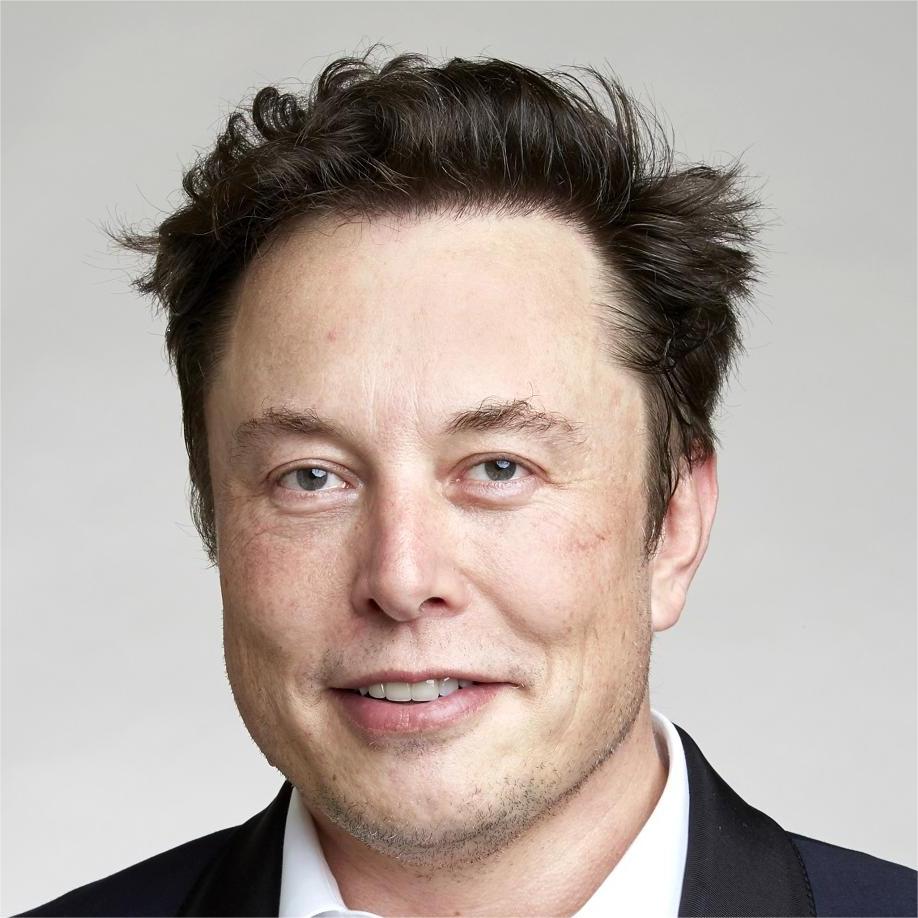}\\
    \includegraphics[width=1\textwidth]{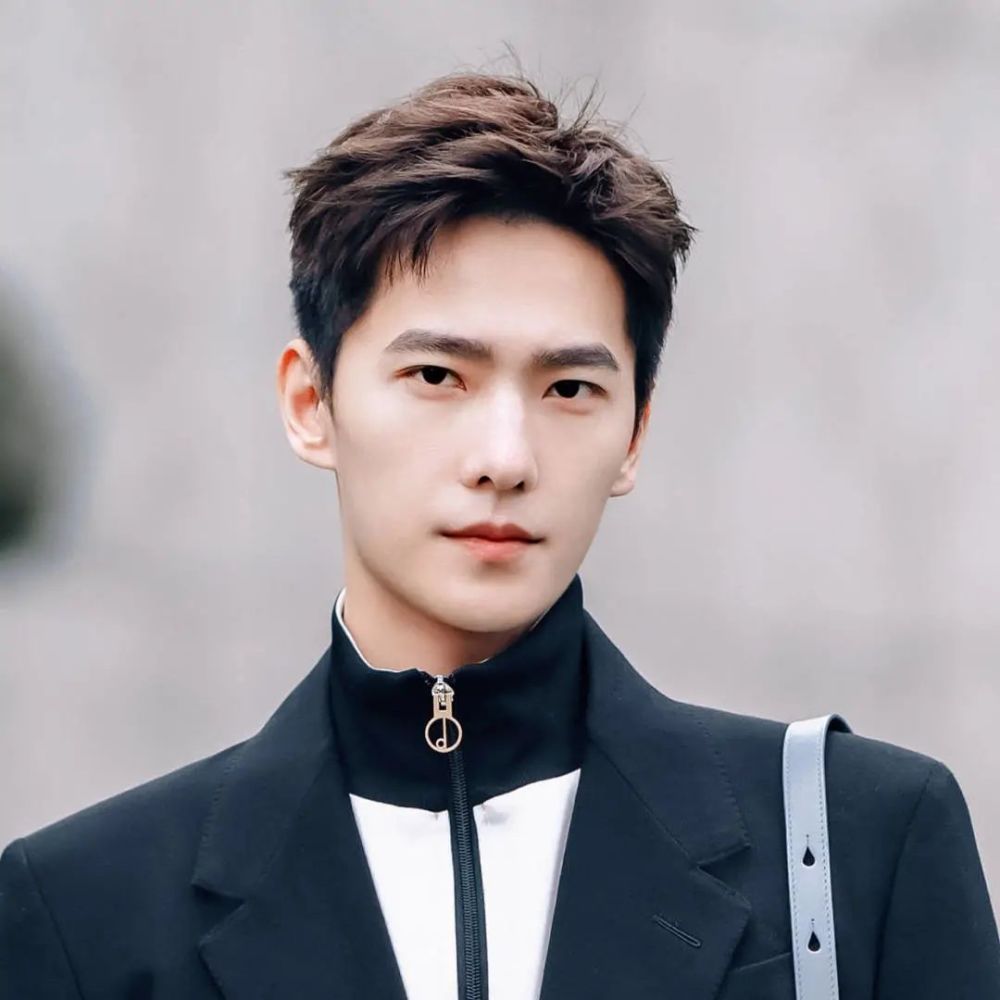}\\
    \includegraphics[width=1\textwidth]{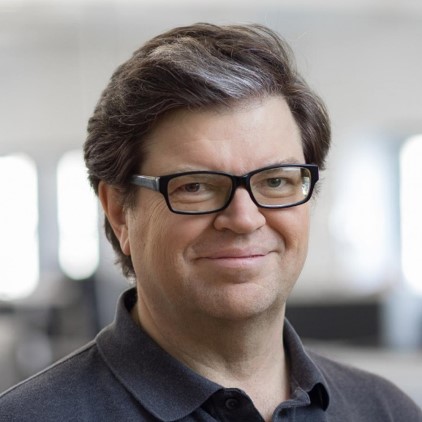}\\
    \includegraphics[width=1\textwidth]{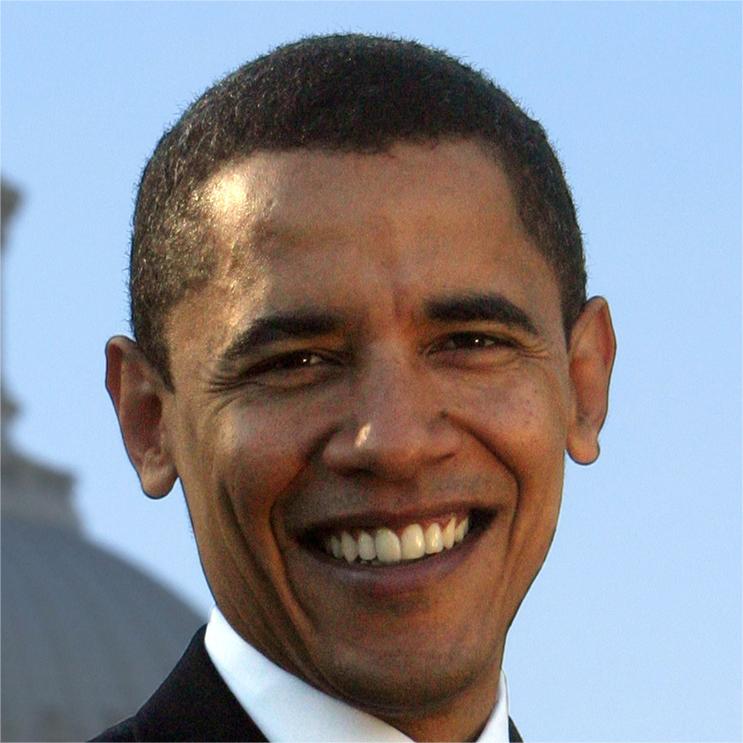}\\
    \includegraphics[width=1\textwidth]{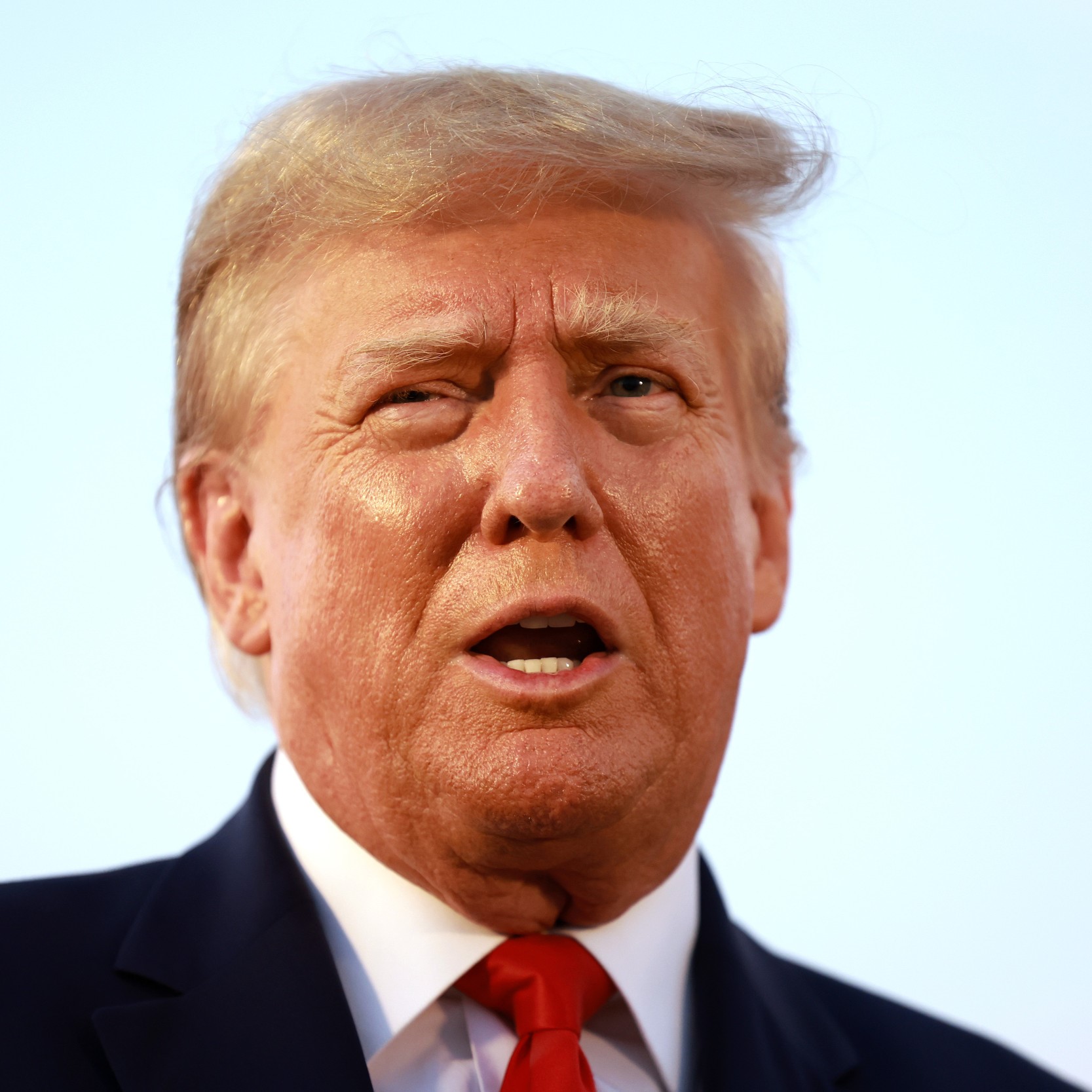}
\end{minipage}
\begin{minipage}[b]{0.105\textwidth}
    \includegraphics[clip, trim=0 105 0 105, width=1\textwidth]{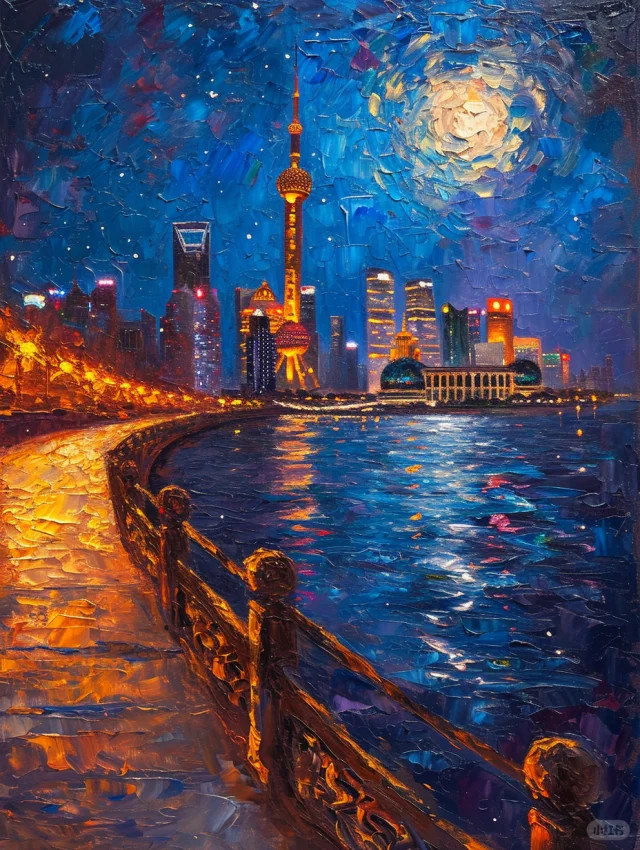}\\
    \includegraphics[width=1\textwidth]{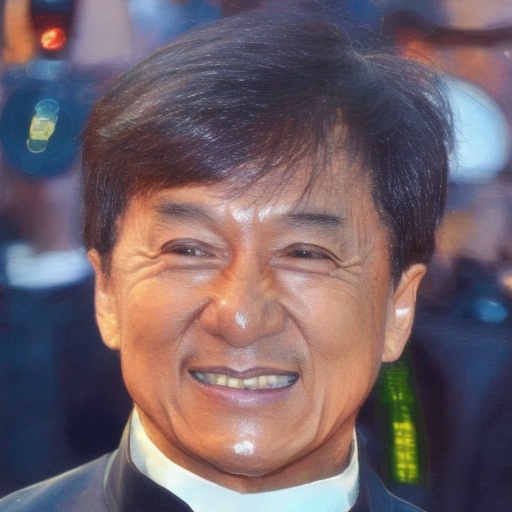}\\
    \includegraphics[width=1\textwidth]{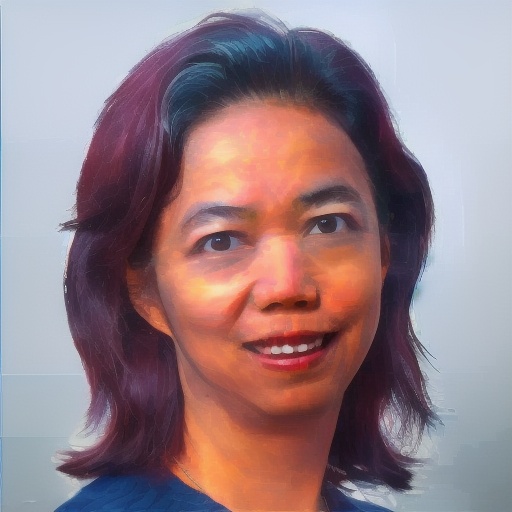}\\
    \includegraphics[width=1\textwidth]{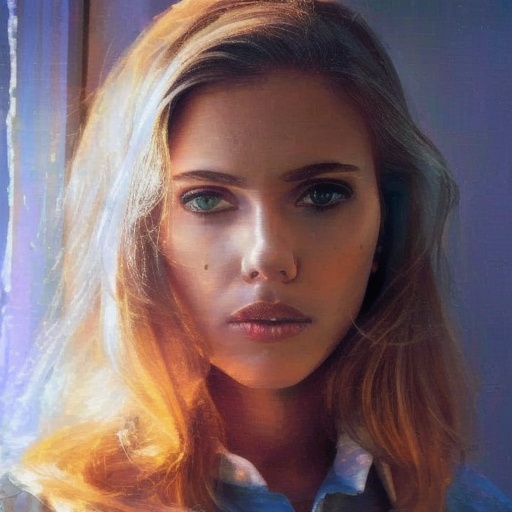}\\
    \includegraphics[width=1\textwidth]{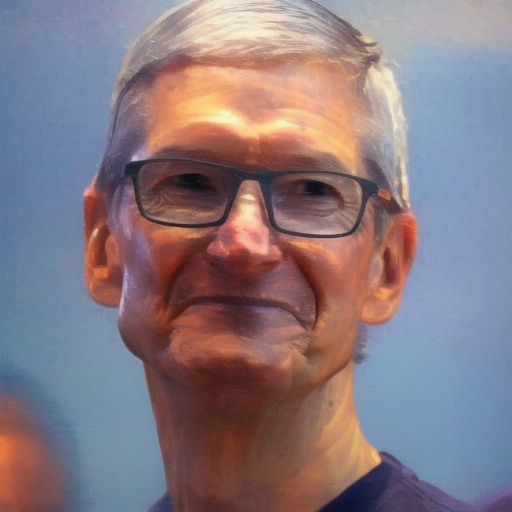}\\
    \includegraphics[width=1\textwidth]{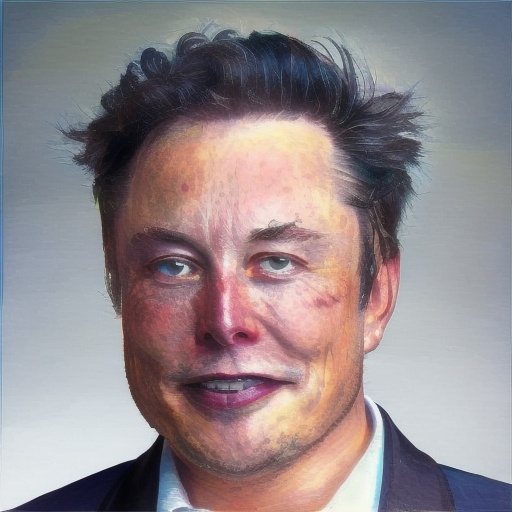}\\
    \includegraphics[width=1\textwidth]{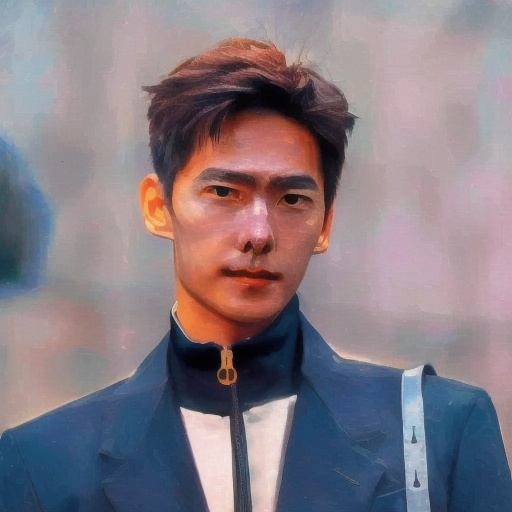}\\
    \includegraphics[width=1\textwidth]{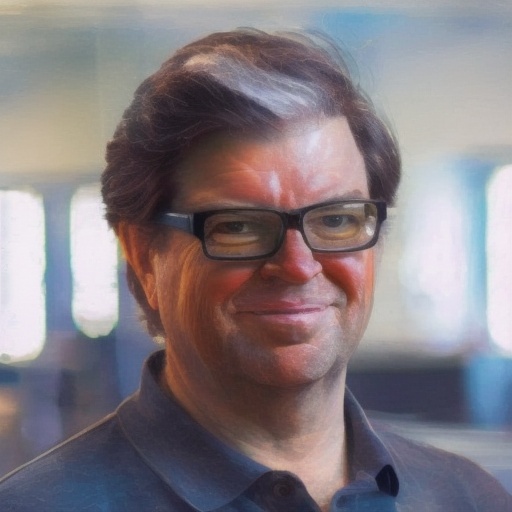}\\
    \includegraphics[width=1\textwidth]{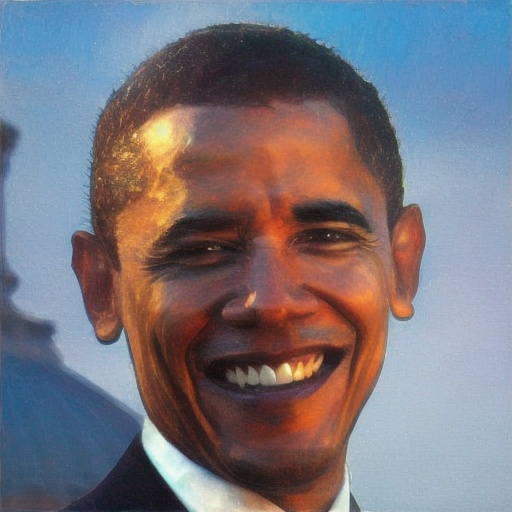}\\
    \includegraphics[width=1\textwidth]{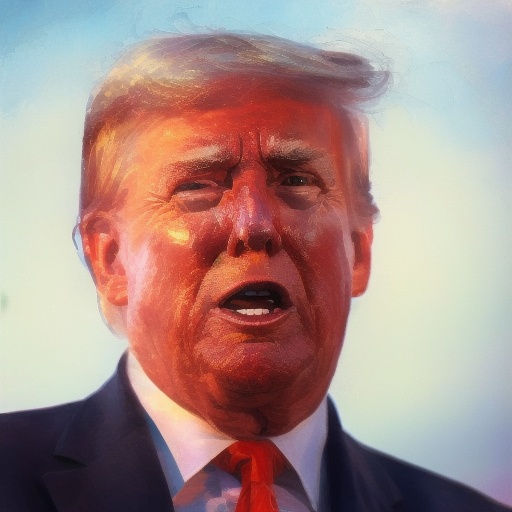}
\end{minipage}
\begin{minipage}[b]{0.105\textwidth}
    \includegraphics[width=1\textwidth]{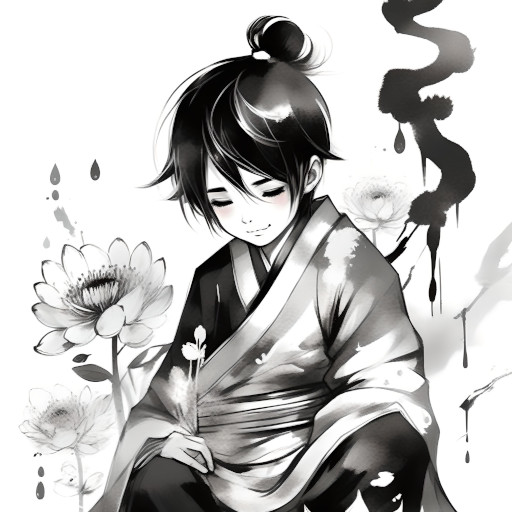}\\
    \includegraphics[width=1\textwidth]{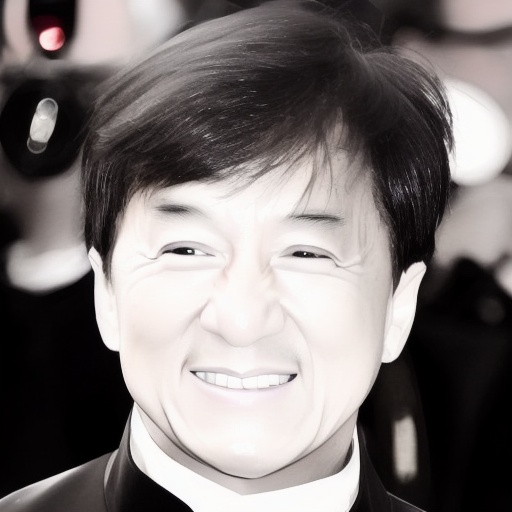}\\
    \includegraphics[width=1\textwidth]{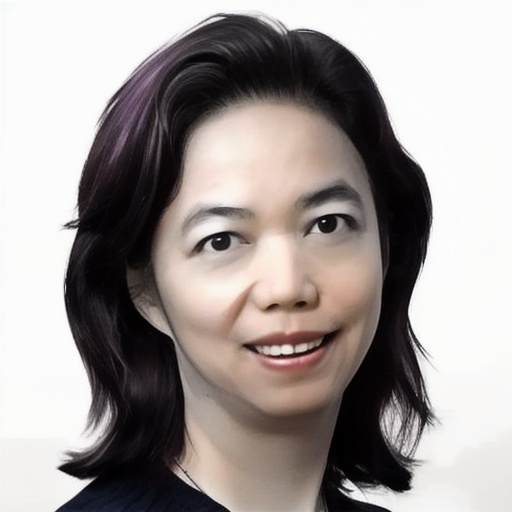}\\
    \includegraphics[width=1\textwidth]{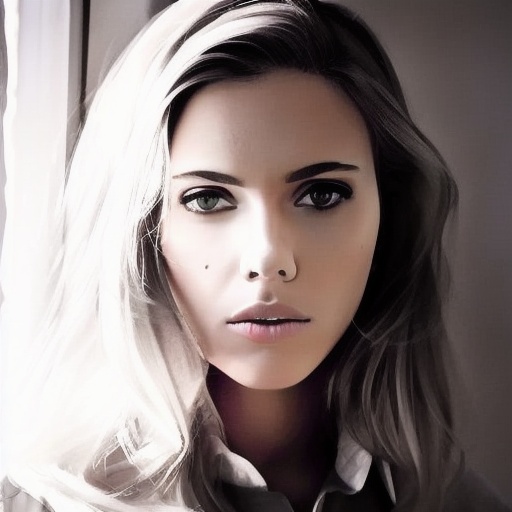}\\
    \includegraphics[width=1\textwidth]{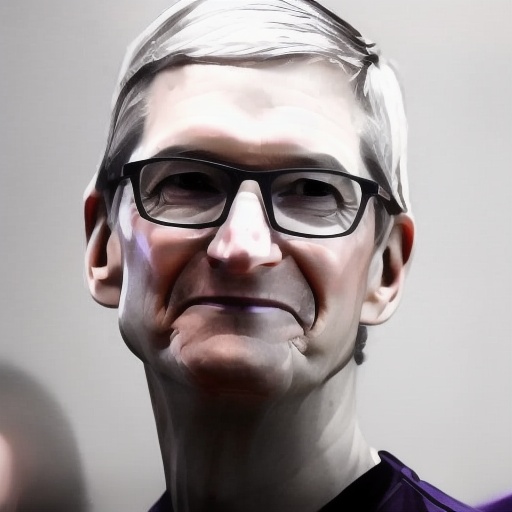}\\
    \includegraphics[width=1\textwidth]{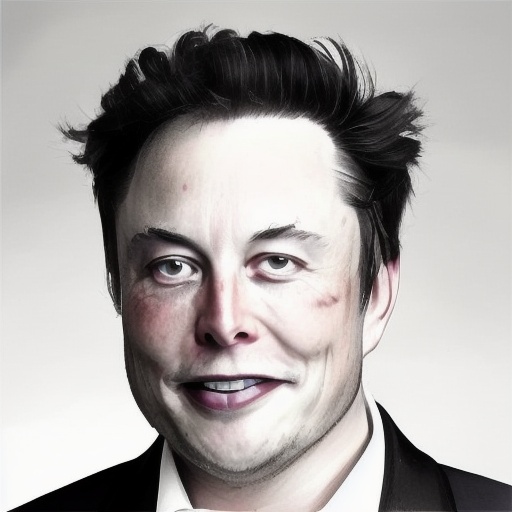}\\
    \includegraphics[width=1\textwidth]{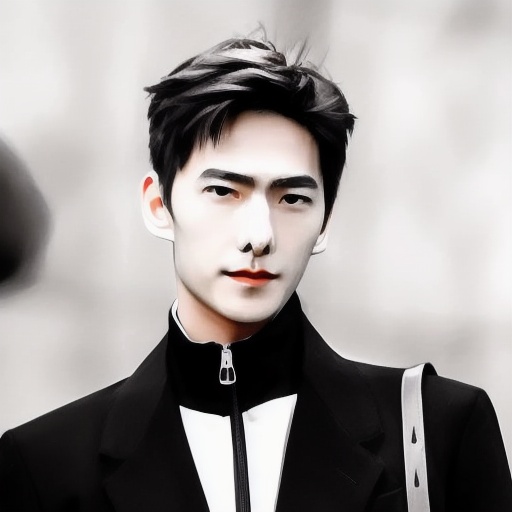}\\
    \includegraphics[width=1\textwidth]{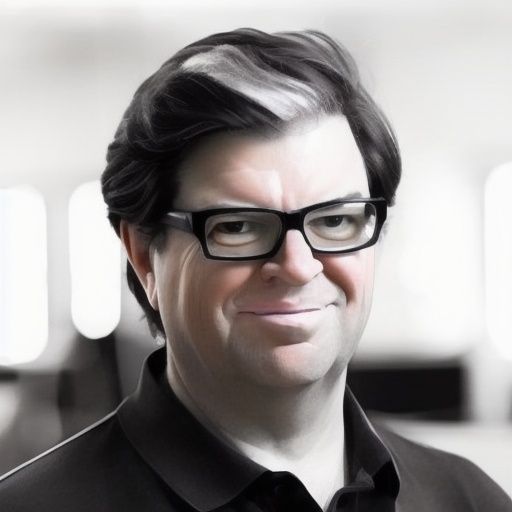}\\
    \includegraphics[width=1\textwidth]{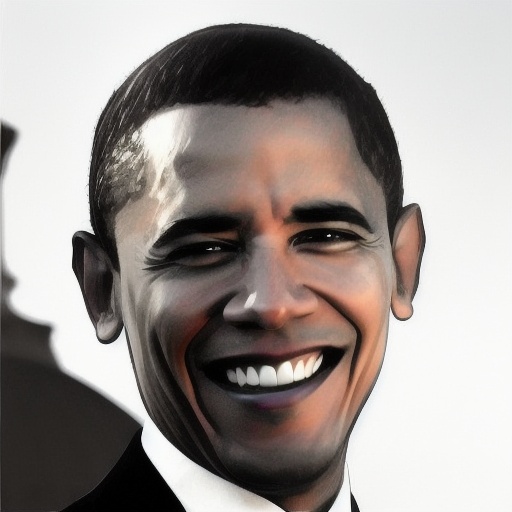}\\
    \includegraphics[width=1\textwidth]{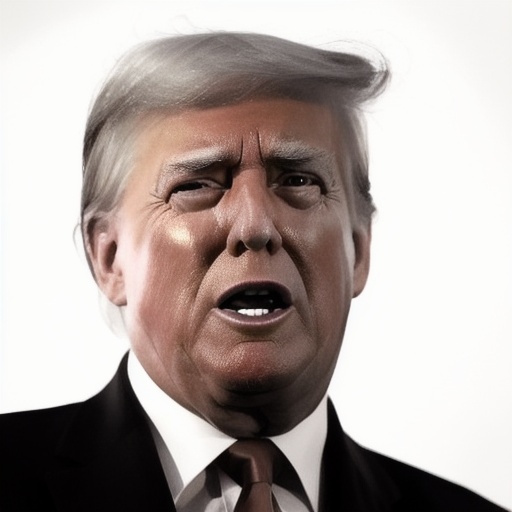}
\end{minipage}
\begin{minipage}[b]{0.105\textwidth}
    \includegraphics[width=1\textwidth]{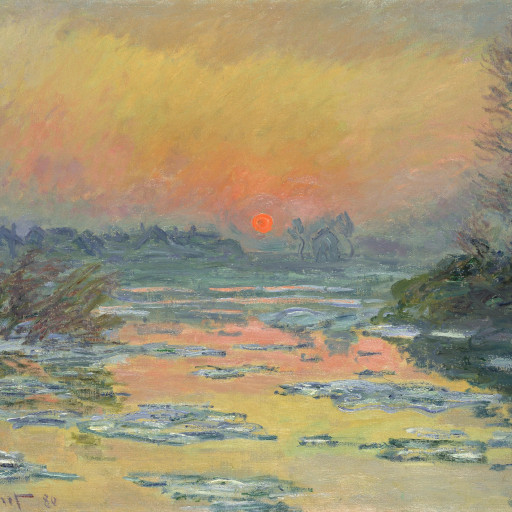}\\
    \includegraphics[width=1\textwidth]{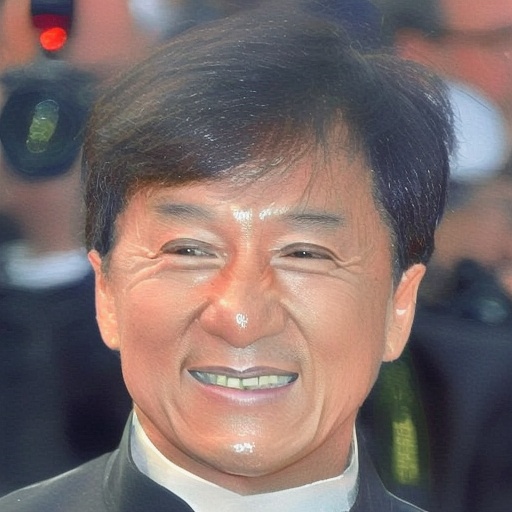}\\
    \includegraphics[width=1\textwidth]{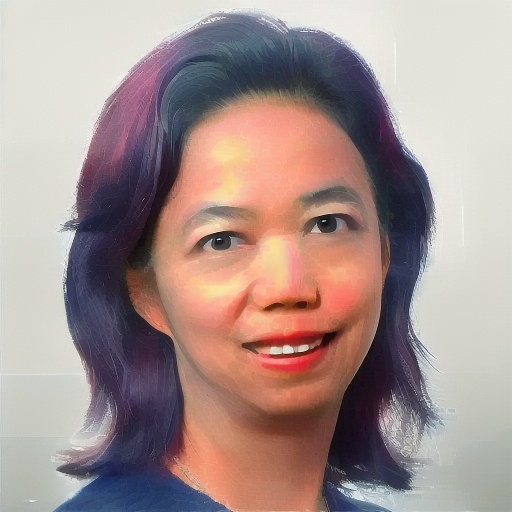}\\
    \includegraphics[width=1\textwidth]{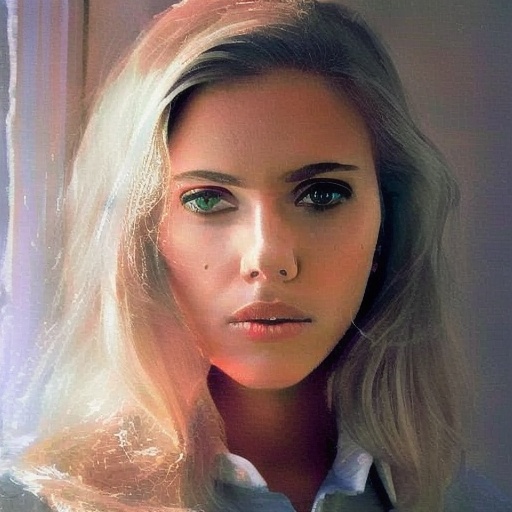}\\
    \includegraphics[width=1\textwidth]{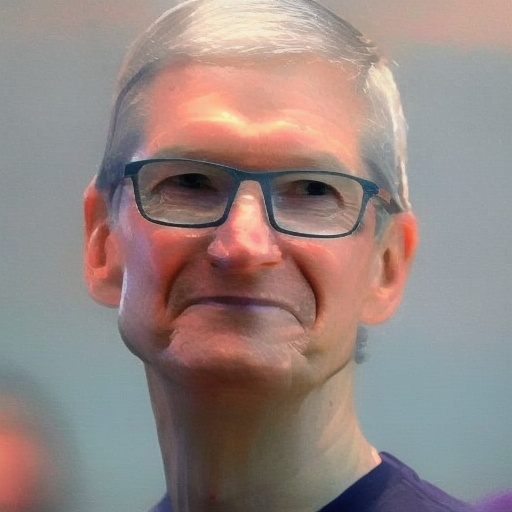}\\
    \includegraphics[width=1\textwidth]{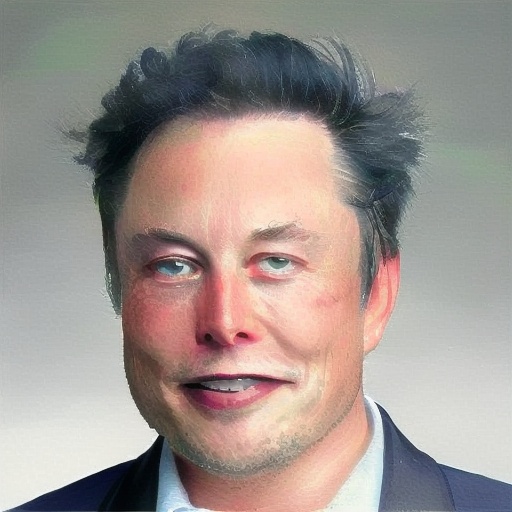}\\
    \includegraphics[width=1\textwidth]{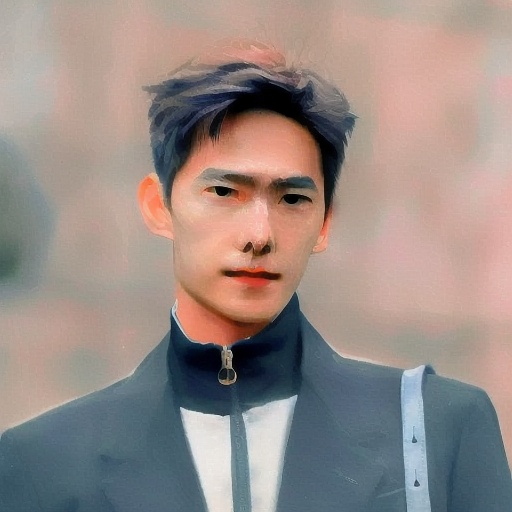}\\
    \includegraphics[width=1\textwidth]{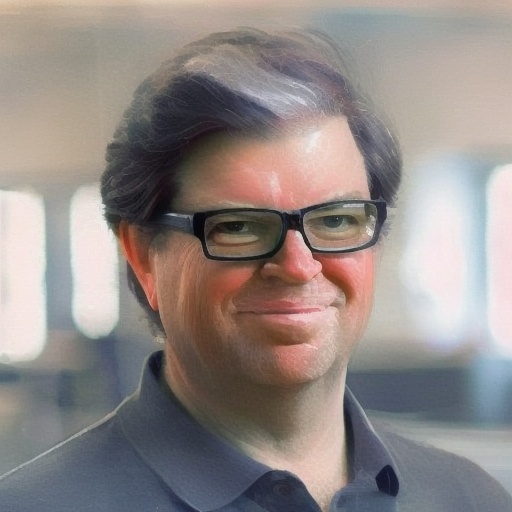}\\
    \includegraphics[width=1\textwidth]{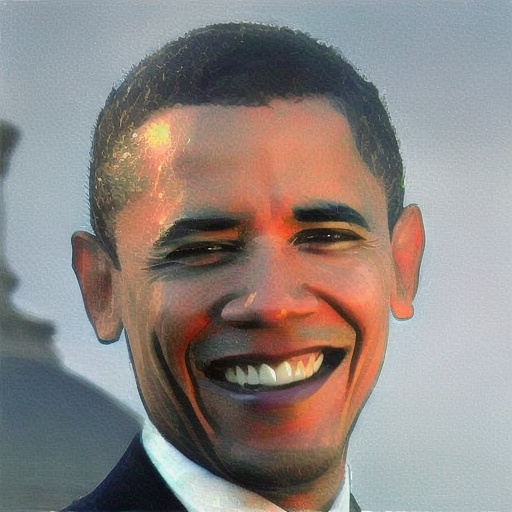}\\
    \includegraphics[width=1\textwidth]{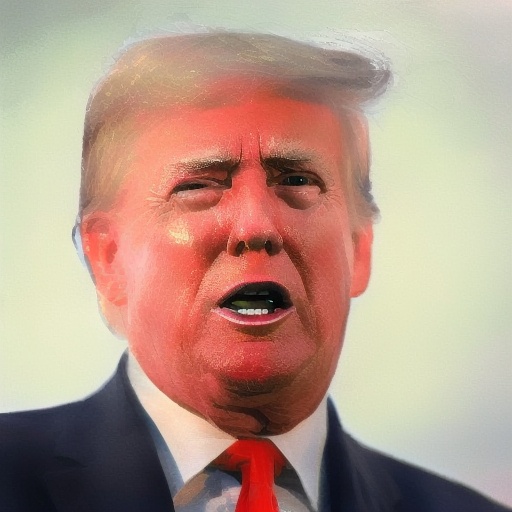}
\end{minipage}
\begin{minipage}[b]{0.105\textwidth}
    \includegraphics[clip, trim=100 0 100 0,width=1\textwidth]{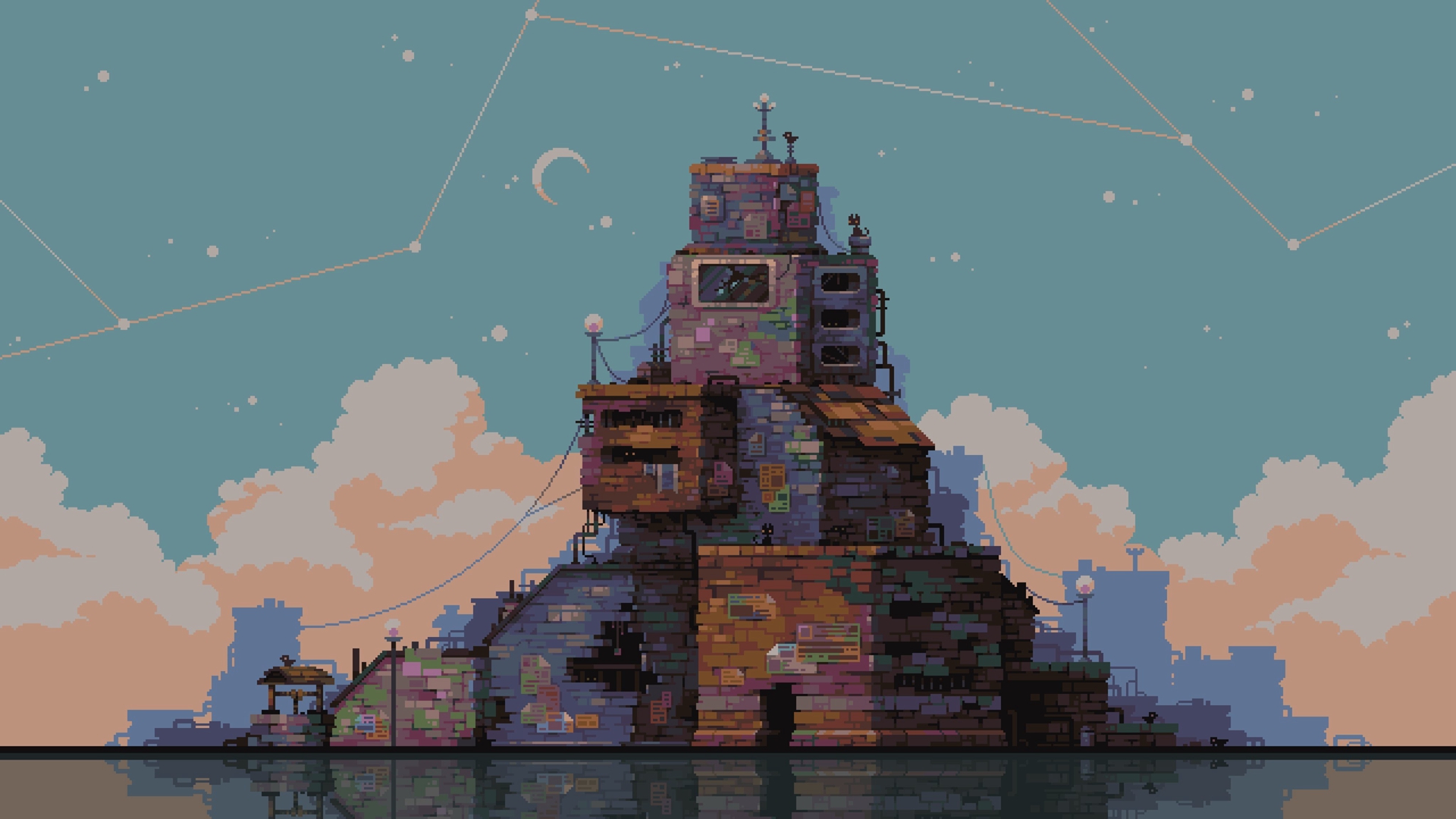}\\
    \includegraphics[width=1\textwidth]{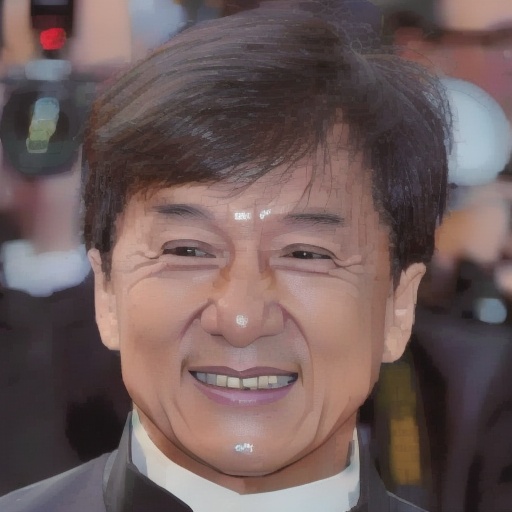}\\
    \includegraphics[width=1\textwidth]{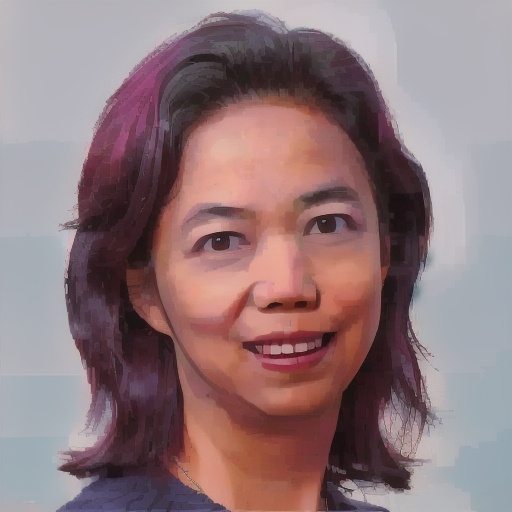}\\
    \includegraphics[width=1\textwidth]{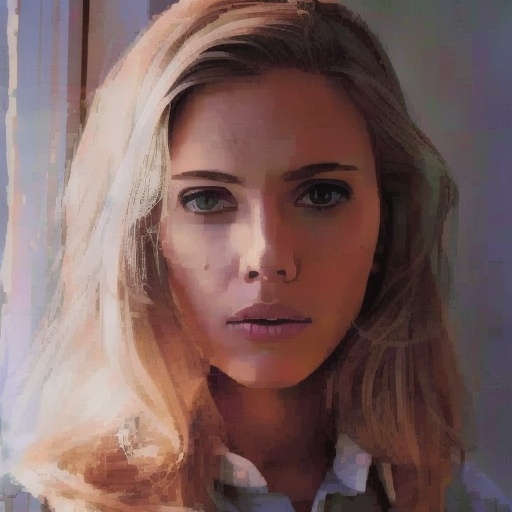}\\
    \includegraphics[width=1\textwidth]{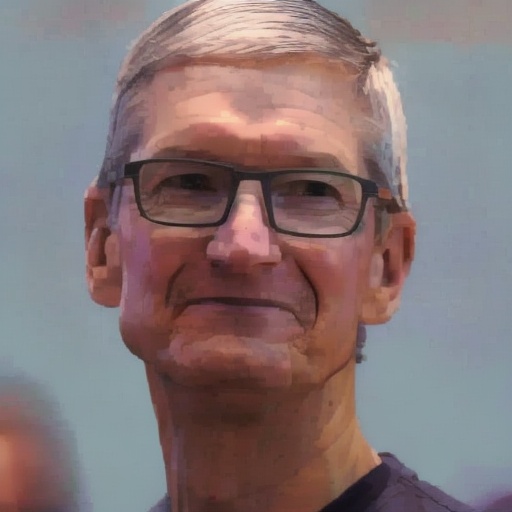}\\
    \includegraphics[width=1\textwidth]{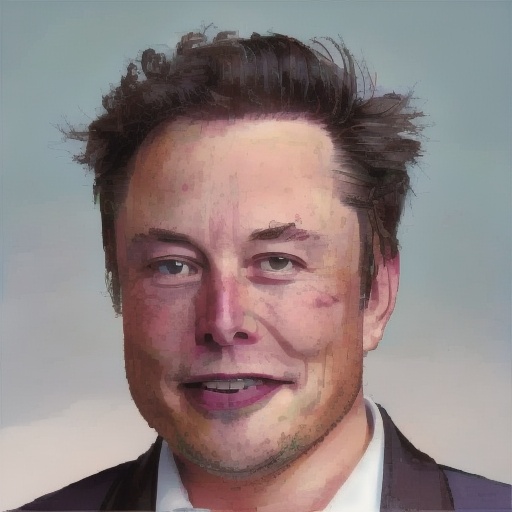}\\
    \includegraphics[width=1\textwidth]{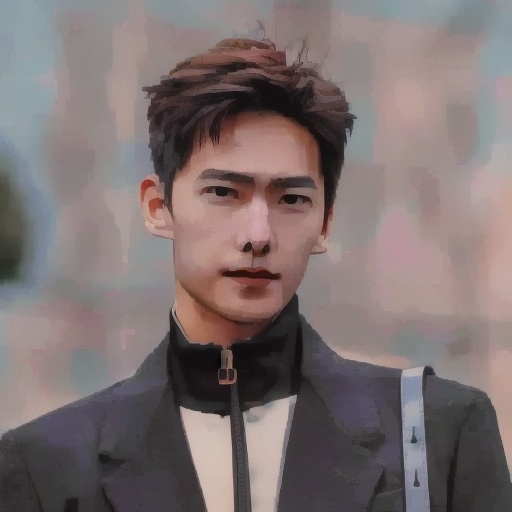}\\
    \includegraphics[width=1\textwidth]{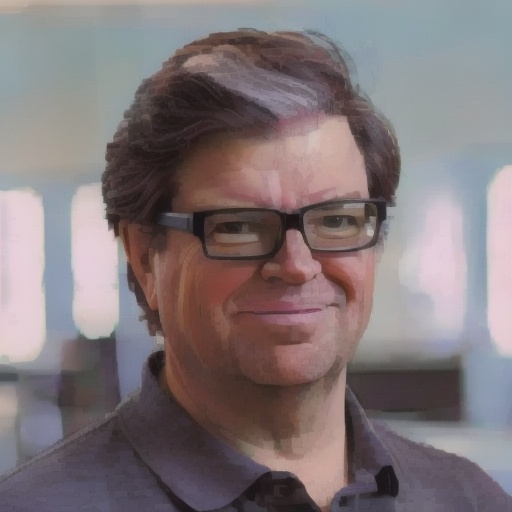}\\
    \includegraphics[width=1\textwidth]{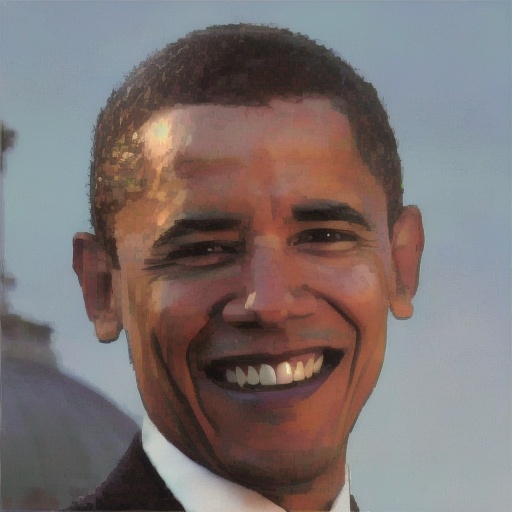}\\
    \includegraphics[width=1\textwidth]{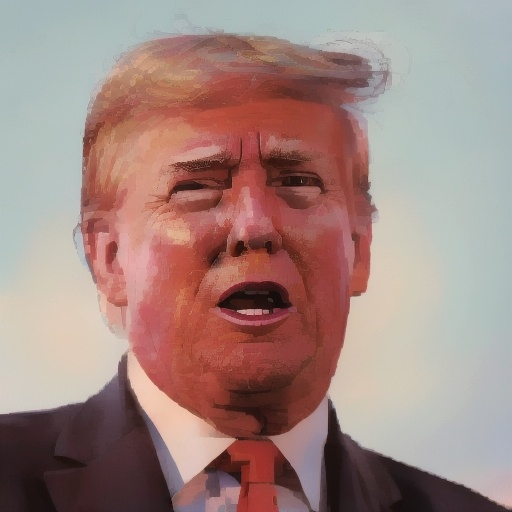}
\end{minipage}
\begin{minipage}[b]{0.105\textwidth}
    \includegraphics[width=1\textwidth]{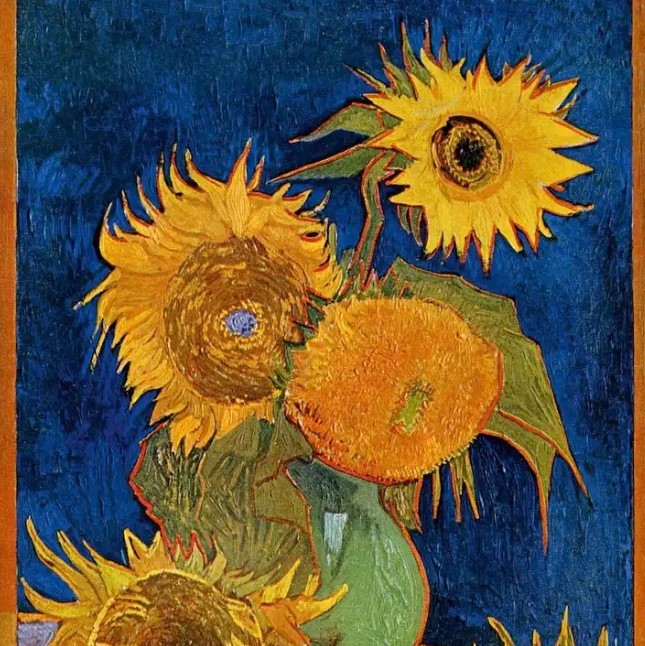}\\
    \includegraphics[width=1\textwidth]{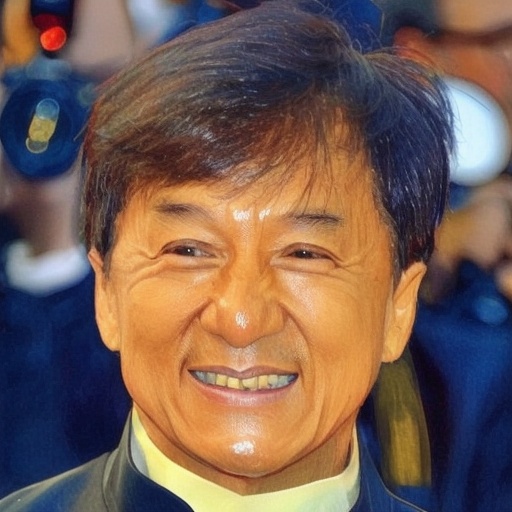}\\
    \includegraphics[width=1\textwidth]{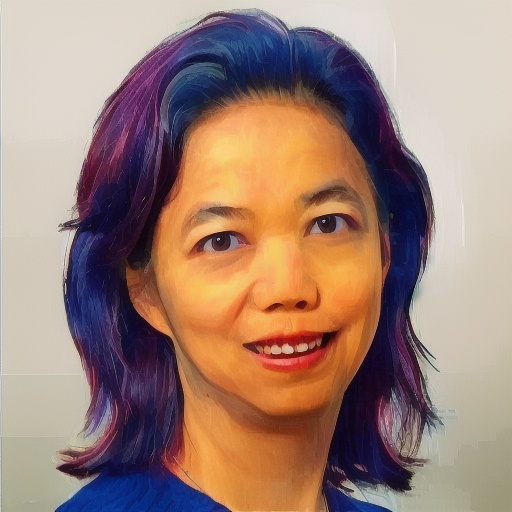}\\
    \includegraphics[width=1\textwidth]{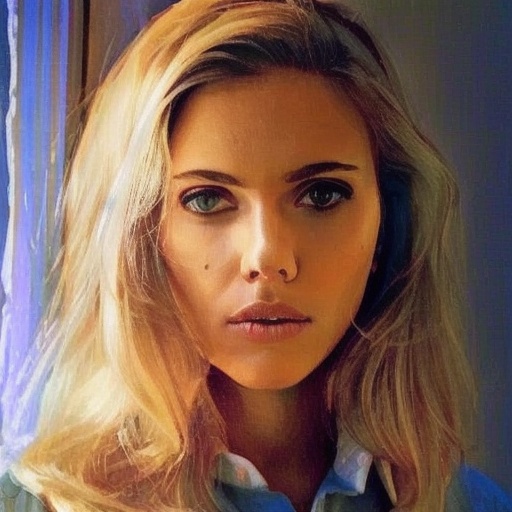}\\
    \includegraphics[width=1\textwidth]{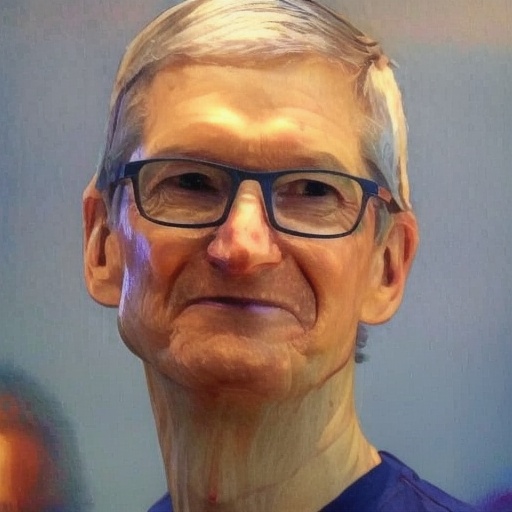}\\
    \includegraphics[width=1\textwidth]{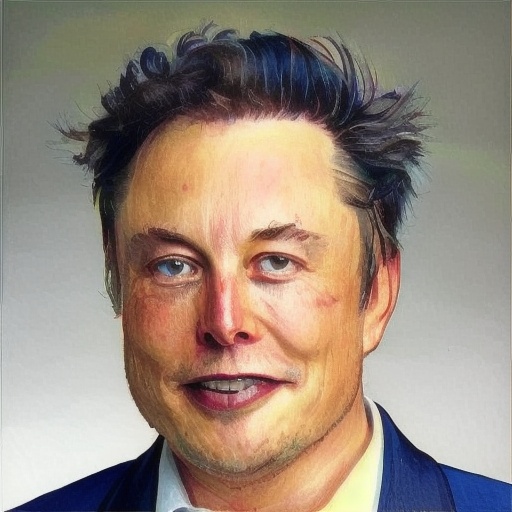}\\
    \includegraphics[width=1\textwidth]{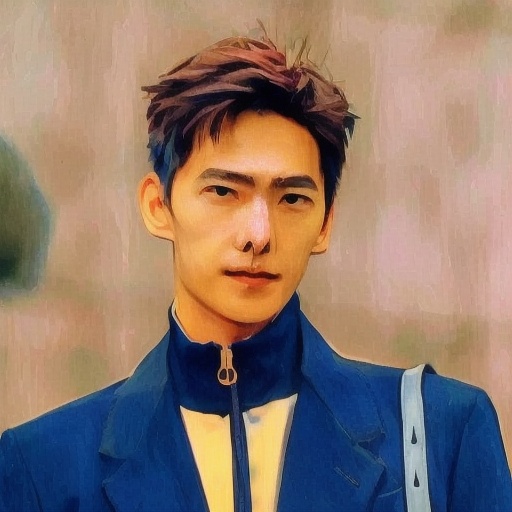}\\
    \includegraphics[width=1\textwidth]{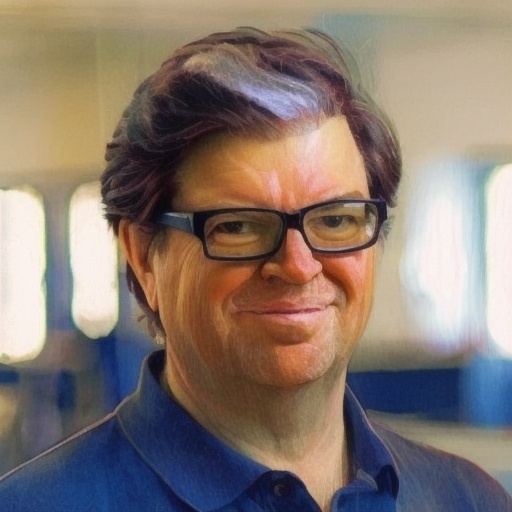}\\
    \includegraphics[width=1\textwidth]{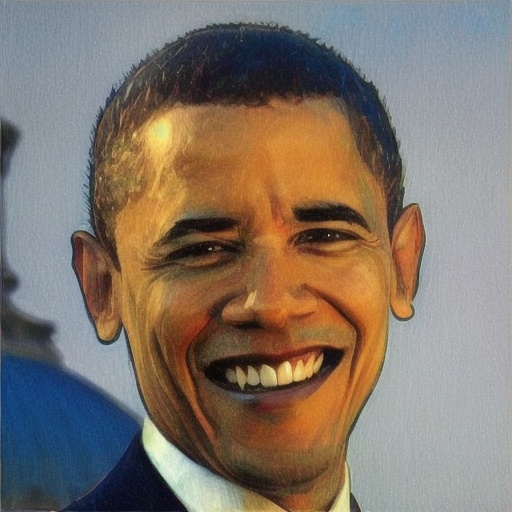}\\
    \includegraphics[width=1\textwidth]{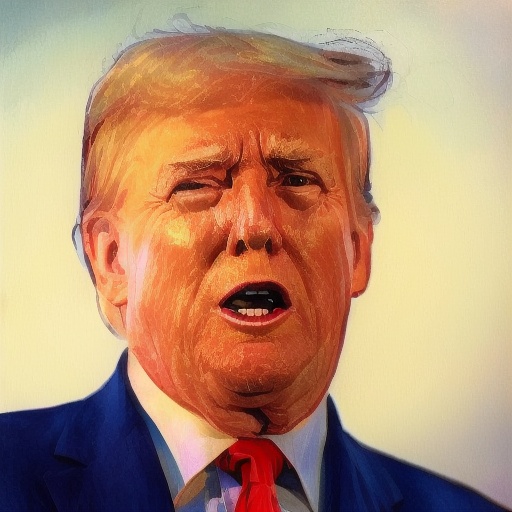}
\end{minipage}
\begin{minipage}[b]{0.105\textwidth}
    \includegraphics[width=1\textwidth]{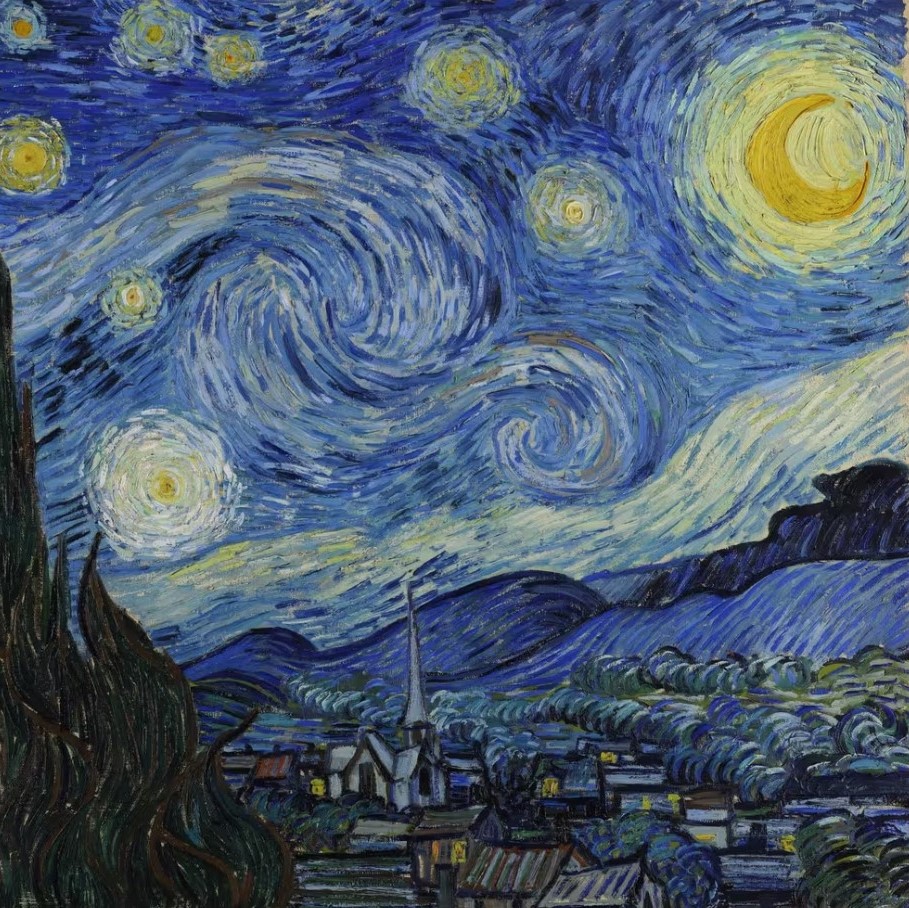}\\
    \includegraphics[width=1\textwidth]{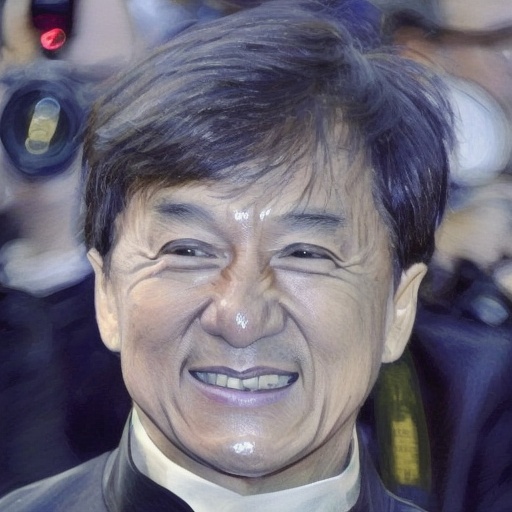}\\
    \includegraphics[width=1\textwidth]{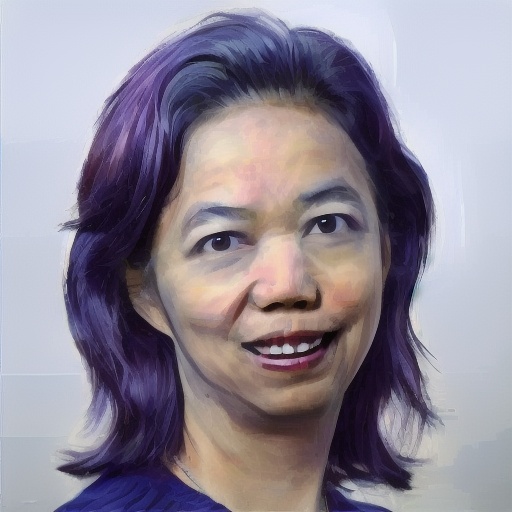}\\
    \includegraphics[width=1\textwidth]{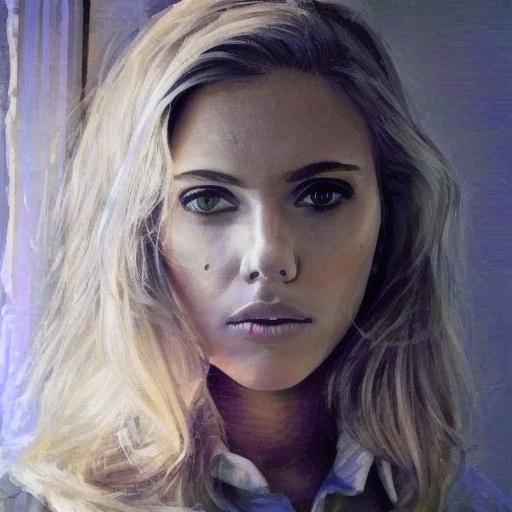}\\
    \includegraphics[width=1\textwidth]{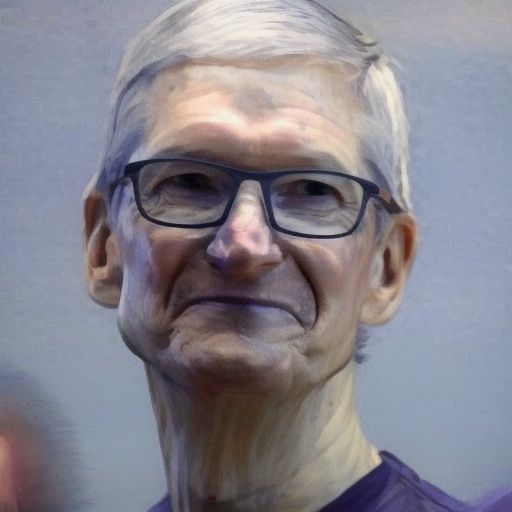}\\
    \includegraphics[width=1\textwidth]{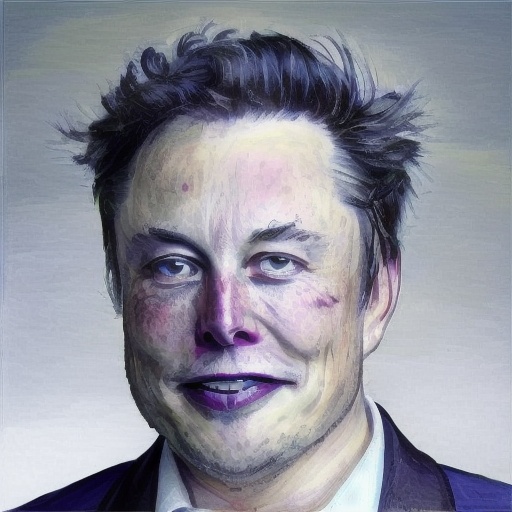}\\
    \includegraphics[width=1\textwidth]{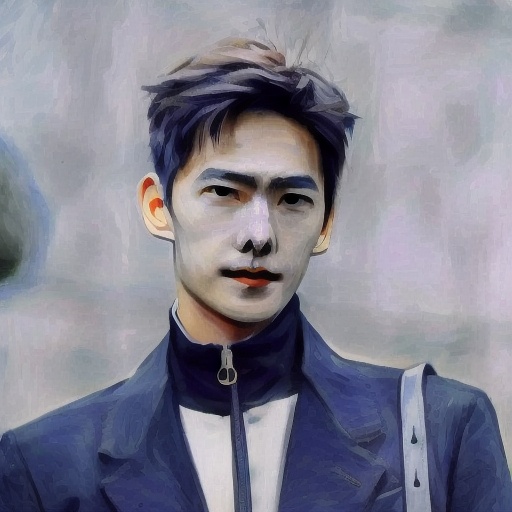}\\
    \includegraphics[width=1\textwidth]{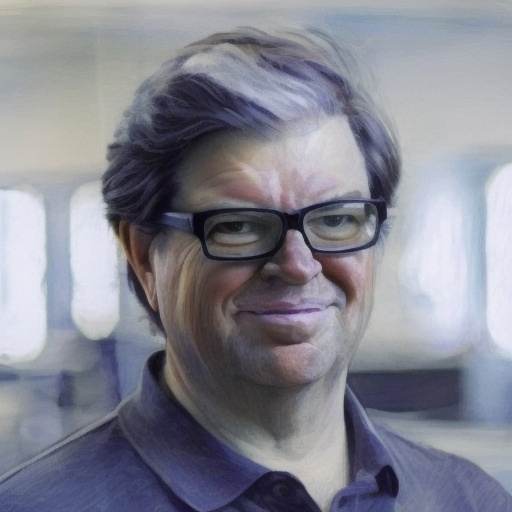}\\
    \includegraphics[width=1\textwidth]{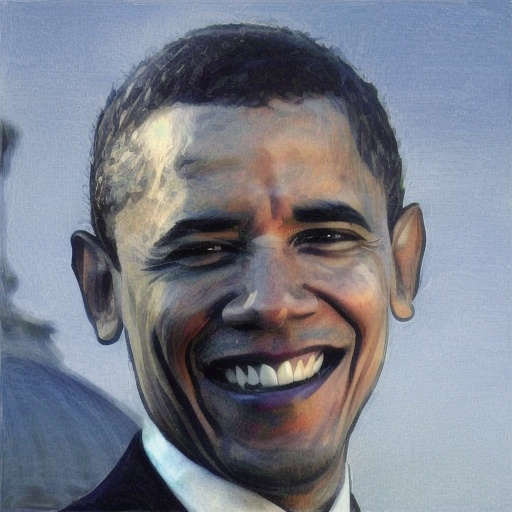}\\
    \includegraphics[width=1\textwidth]{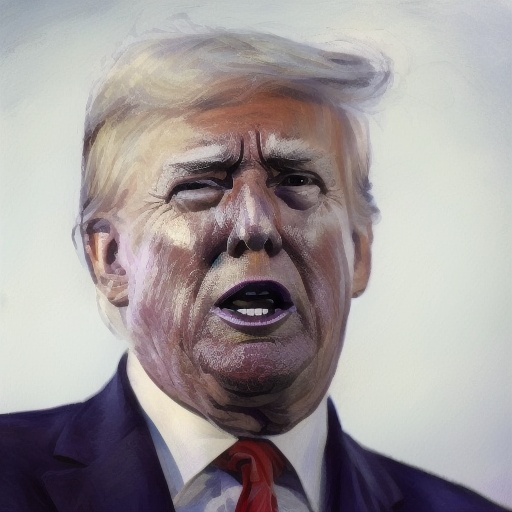}
\end{minipage}
\begin{minipage}[b]{0.105\textwidth}
    \includegraphics[width=1\textwidth]{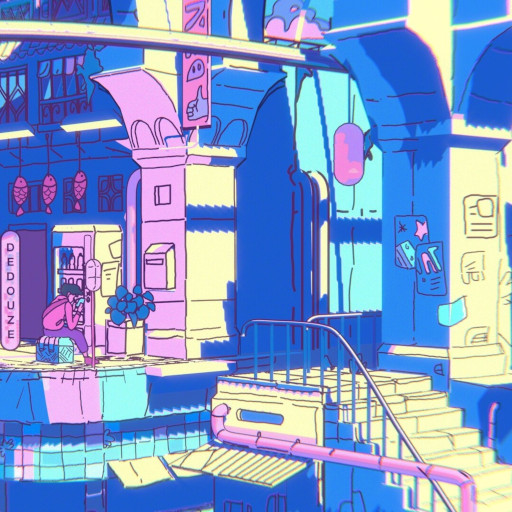}\\
    \includegraphics[width=1\textwidth]{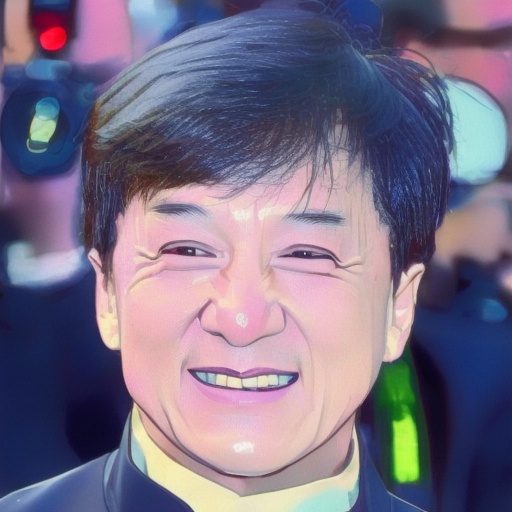}\\
    \includegraphics[width=1\textwidth]{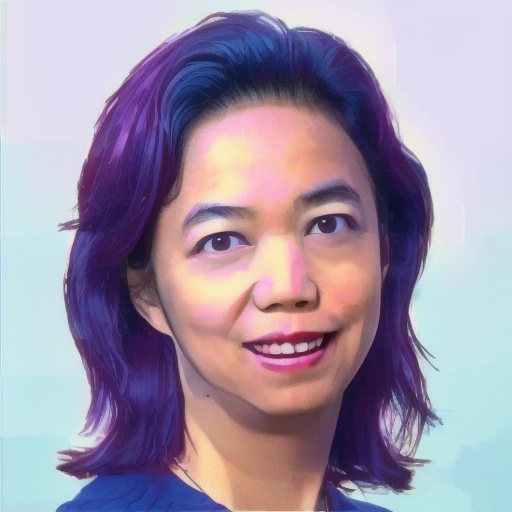}\\
    \includegraphics[width=1\textwidth]{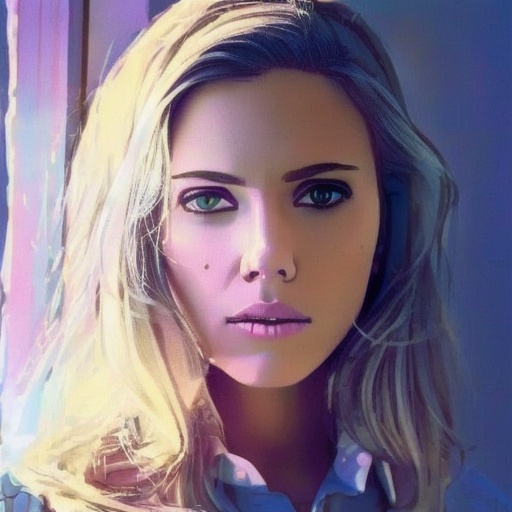}\\
    \includegraphics[width=1\textwidth]{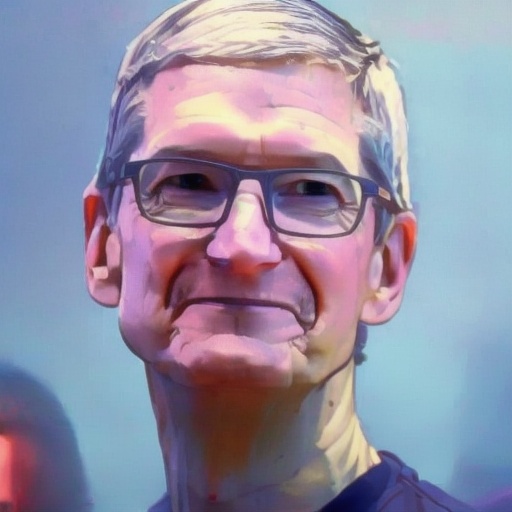}\\
    \includegraphics[width=1\textwidth]{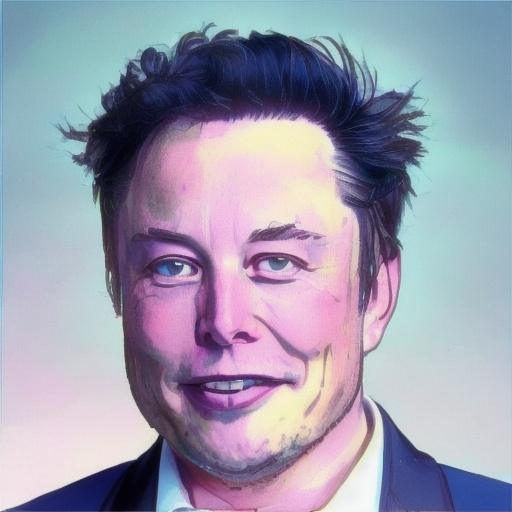}\\
    \includegraphics[width=1\textwidth]{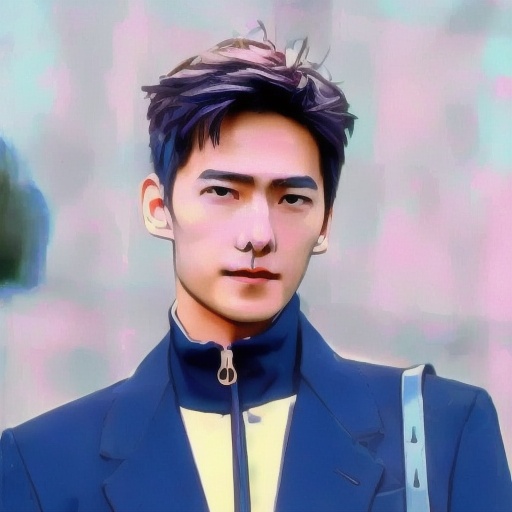}\\
    \includegraphics[width=1\textwidth]{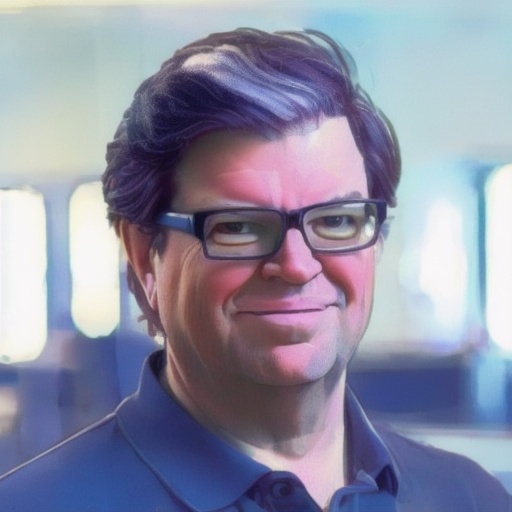}\\
    \includegraphics[width=1\textwidth]{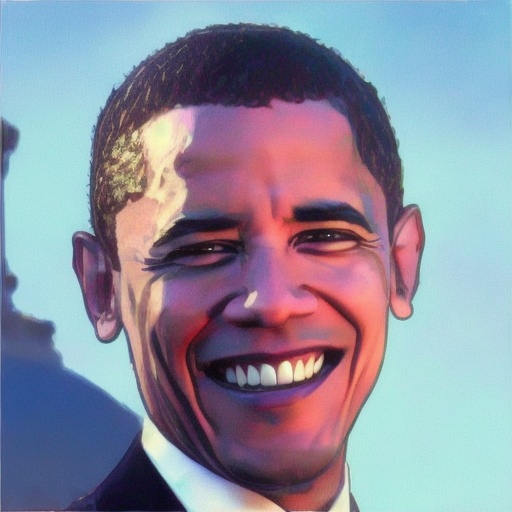}\\
    \includegraphics[width=1\textwidth]{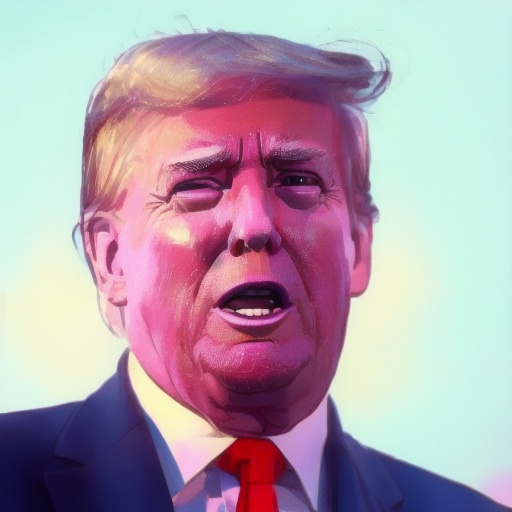}
\end{minipage}
\begin{minipage}[b]{0.105\textwidth}
    \includegraphics[width=1\textwidth]{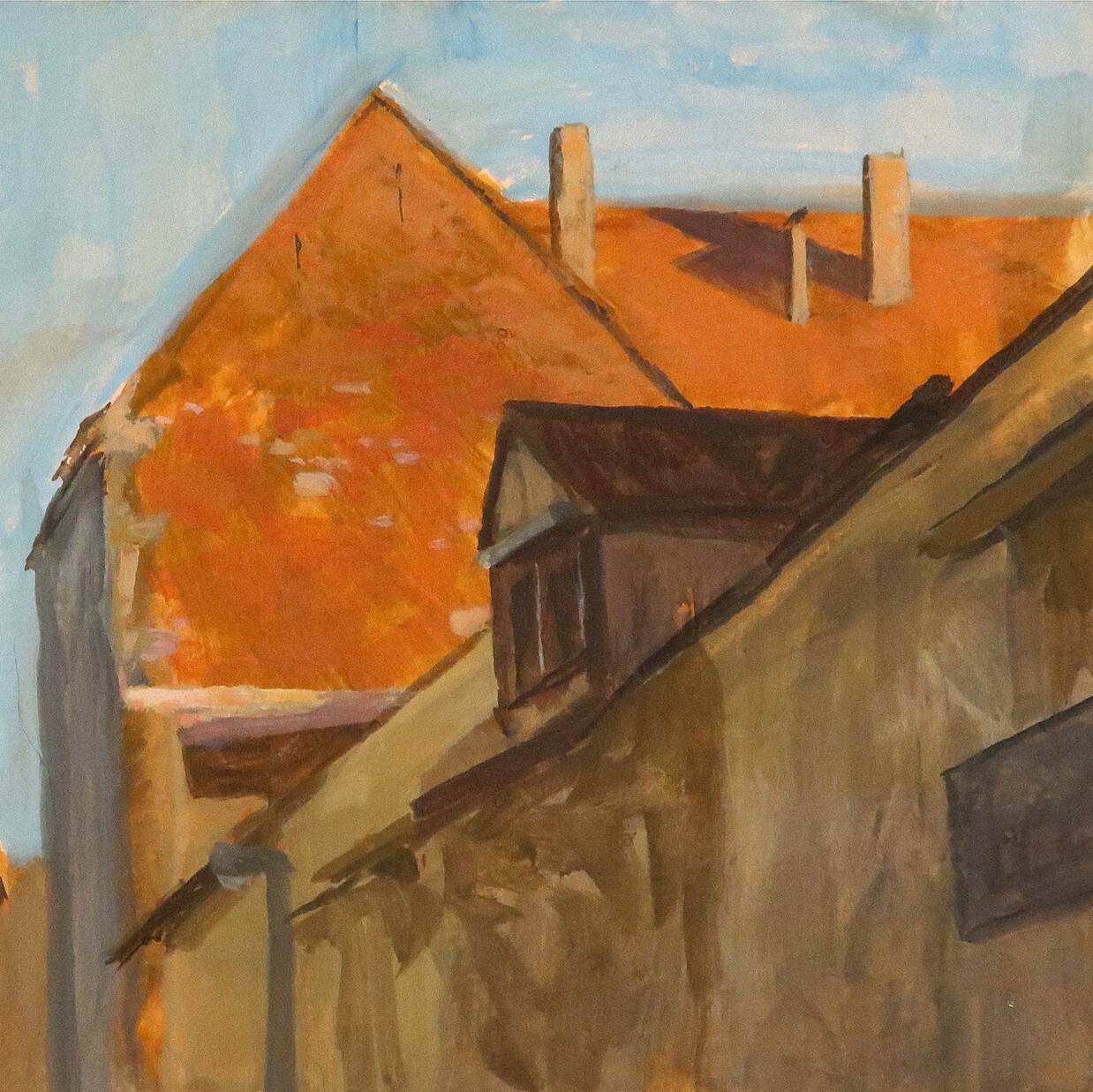}\\
    \includegraphics[width=1\textwidth]{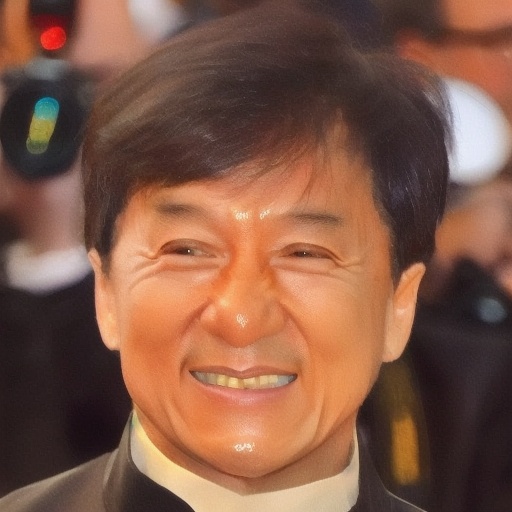}\\
    \includegraphics[width=1\textwidth]{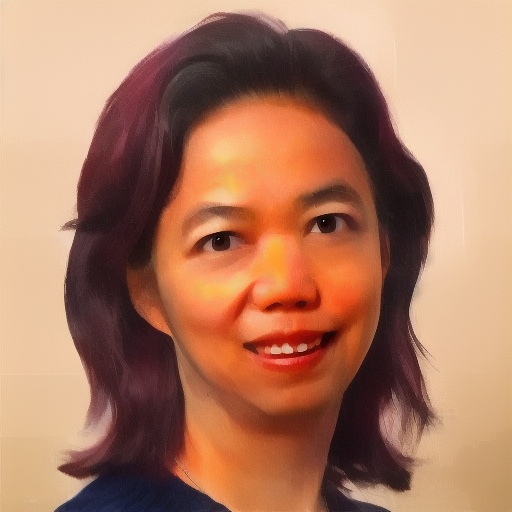}\\
    \includegraphics[width=1\textwidth]{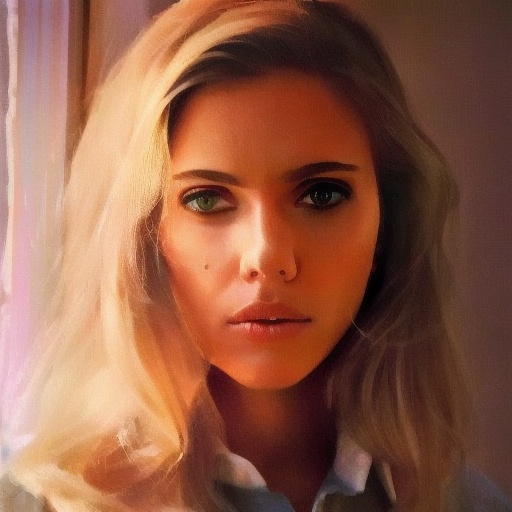}\\
    \includegraphics[width=1\textwidth]{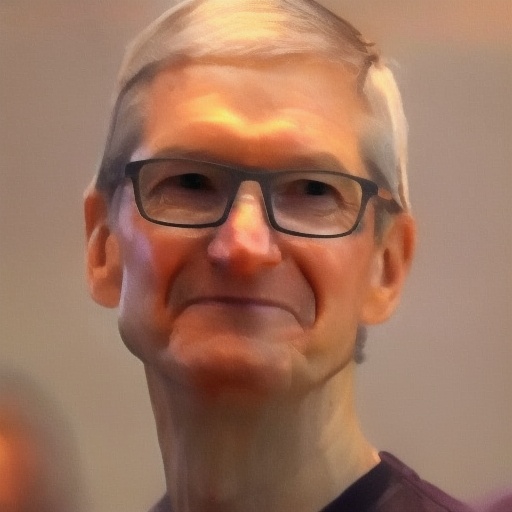}\\
    \includegraphics[width=1\textwidth]{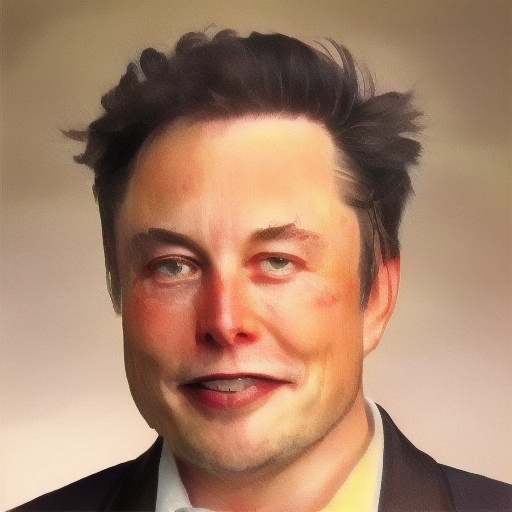}\\
    \includegraphics[width=1\textwidth]{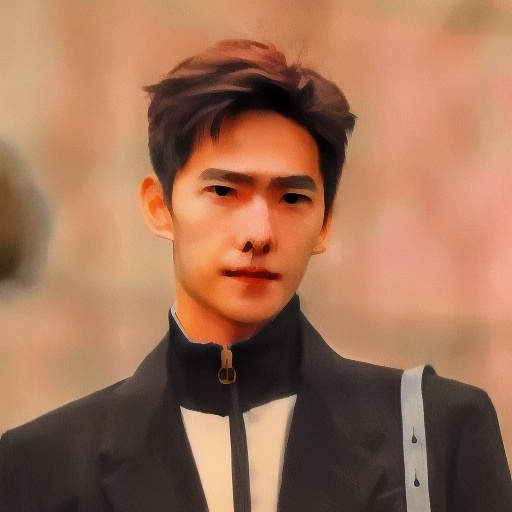}\\
    \includegraphics[width=1\textwidth]{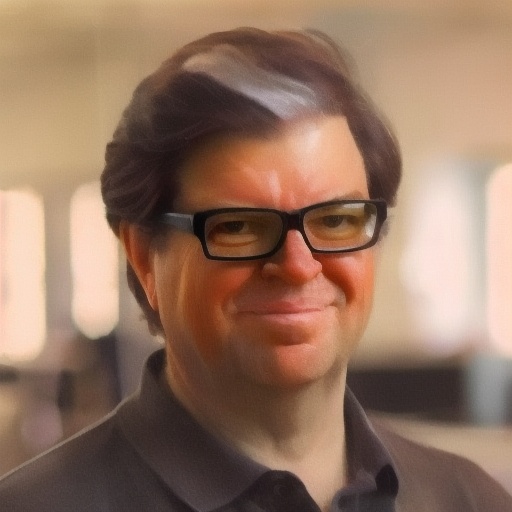}\\
    \includegraphics[width=1\textwidth]{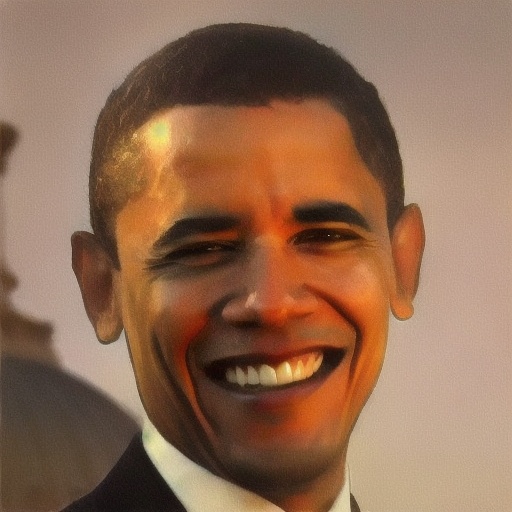}\\
    \includegraphics[width=1\textwidth]{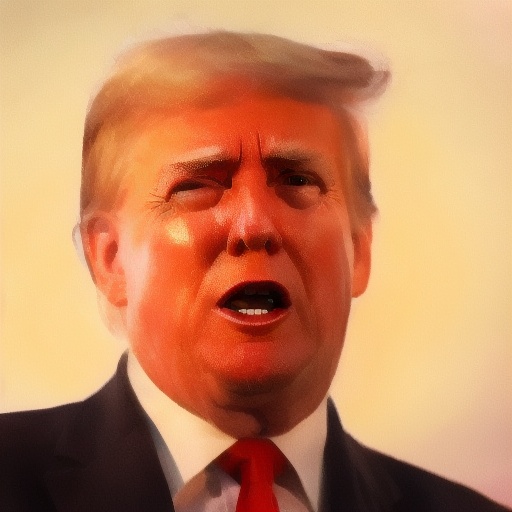}
\end{minipage}

\caption{Generation results of MagicStyle with various of contents and styles, $\alpha=0.8, \beta=0.2$. MagicStyle can generate stylized images for portrait content images of different genders, ages and colors, as well as for various style images.}
\label{fig:styles}
\end{figure*}

\begin{figure*}[!h]
\centering
\begin{minipage}[b]{0.105\textwidth}
    \includegraphics[width=1\textwidth]{Figures/Content_Style.jpg}\\
    \vspace{4pt} 
    \includegraphics[width=1\textwidth]{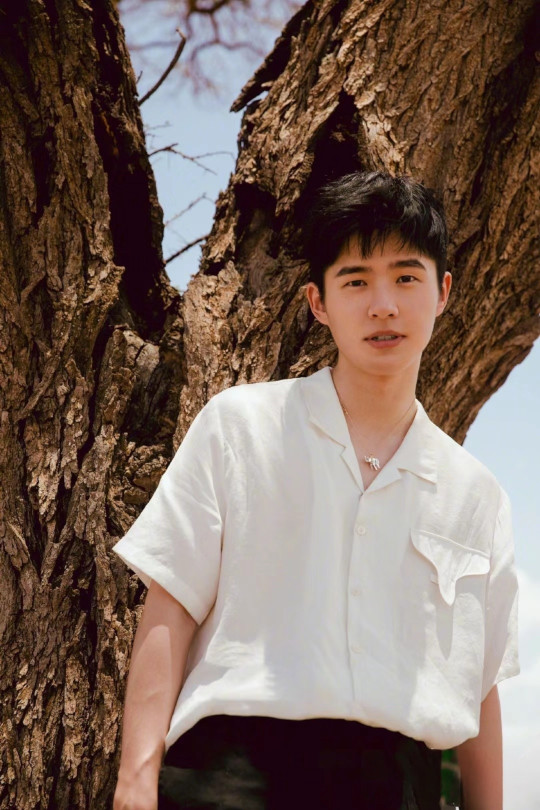}\\
    \vspace{10pt} 
    \includegraphics[ width=1\textwidth]{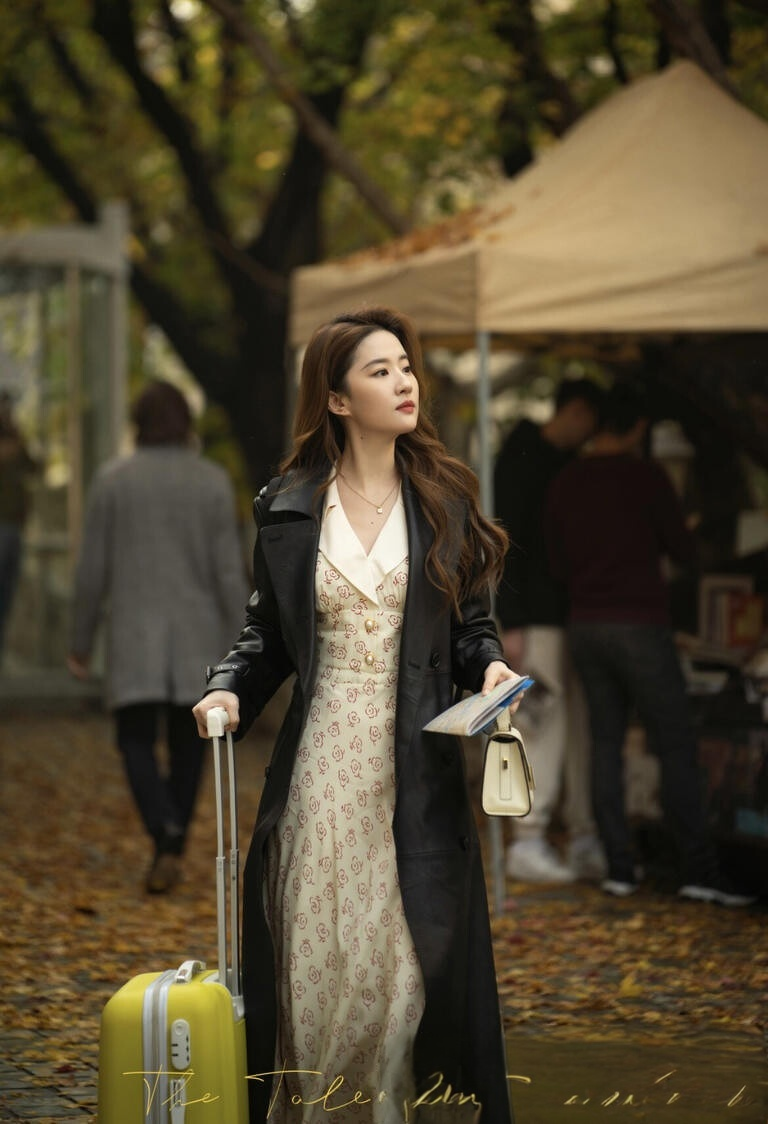}\\
    \vspace{5pt} 
\end{minipage}
\begin{minipage}[b]{0.105\textwidth}
    \includegraphics[clip, trim=0 105 0 105, width=1\textwidth]{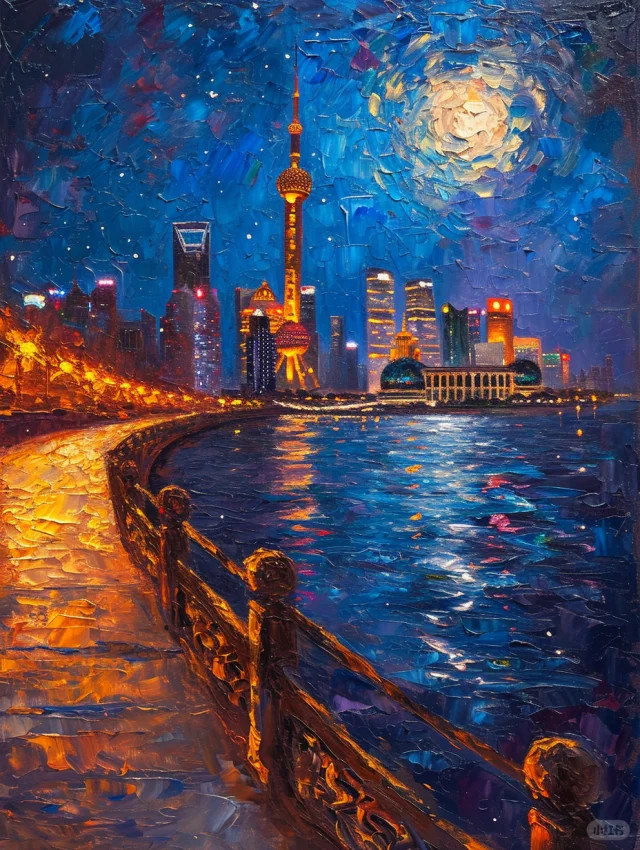}\\
    \includegraphics[width=1\textwidth]{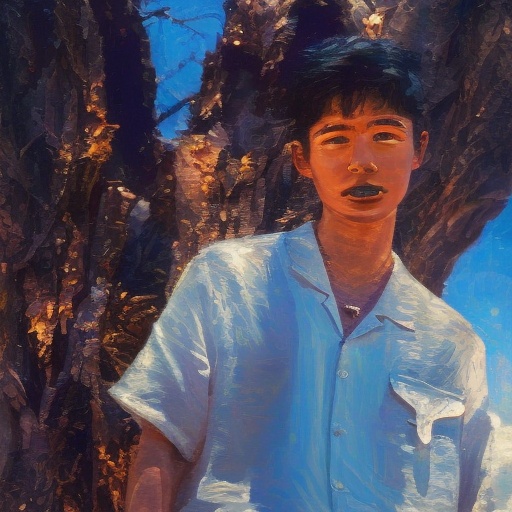}\\
    \includegraphics[width=1\textwidth]{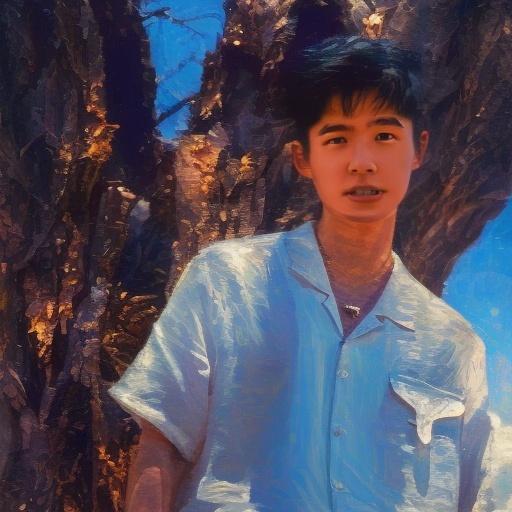}\\
    \includegraphics[width=1\textwidth]{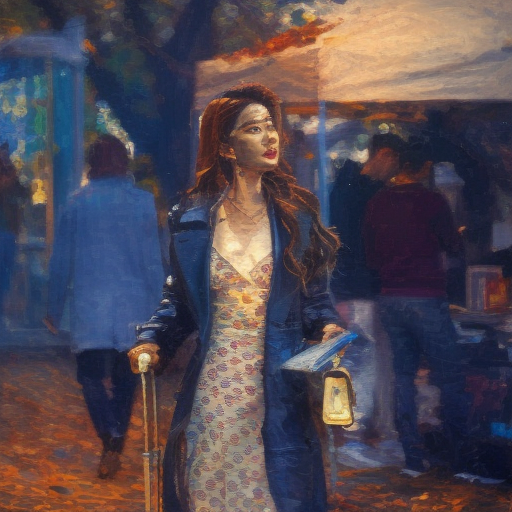}\\
    \includegraphics[width=1\textwidth]{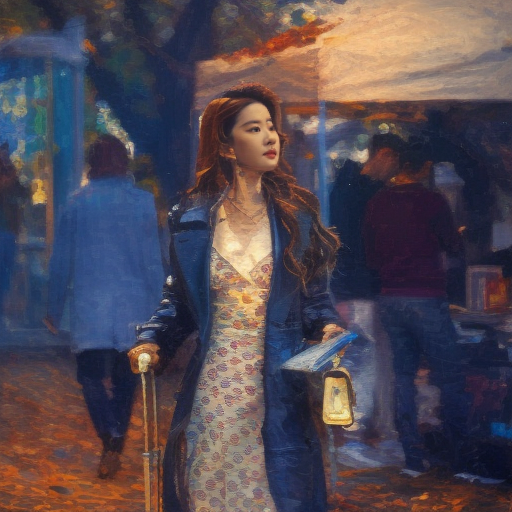}
\end{minipage}
\begin{minipage}[b]{0.105\textwidth}
    \includegraphics[clip, trim=0 105 0 105, width=1\textwidth]{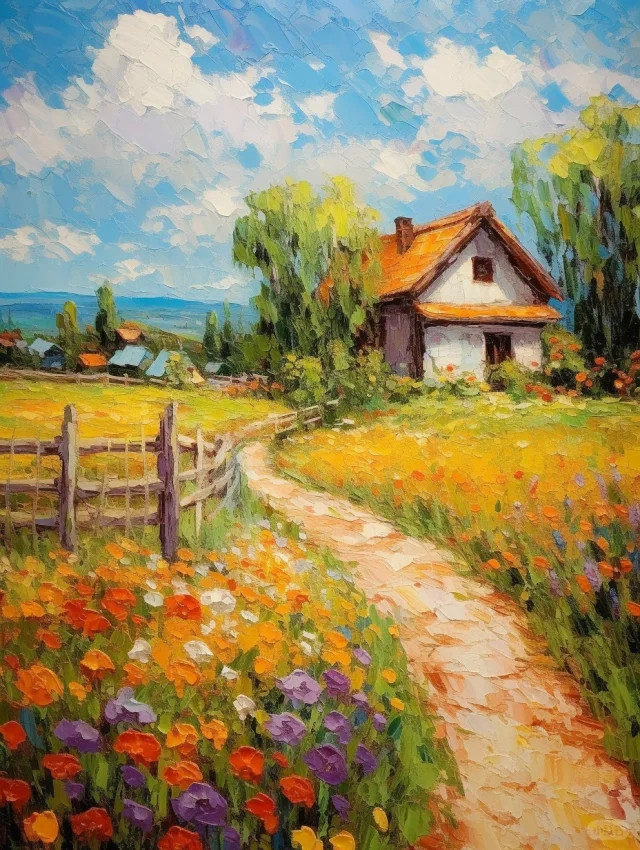}\\
    \includegraphics[width=1\textwidth]{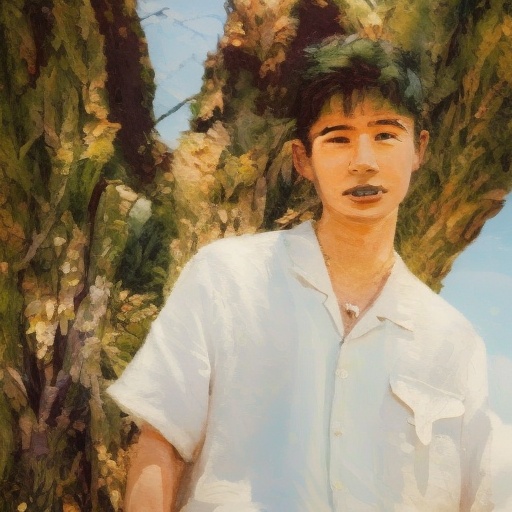}\\
    \includegraphics[width=1\textwidth]{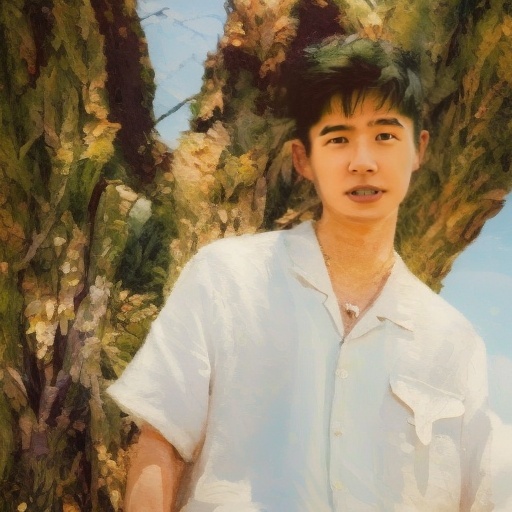}\\
    \includegraphics[width=1\textwidth]{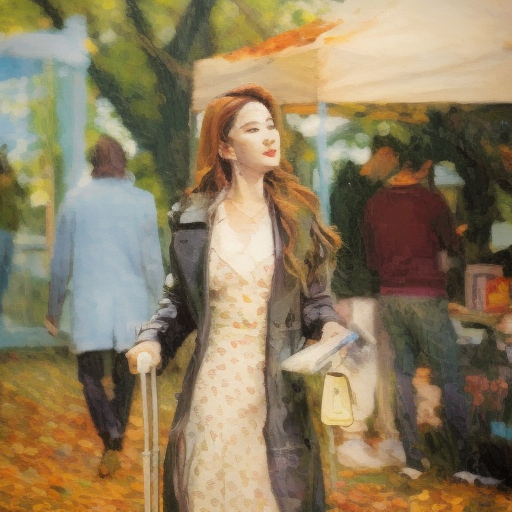}\\
    \includegraphics[width=1\textwidth]{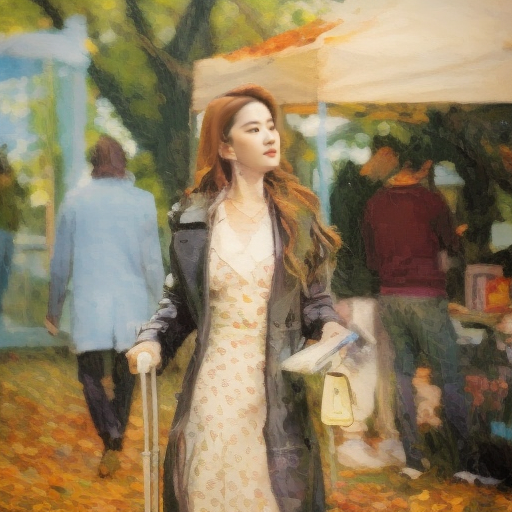}
\end{minipage}
\begin{minipage}[b]{0.105\textwidth}
    \includegraphics[clip, trim=0 220 0 220, width=1\textwidth]{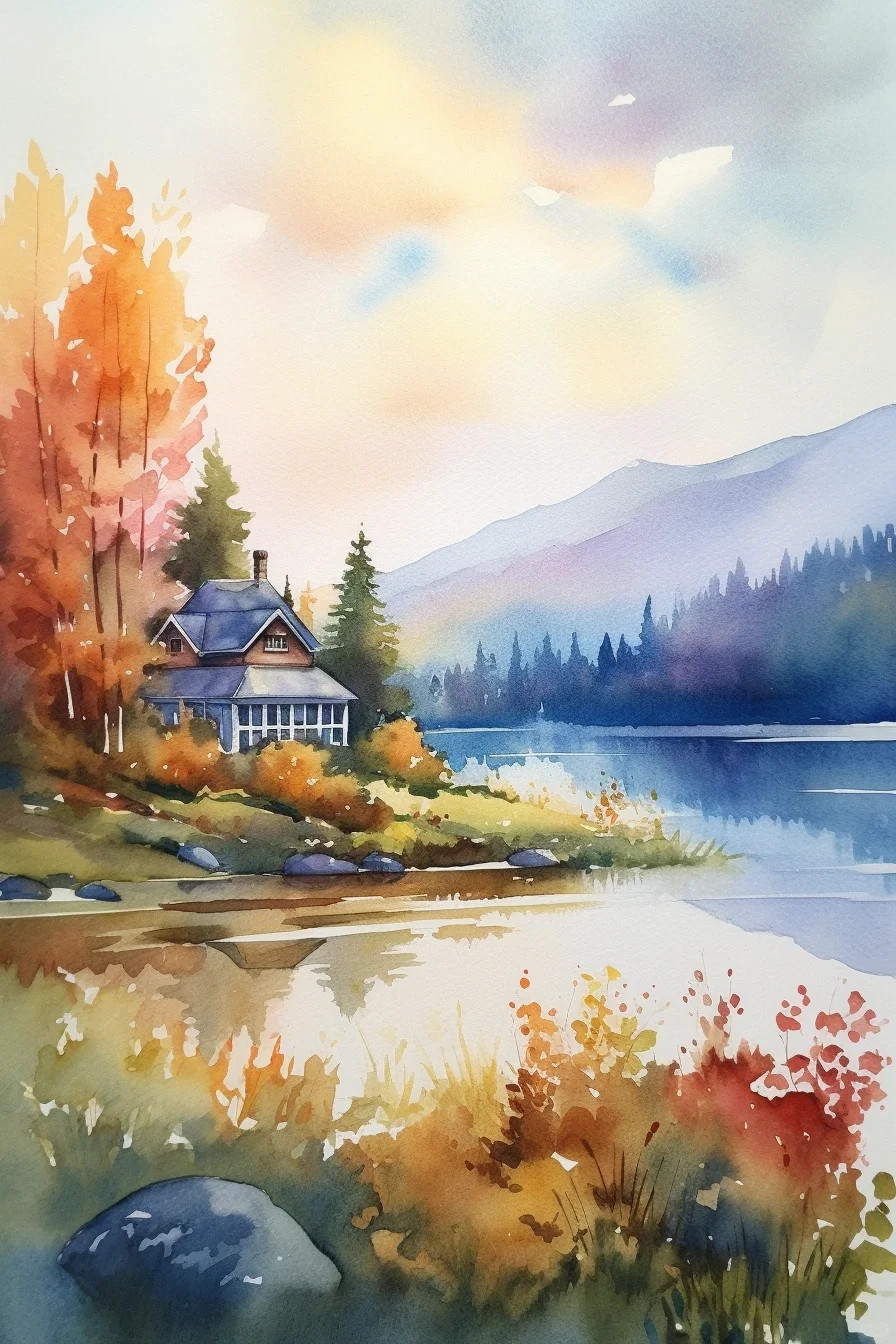}\\
    \includegraphics[width=1\textwidth]{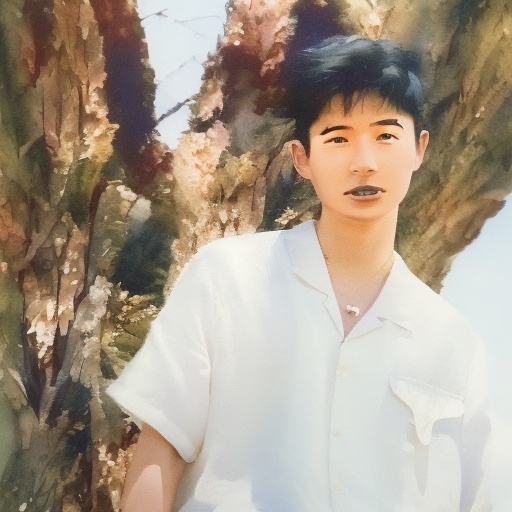}\\
    \includegraphics[width=1\textwidth]{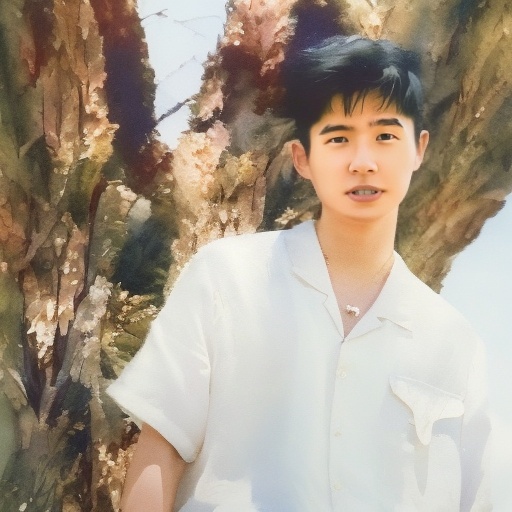}\\
    \includegraphics[width=1\textwidth]{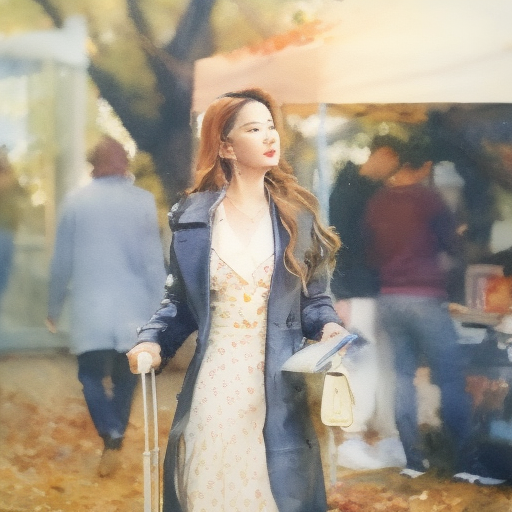}\\
    \includegraphics[width=1\textwidth]{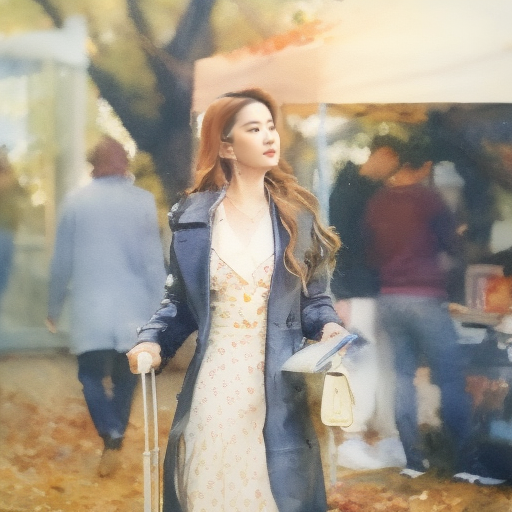}
\end{minipage}
\begin{minipage}[b]{0.105\textwidth}
    \includegraphics[width=1\textwidth,height=0.078\textheight]{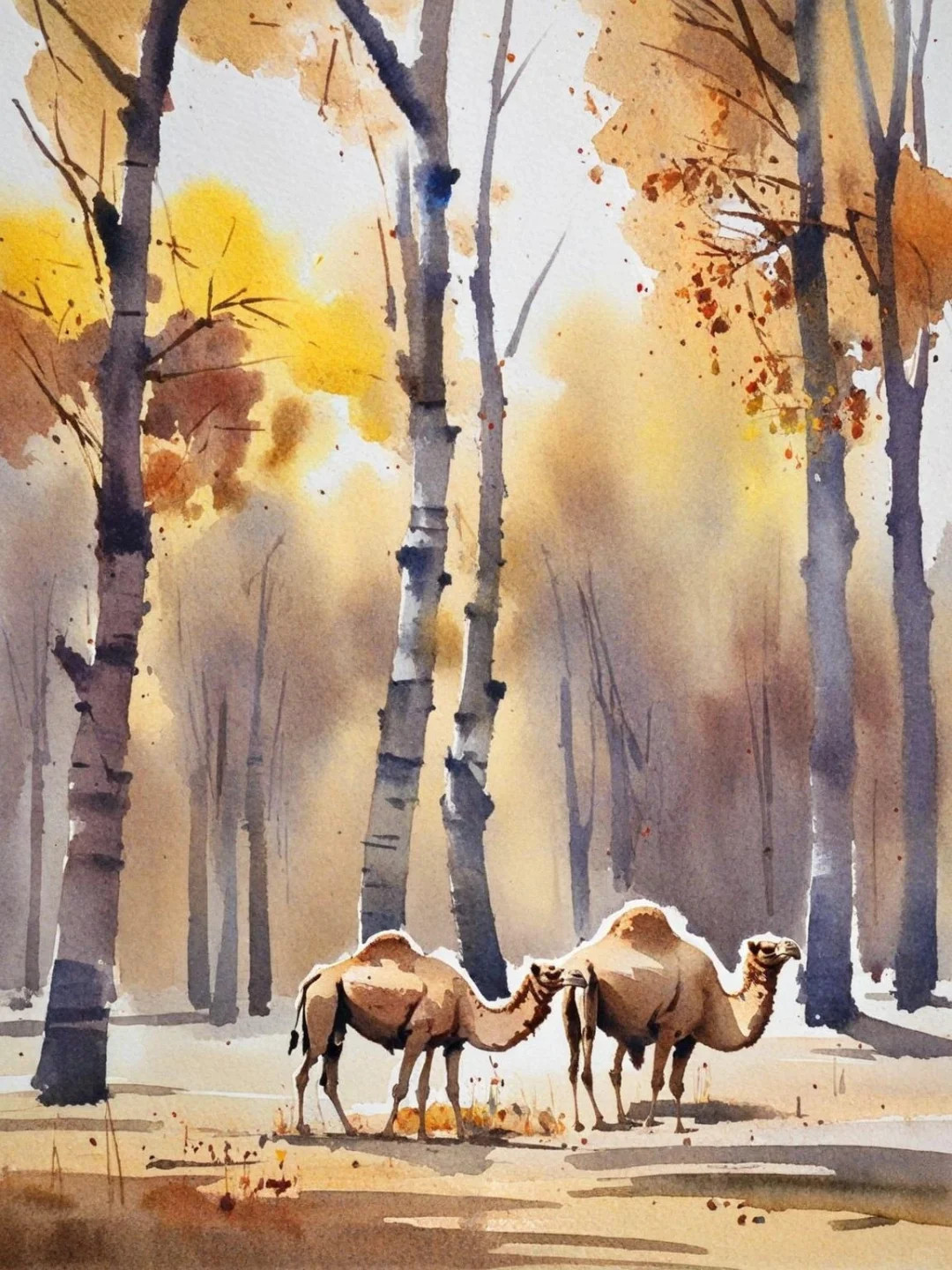}\\
    \includegraphics[width=1\textwidth]{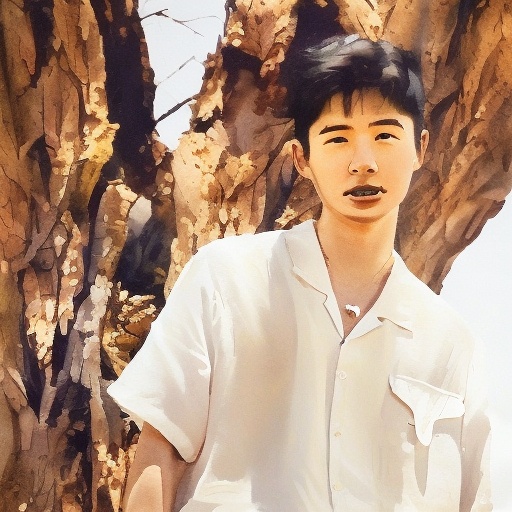}\\
    \includegraphics[width=1\textwidth]{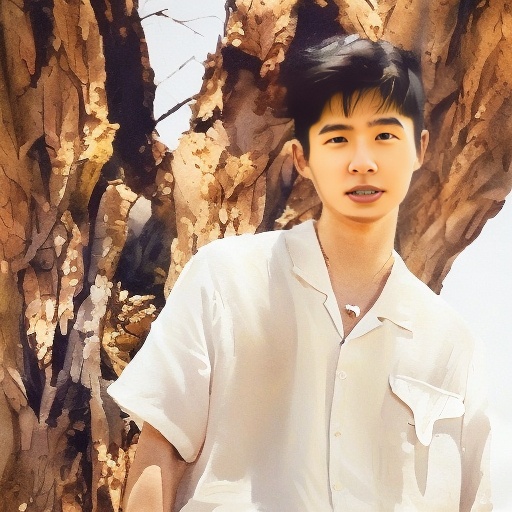}\\
    \includegraphics[width=1\textwidth]{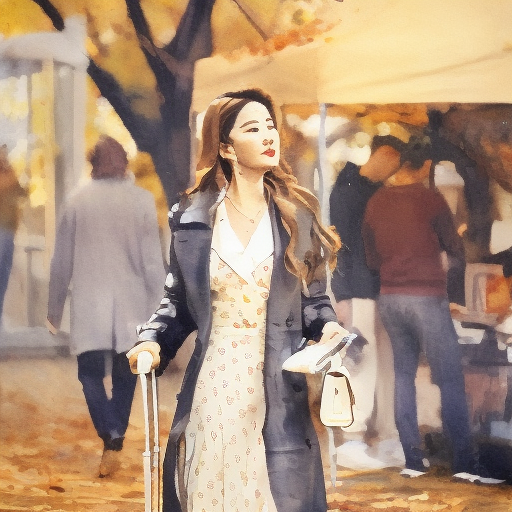}\\
    \includegraphics[width=1\textwidth]{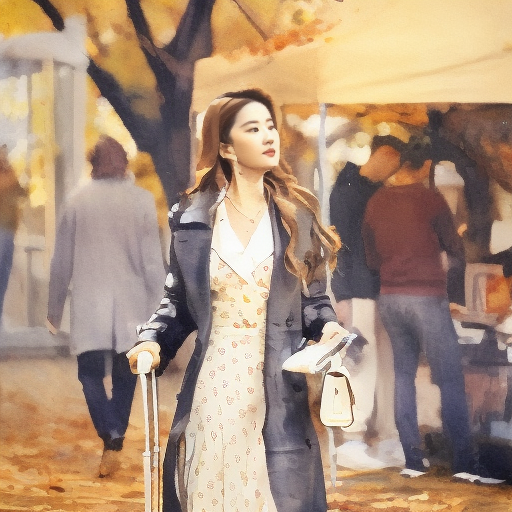}
\end{minipage}
\begin{minipage}[b]{0.105\textwidth}
    \includegraphics[clip, trim=0 105 0 105, width=1\textwidth]{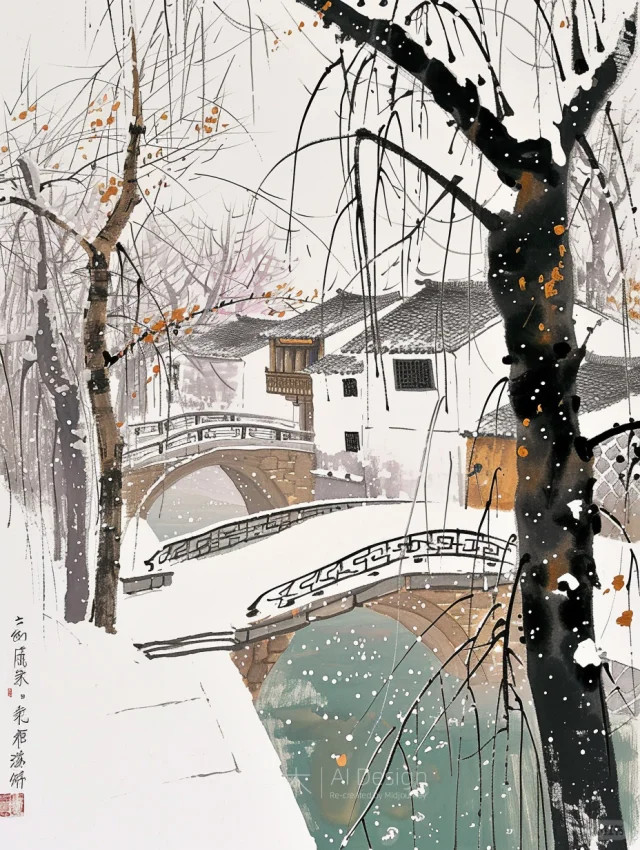}\\
    \includegraphics[width=1\textwidth]{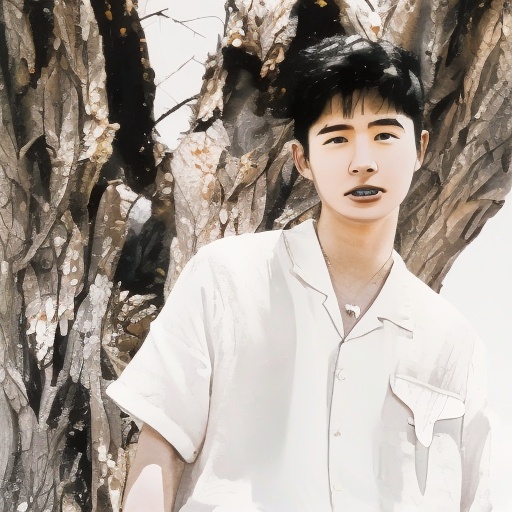}\\
    \includegraphics[width=1\textwidth]{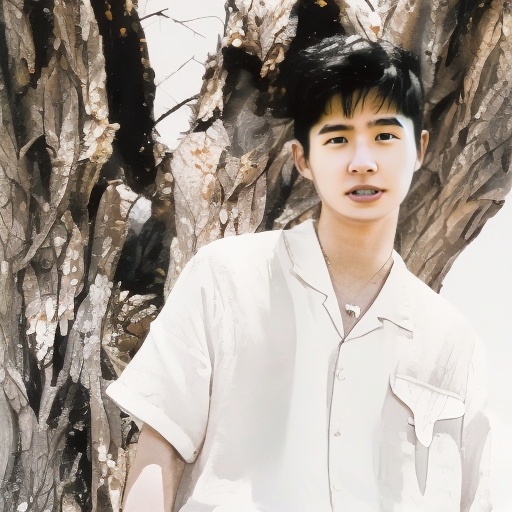}\\
    \includegraphics[width=1\textwidth]{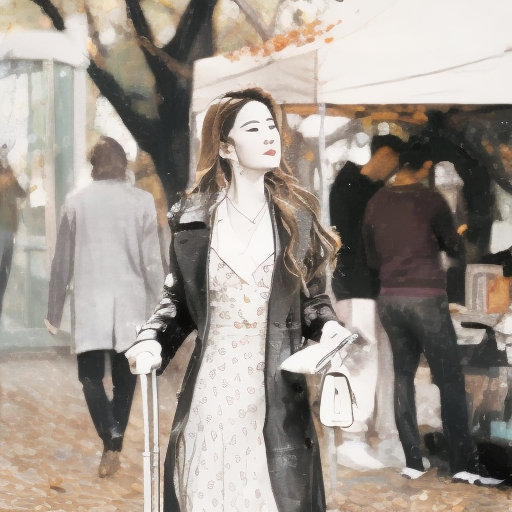}\\
    \includegraphics[width=1\textwidth]{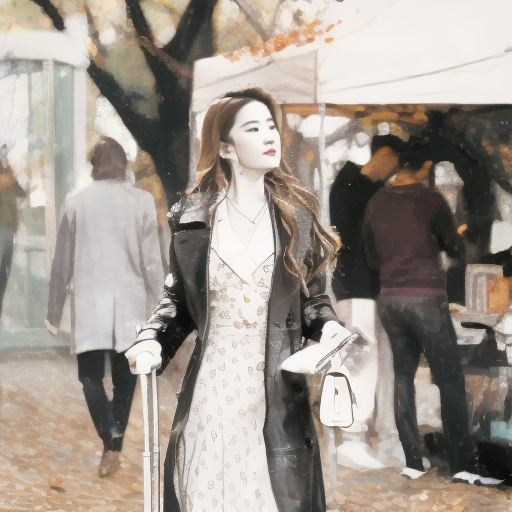}
\end{minipage}
\begin{minipage}[b]{0.105\textwidth}
    \includegraphics[clip, trim=0 100 0 110,width=1\textwidth]{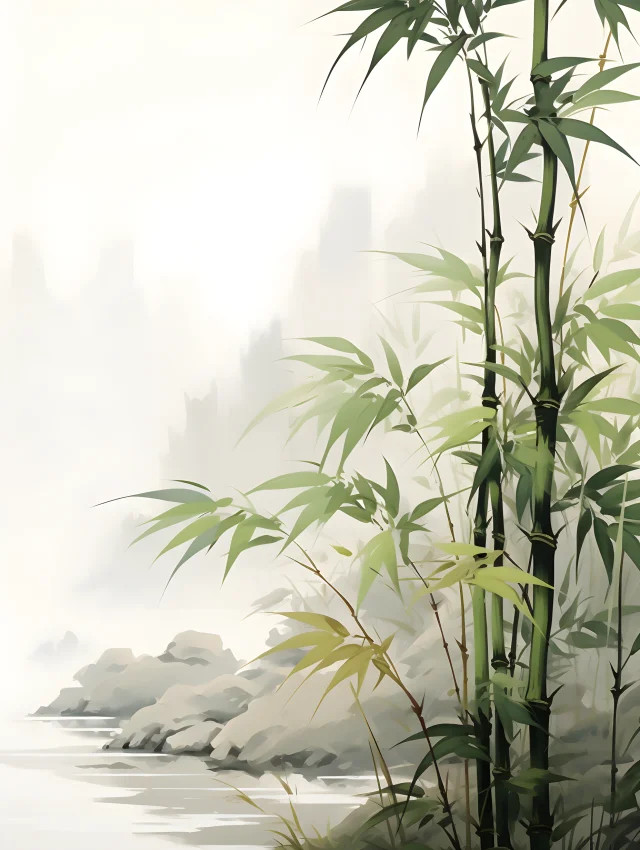}\\
    \includegraphics[width=1\textwidth]{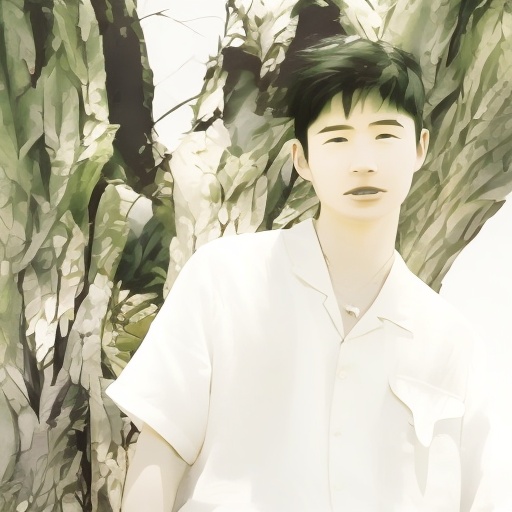}\\
    \includegraphics[width=1\textwidth]{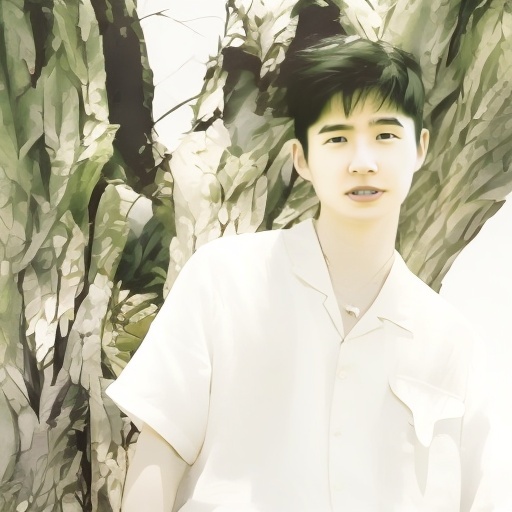}\\
    \includegraphics[width=1\textwidth]{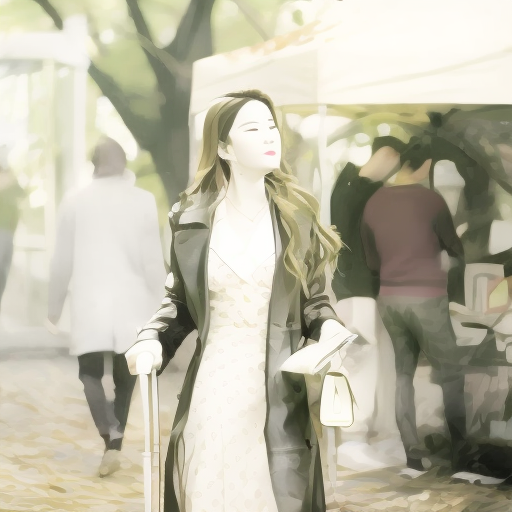}\\
    \includegraphics[width=1\textwidth]{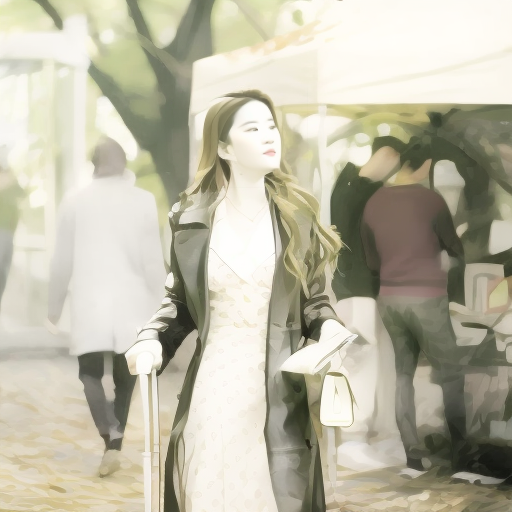}
\end{minipage}
\begin{minipage}[b]{0.105\textwidth}
    \includegraphics[clip, trim=0 180 0 170,width=1\textwidth]{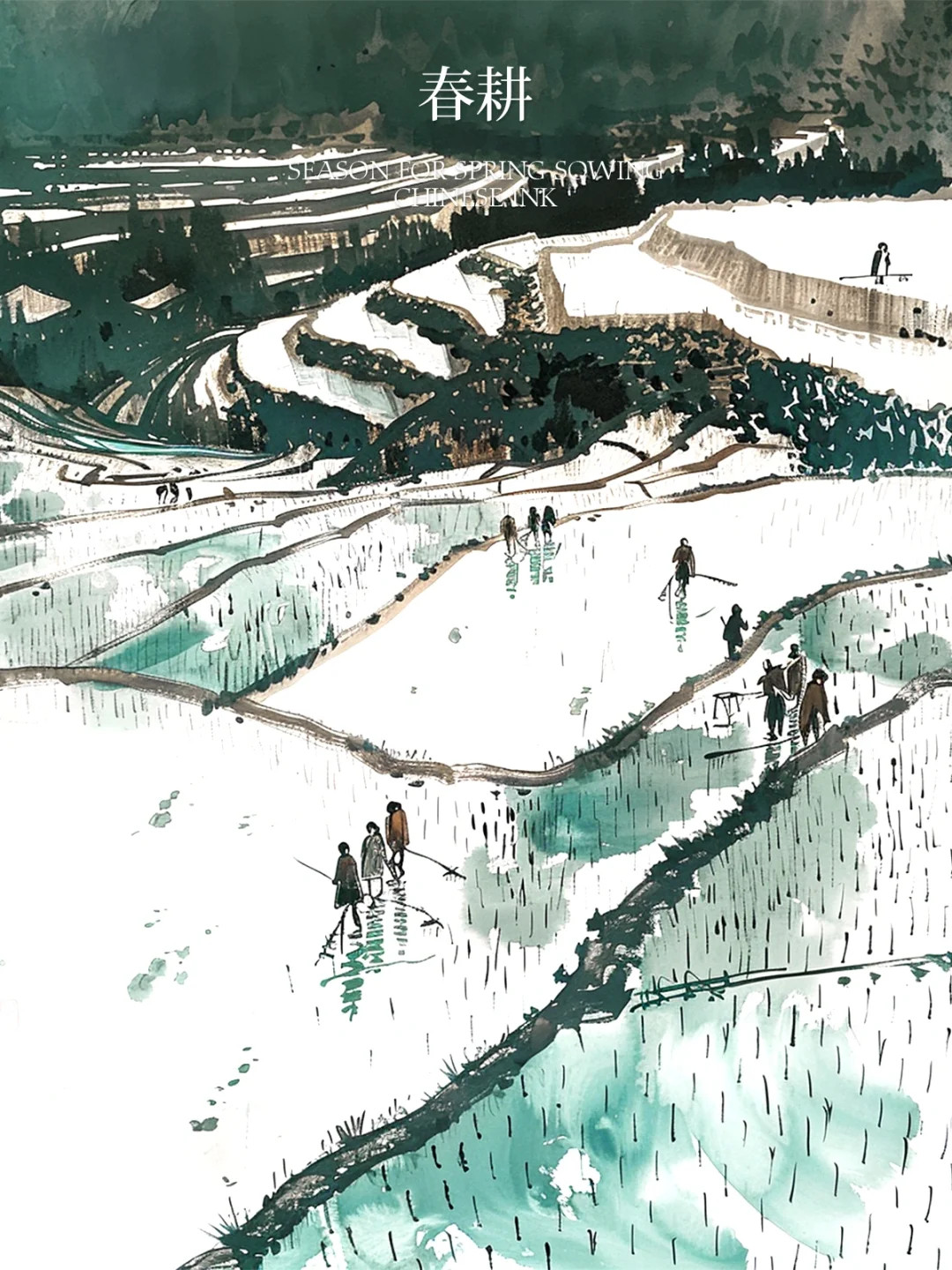}\\
    \includegraphics[width=1\textwidth]{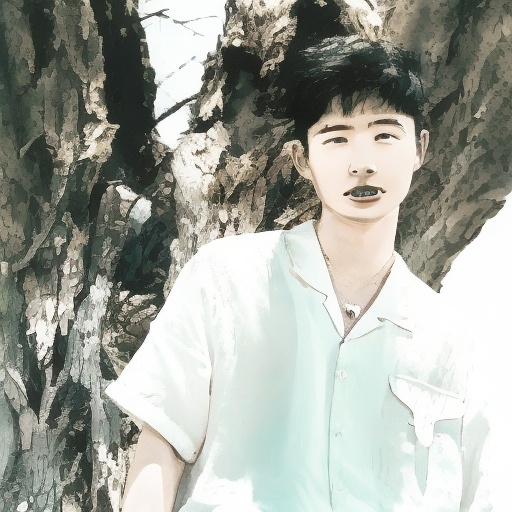}\\
    \includegraphics[width=1\textwidth]{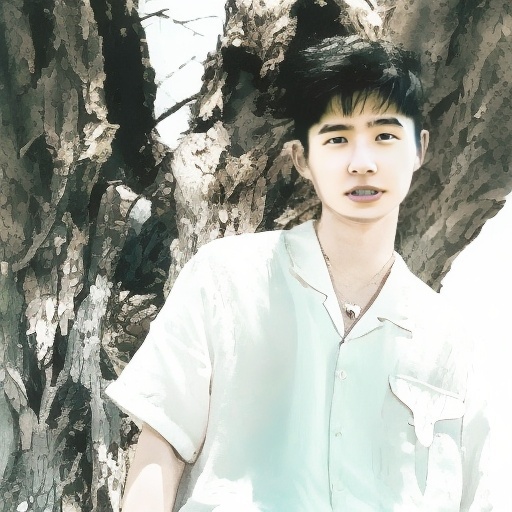}\\
    \includegraphics[width=1\textwidth]{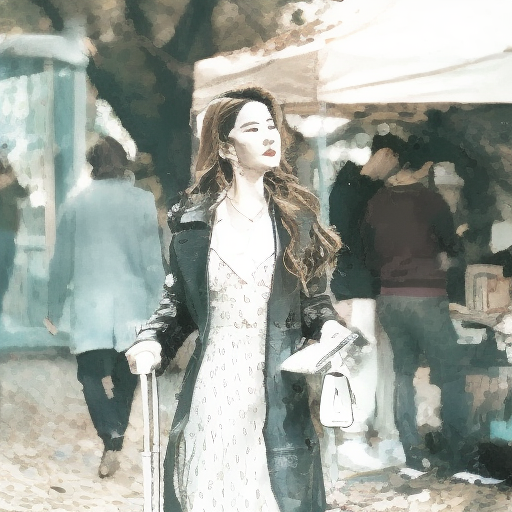}\\
    \includegraphics[width=1\textwidth]{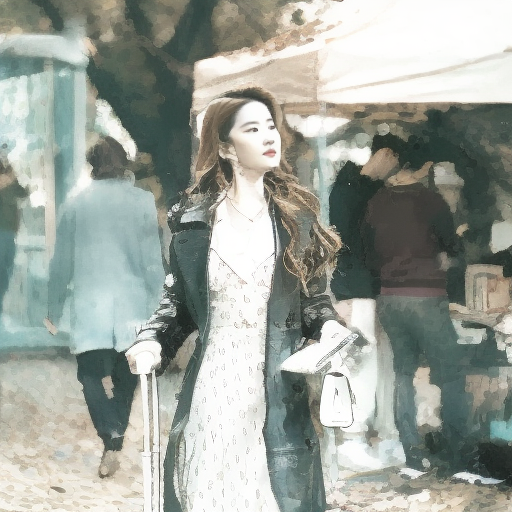}
\end{minipage}
\begin{minipage}[b]{0.105\textwidth}
    \includegraphics[clip, trim=105 0 100 0,width=1\textwidth]{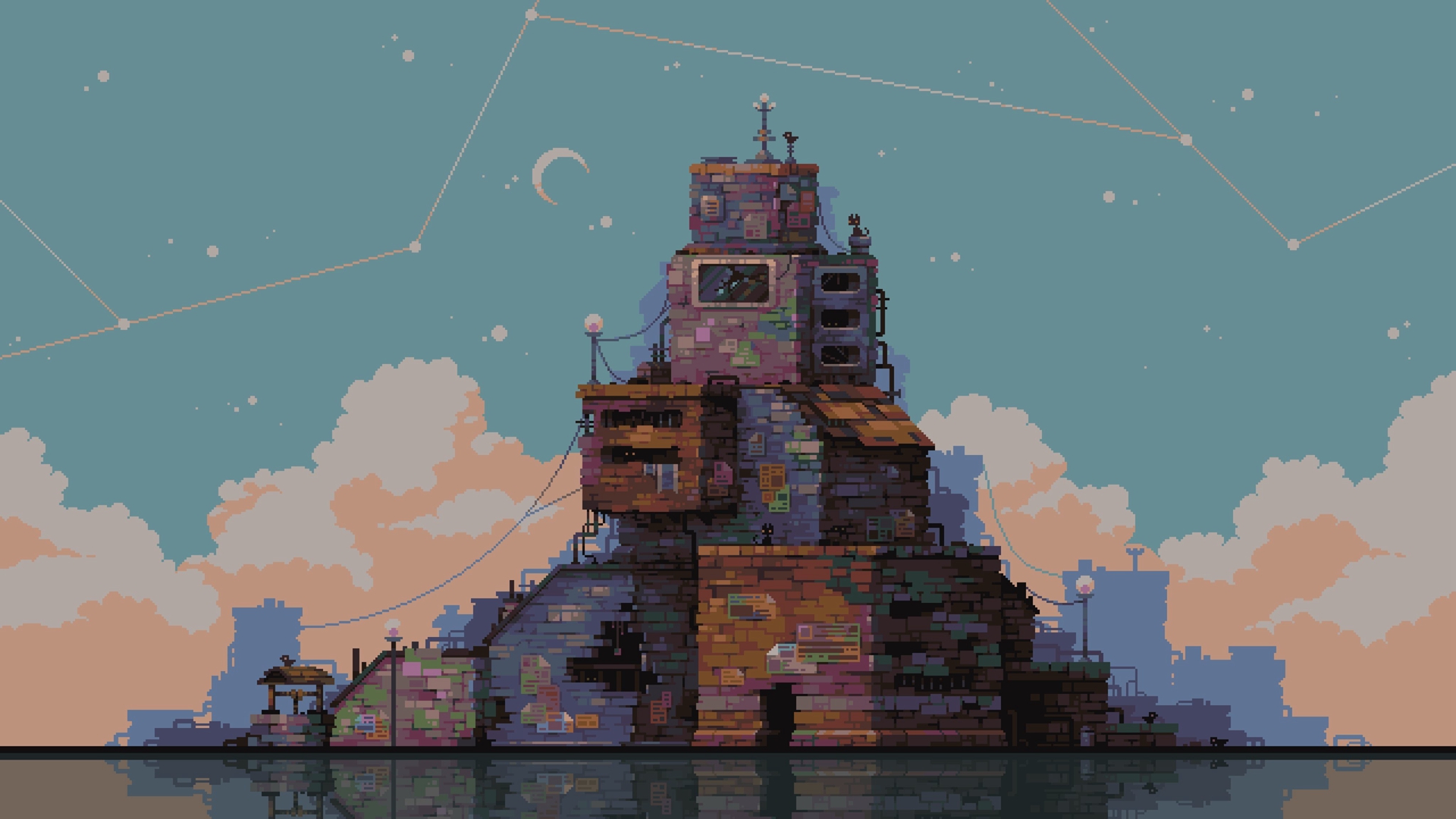}\\
    \includegraphics[width=1\textwidth]{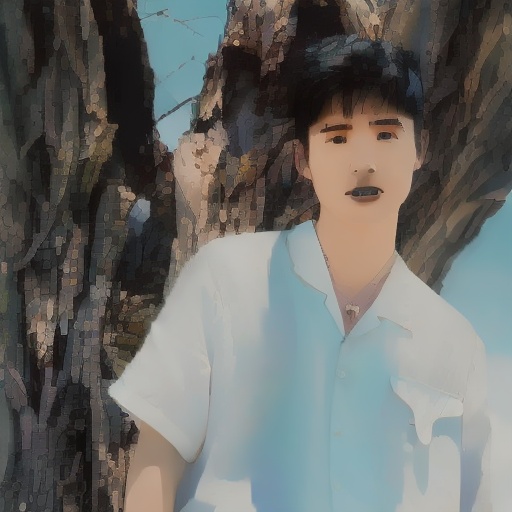}\\
    \includegraphics[width=1\textwidth]{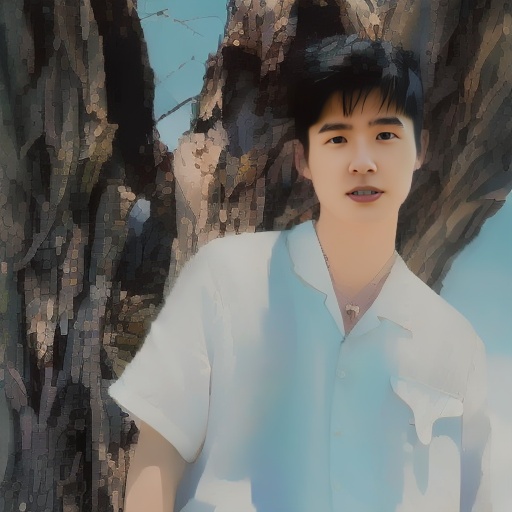}\\
    \includegraphics[width=1\textwidth]{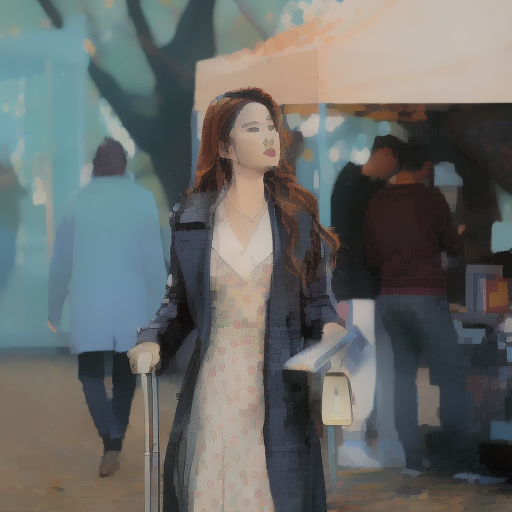}\\
    \includegraphics[width=1\textwidth]{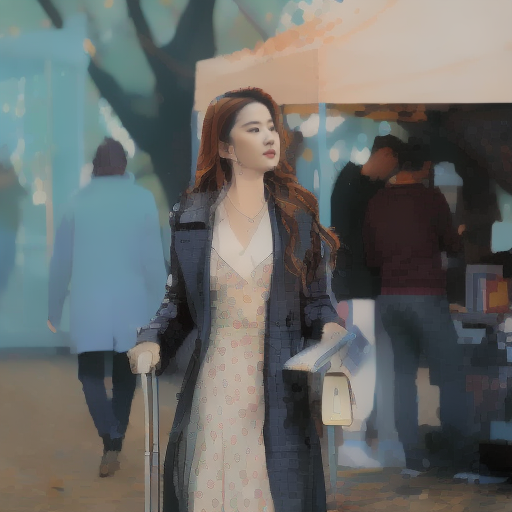}
\end{minipage}

\caption{Generation results of MagicStyle with the face resolution in content image is relatively small, $\alpha=0.8, \beta=0.2$. The second row of each content image are results in coorperated with DIIR proposed in MagicID~\cite{deng2024magicid}.}
\label{fig:smallface}
\end{figure*}

In our experiments, we utilized various content and style images for portrait stylization. The visualization results are shown in Fig. \ref{fig:styles}. As can be seen, our proposed MagicStyle effectively incorporates the texture style from the style image into portraits of varying genders and ages, resulting in stylized image generation. Moreover, the details of the content image, such as facial identity, expressions, and background, are well preserved. 
In conjunction with DIIR proposed in MagicID~\cite{deng2024magicid}, as illustrated in Fig. \ref{fig:smallface}, our MagicStyle can be extended to apply to scenarios where the face occupies a relatively small portion of the scene.

\begin{figure*}[t]
\centering
\begin{minipage}[b]{0.112\textwidth}
    \makebox[1\textwidth]{\raisebox{0.5em}{\vphantom{X} Content}}\\
    \includegraphics[width=1\textwidth]{Figures/input_img/content_crop/chenglong.jpeg}\\
    \includegraphics[width=1\textwidth]{Figures/input_img/content_crop/Feifei.jpg}\\
    \includegraphics[width=1\textwidth]{Figures/input_img/content_crop/hgf.jpeg}\\
    \includegraphics[width=1\textwidth]{Figures/input_img/content_crop/kuke.jpeg}\\
    \includegraphics[width=1\textwidth]{Figures/input_img/content_crop/yangyang.jpeg}\\
    \includegraphics[width=1\textwidth]{Figures/input_img/content_crop/Musk2.jpg}
\end{minipage}
\begin{minipage}[b]{0.112\textwidth}
    \makebox[1\textwidth]{\raisebox{0.5em}{\vphantom{X} Style}} \\
    \includegraphics[width=1\textwidth]{Figures/input_img/style/andry-dedouze-train-final.jpg}\\
    \includegraphics[width=1\textwidth]{Figures/input_img/style/mannai_sunset.jpg}\\
    \includegraphics[width=1\textwidth]{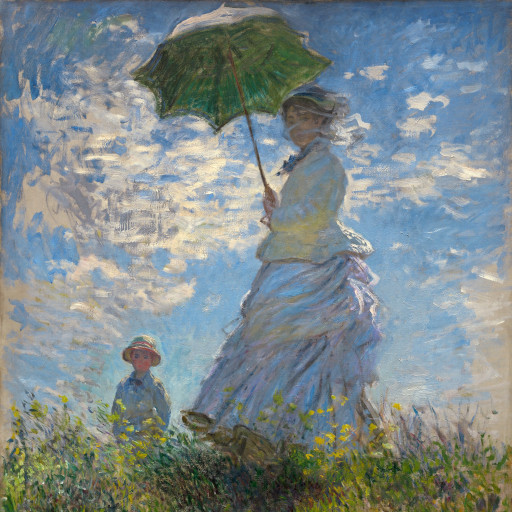}\\
    \includegraphics[width=1\textwidth]{Figures/input_img/style/ink_boy.jpg}\\
    \includegraphics[width=1\textwidth,height=0.085\textheight]{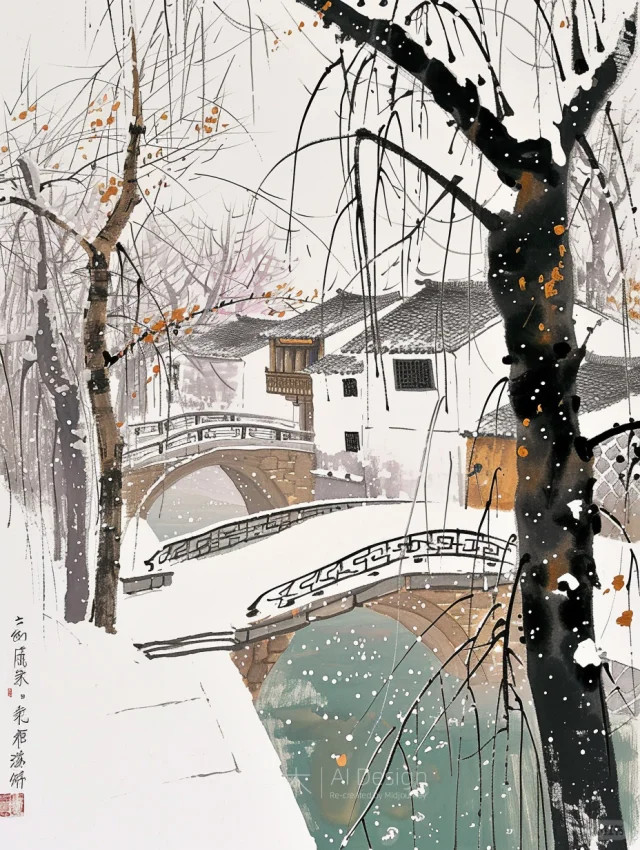}\\
    \includegraphics[width=1\textwidth]{Figures/input_img/style/vangogh_flower.jpg}
\end{minipage}
\begin{minipage}[b]{0.112\textwidth}
    \makebox[1\textwidth]{\raisebox{0.5em}{\vphantom{X} InstantID}} \\
    \includegraphics[width=1\textwidth]{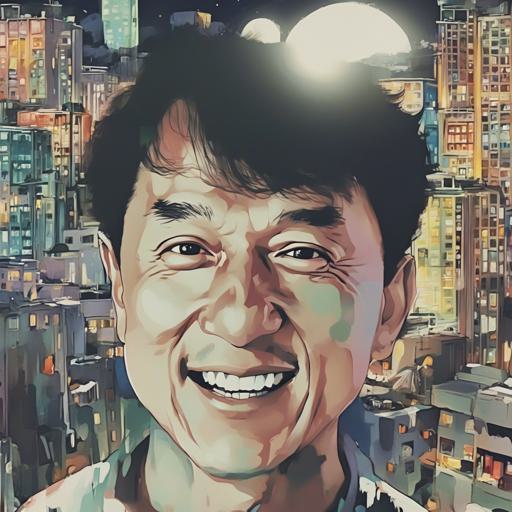}\\
    \includegraphics[width=1\textwidth]{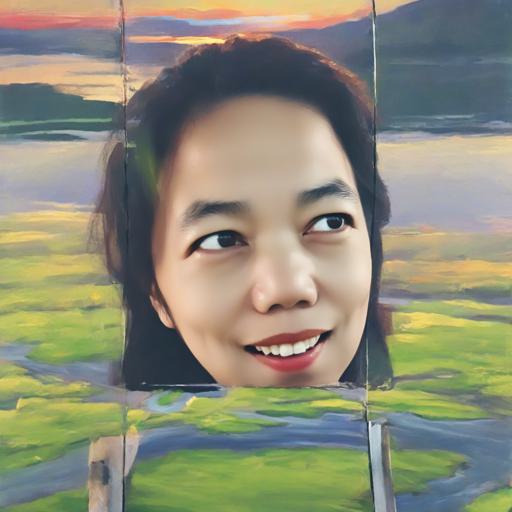}\\
    \includegraphics[width=1\textwidth]{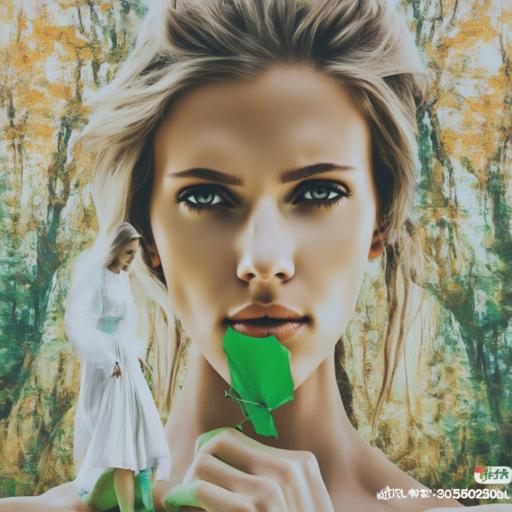}\\
    \includegraphics[width=1\textwidth]{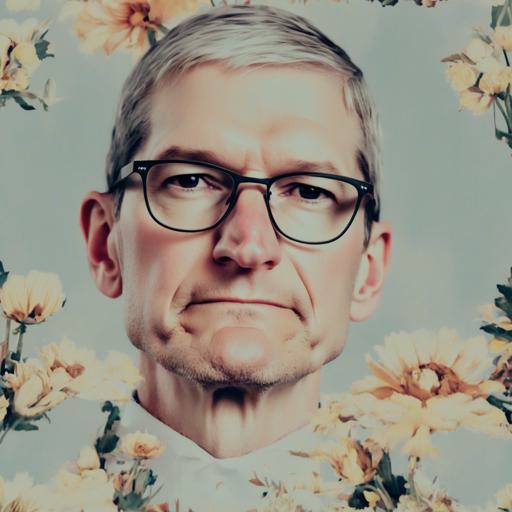}\\
    \includegraphics[width=1\textwidth]{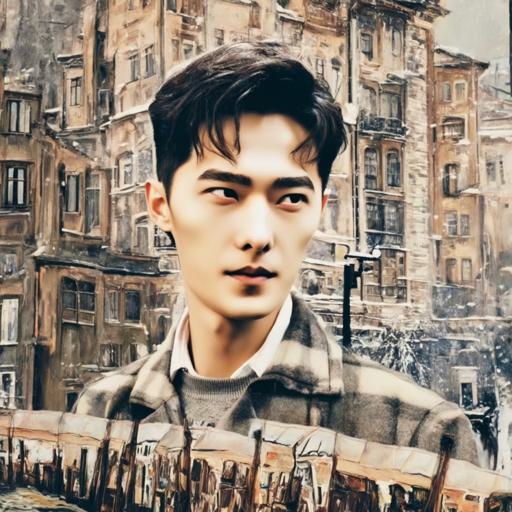}\\
    \includegraphics[width=1\textwidth]{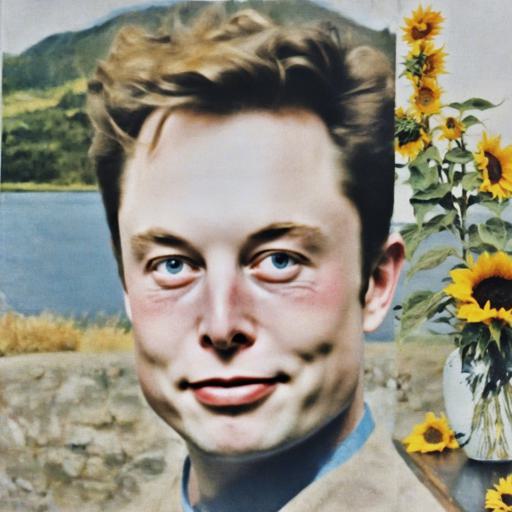}
\end{minipage}
\begin{minipage}[b]{0.112\textwidth}
    \makebox[1\textwidth]{\raisebox{0.5em}{\vphantom{X} PhotoMakerV2}} \\
    \includegraphics[width=1\textwidth]{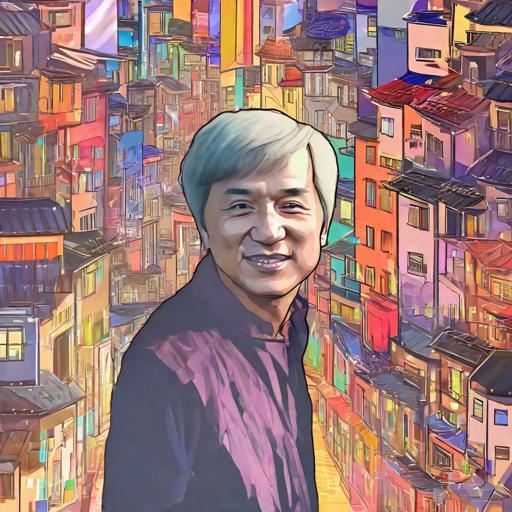}\\
    \includegraphics[width=1\textwidth]{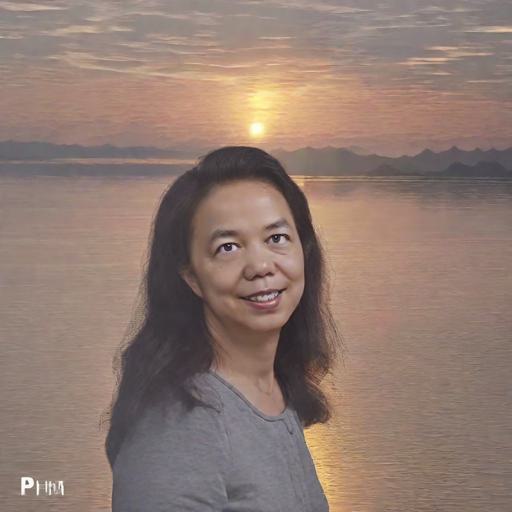}\\
    \includegraphics[width=1\textwidth]{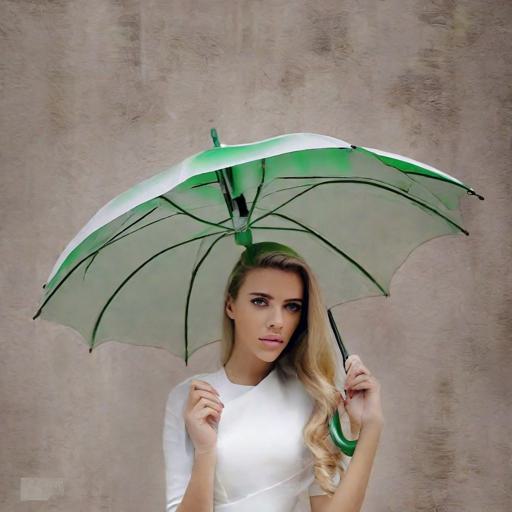}\\
    \includegraphics[width=1\textwidth]{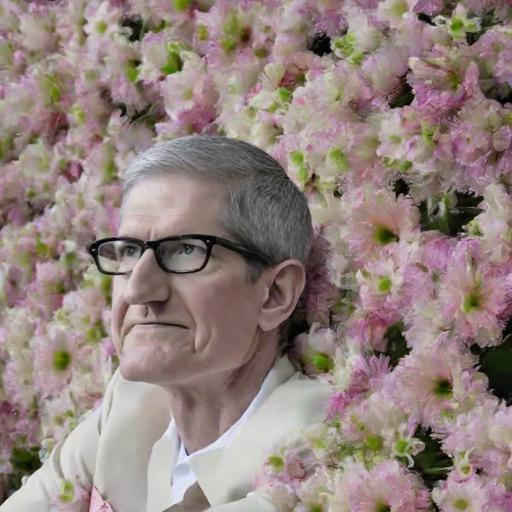}\\
    \includegraphics[width=1\textwidth]{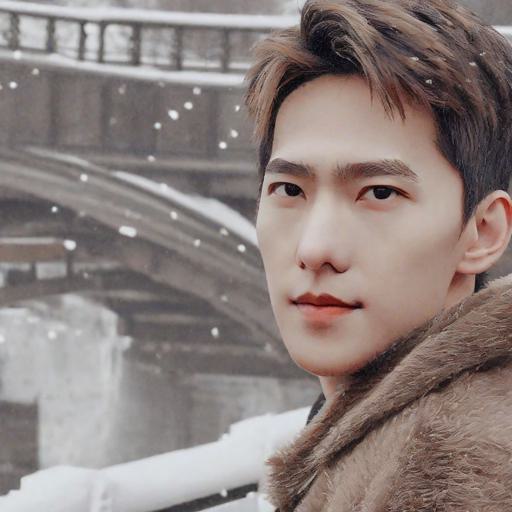}\\
    \includegraphics[width=1\textwidth]{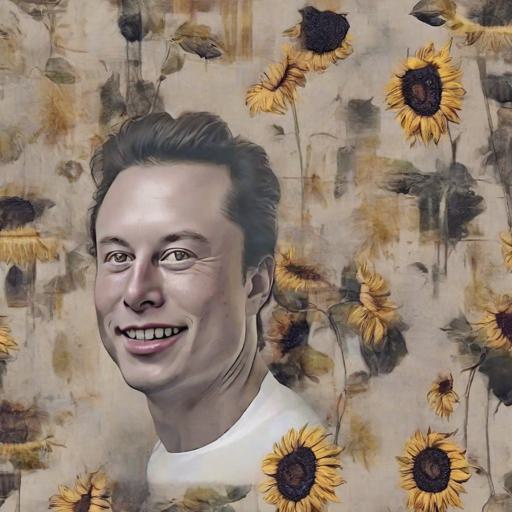}
\end{minipage}
\begin{minipage}[b]{0.112\textwidth}
    \makebox[1\textwidth]{\raisebox{0.5em}{\vphantom{X} StyleAligned}} \\
    \includegraphics[width=1\textwidth]{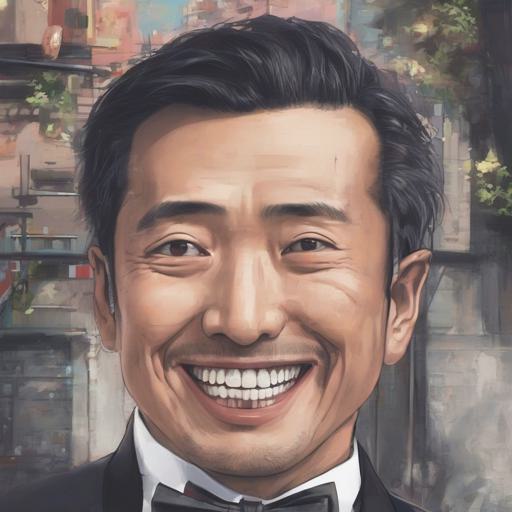}\\
    \includegraphics[width=1\textwidth]{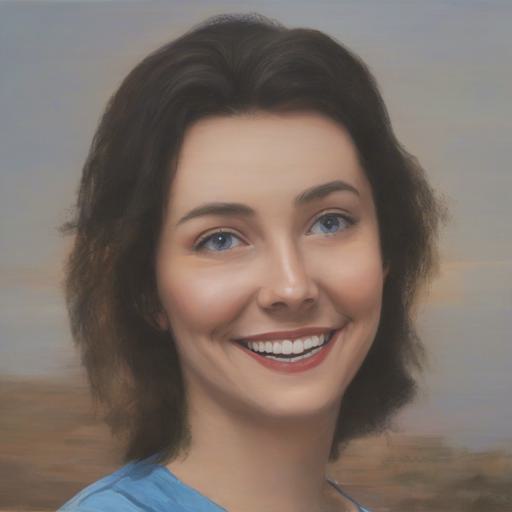}\\
    \includegraphics[width=1\textwidth]{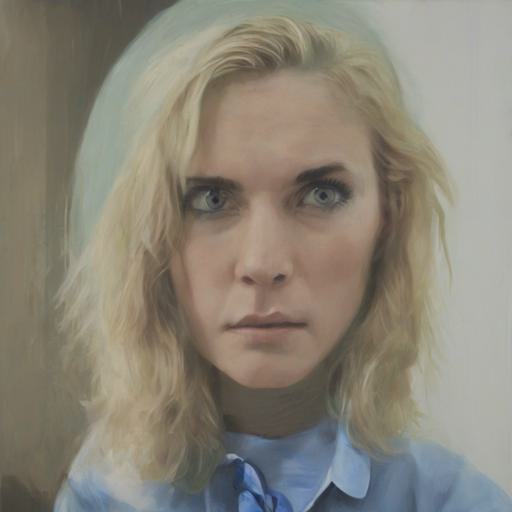}\\
    \includegraphics[width=1\textwidth]{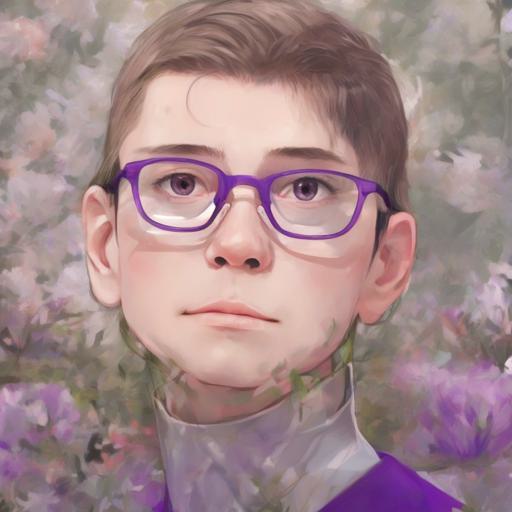}\\
    \includegraphics[width=1\textwidth]{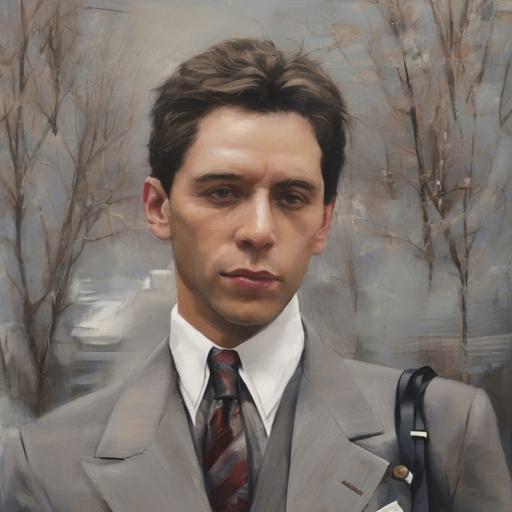}\\
    \includegraphics[width=1\textwidth]{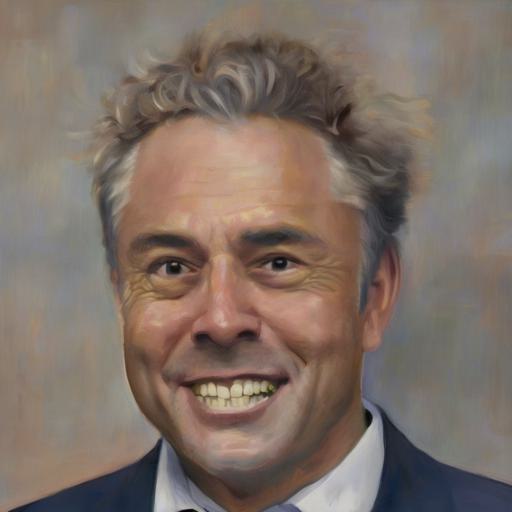}
\end{minipage}
\begin{minipage}[b]{0.112\textwidth}
    \makebox[1\textwidth]{\raisebox{0.5em}{\vphantom{X} CSGO}} \\
    \includegraphics[width=1\textwidth]{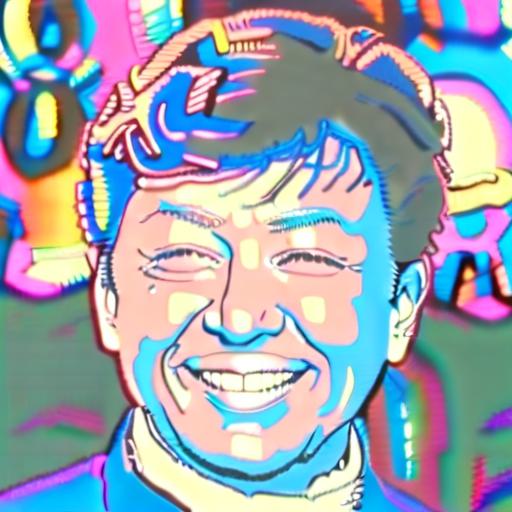}\\
    \includegraphics[width=1\textwidth]{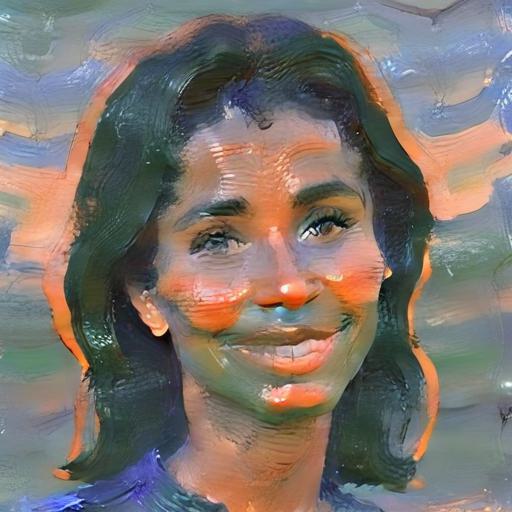}\\
    \includegraphics[width=1\textwidth]{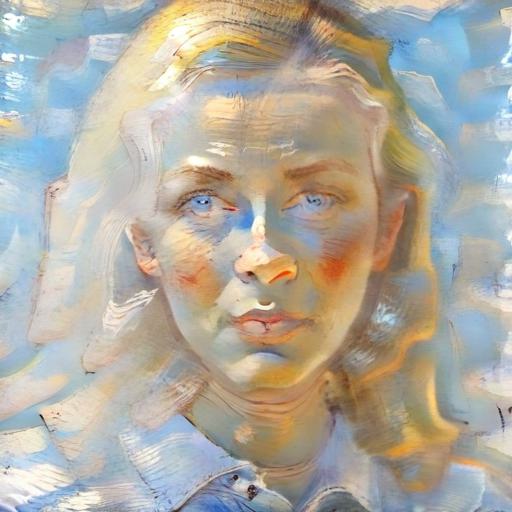}\\
    \includegraphics[width=1\textwidth]{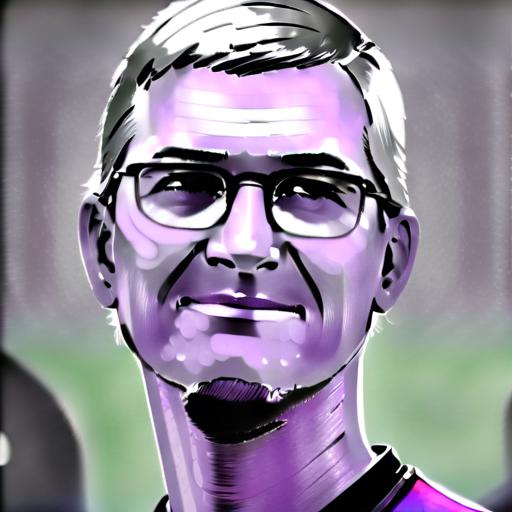}\\
    \includegraphics[width=1\textwidth]{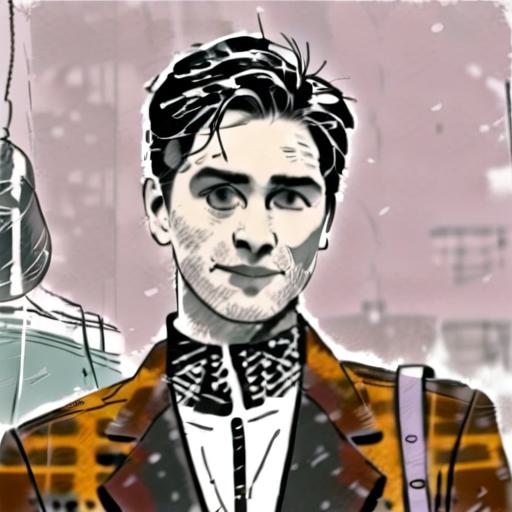}\\
    \includegraphics[width=1\textwidth]{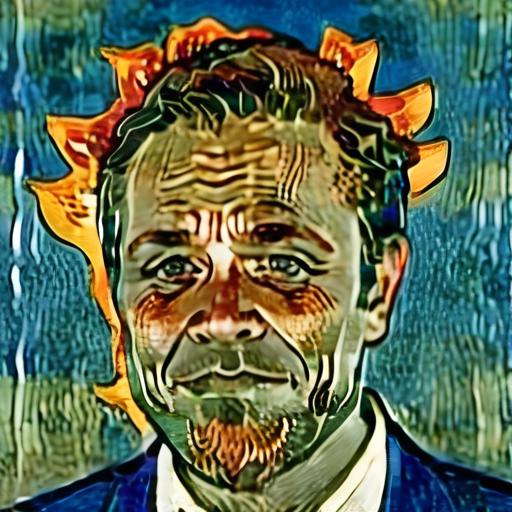}
\end{minipage}
\begin{minipage}[b]{0.112\textwidth}
    \makebox[1\textwidth]{\raisebox{0.5em}{\vphantom{X} StyleID}} \\
    \includegraphics[width=1\textwidth]{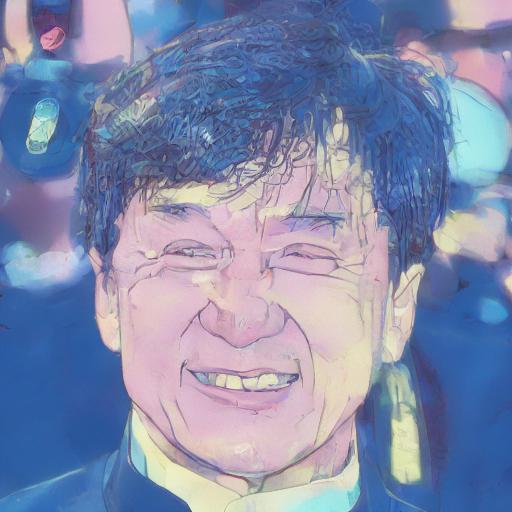}\\
    \includegraphics[width=1\textwidth]{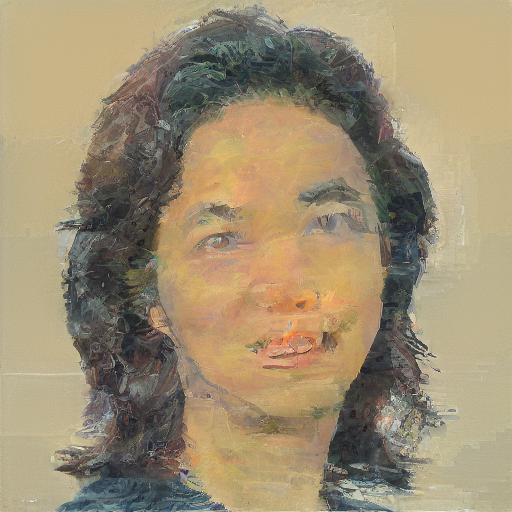}\\
    \includegraphics[width=1\textwidth]{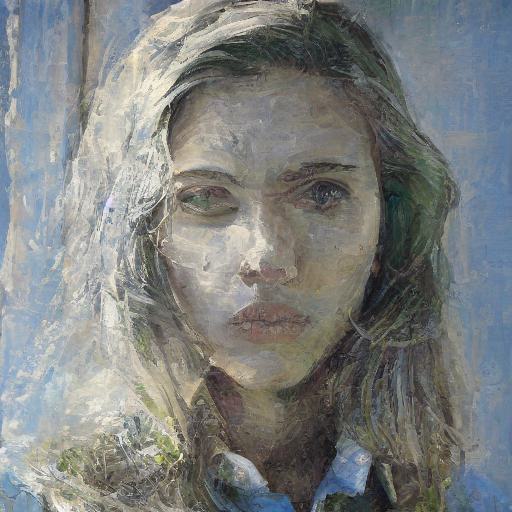}\\
    \includegraphics[width=1\textwidth]{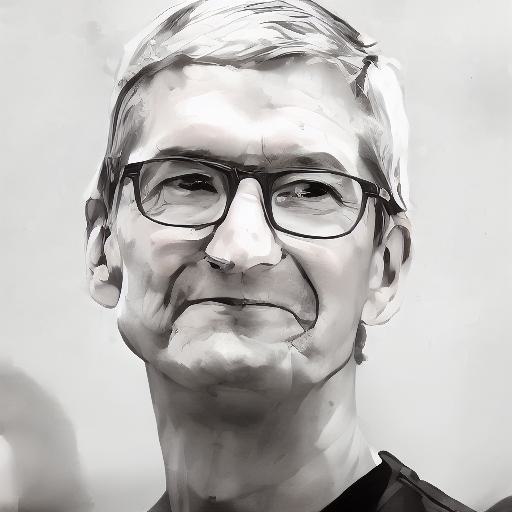}\\
    \includegraphics[width=1\textwidth]{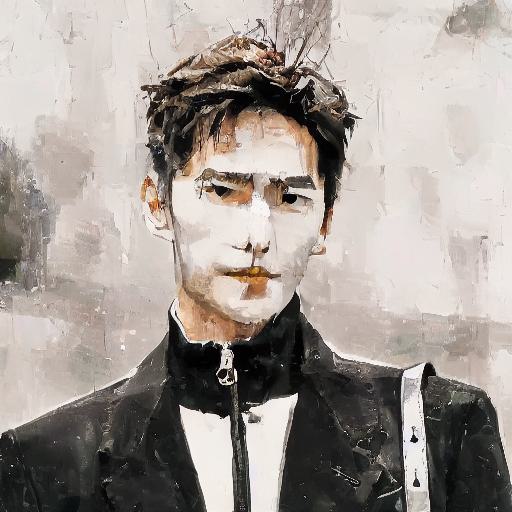}\\
    \includegraphics[width=1\textwidth]{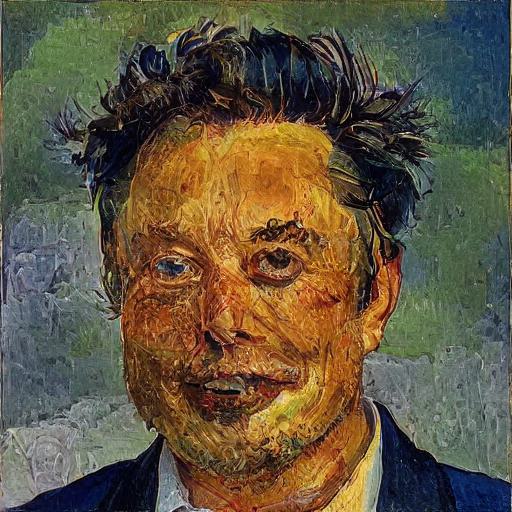}
\end{minipage}
\begin{minipage}[b]{0.112\textwidth}
    \makebox[1\textwidth]{\raisebox{0.5em}{\vphantom{X} Ours}} \\
    \includegraphics[width=1\textwidth]{Figures/MagicStyle/styles_alpha0.8/andry-dedouze-train-final/chenglong.jpeg}\\
    \includegraphics[width=1\textwidth]{Figures/MagicStyle/styles_alpha0.8/mannai_sunset/Feifei.jpg}\\
    \includegraphics[width=1\textwidth]{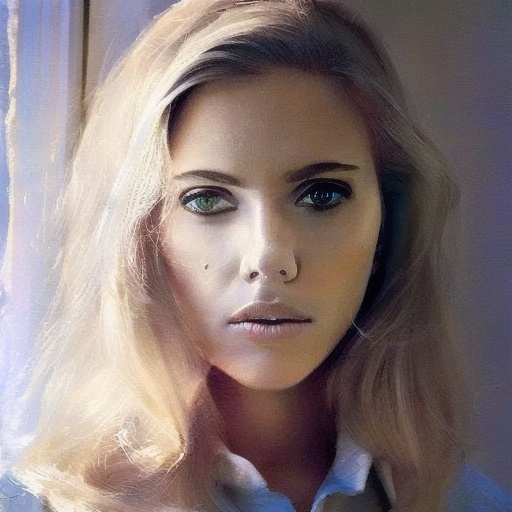}\\
    \includegraphics[width=1\textwidth]{Figures/MagicStyle/styles_alpha0.8/ink_boy/kuke.jpeg}\\
    \includegraphics[width=1\textwidth]{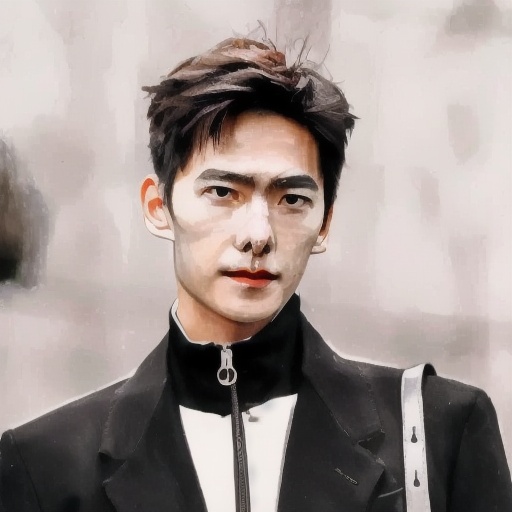}\\
    \includegraphics[width=1\textwidth]{Figures/MagicStyle/styles_alpha0.8/vangogh_flower/Musk2.jpg}
\end{minipage}

\caption{Visualization comparison with baseline models, MagicStyle ($\alpha=0.8, \beta=0.2$) can not only provide excellent image stylization results, but also preserve details such as facial identity, expressions, and background from the content image.}
\label{fig:baseline}
\end{figure*}

We also performed visual comparisons with several baseline models, including InstantID \cite{wang2024instantid}, PhotoMakerV2 \cite{li2024photomaker}, StyleAligned \cite{hertz2024style}, CSGO \cite{xing2024csgo} and StyleID \cite{chung2024style}. It is to be noted that InstantID, PhotoMakerV2, and StyleAligned cannot provide images as styles; instead, we obtained image description texts as prompts for the image style using BLIP \cite{li2022blip}. As illustrated in Fig. \ref{fig:baseline}, both InstantID and PhotoMakerV2, as personalized image generation methods, are able to retain the facial identity from the content image in the generated images, although the background of the face changes. Furthermore, since they cannot provide images as styles, the generated images do not consistently match the input style images. The stylization results from StyleAligned, StyleID, and CSGO appear more natural, with the generated portraits blending well with the background. However, since StyleAligned and CSGO are not specifically designed for portrait generation, the facial identity in the content images cannot be preserved. The results most similar to ours are from StyleID, which, like MagicStyle, maintains both the content of the image and its stylization. However, a comparison reveals that StyleID's retention of facial details and background is not as effective as our method.

\subsection{Quantitative Comparison Results} 

To quantitatively validate our method, we use the DINO \cite{caron2021emerging} and CLIP-I \cite{radford2021learning} metrics to measure image fidelity, and the FID metric to evaluate the quality of the generated images. Finally, a facial recognition model \cite{deng2020sub} is employed to measure facial similarity (FaceSim).
As shown in Table \ref{tab:Quantitative}, compared to other methods, our MagicStyle achieves SOTA results across all metrics.

\begin{table}[!t]
\centering
  \setlength{\tabcolsep}{0pt} 
  \begin{tabular*}{\columnwidth}{@{\extracolsep{\fill}}lcccccc@{}}
    \toprule
    \textbf{Method} & ClipI $\uparrow$ &  FaceSim $\uparrow$ & DINO $\uparrow$  & FID $\downarrow$\\
    \midrule
    InstantID  & 85.7 & 89.5 & 90.8 & 240.3 \\
    PhotoMakerV2  & 84.9 & 86.6 & 91.2  &  295.4 \\
    StyleAligned  & 86.8 & 73.6  & 89.7  & 116.9  \\
    CSGO & 84.2 & 67.3 & 84.9  & 188.9 \\
    StyleID & 84.5 & 84.9  & 89.5 & 138.2 \\
    \textbf{MagicStyle(Ours)} &\textbf{89.2}  & \textbf{95.8}  & \textbf{96.9} & \textbf{73.1}\\
    \bottomrule
  \end{tabular*}
 \caption{Quantitative comparison results with baseline models. MagicStyle achieves SOTA results across all metrics.}
\label{tab:Quantitative}
\end{table}

\subsection{Ablation Results}

\begin{figure*}[!h]
\centering
\begin{minipage}[b]{0.112\textwidth}
    \makebox[1\textwidth]{\raisebox{0.5em}{\vphantom{X} Content}}\\
    \includegraphics[width=1\textwidth]{Figures/input_img/content_crop/chenglong.jpeg}\\
    \includegraphics[width=1\textwidth]{Figures/input_img/content_crop/Feifei.jpg}\\
    \includegraphics[width=1\textwidth]{Figures/input_img/content_crop/Musk2.jpg}
\end{minipage}
\begin{minipage}[b]{0.112\textwidth}
    \makebox[1\textwidth]{\raisebox{0.5em}{\vphantom{X} Style}}\\
    \includegraphics[width=1\textwidth]{Figures/input_img/style/andry-dedouze-train-final.jpg}\\
    \includegraphics[width=1\textwidth]{Figures/input_img/style/mannai_sunset.jpg}\\
    \includegraphics[width=1\textwidth]{Figures/input_img/style/vangogh_flower.jpg}
\end{minipage}
\begin{minipage}[b]{0.112\textwidth}
    \makebox[1\textwidth]{\raisebox{0.5em}{\vphantom{X} $\beta=0$}}\\
    \includegraphics[width=1\textwidth]{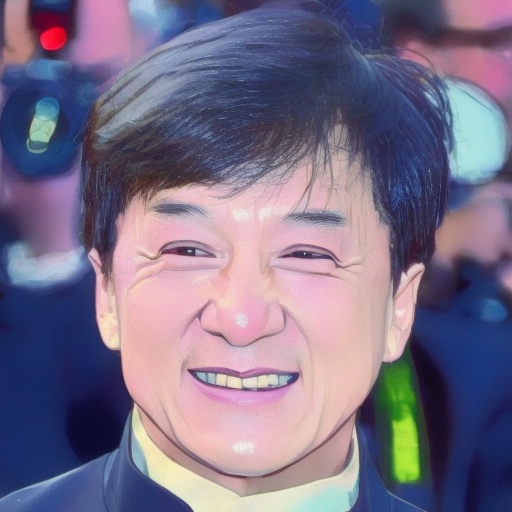}\\
    \includegraphics[width=1\textwidth]{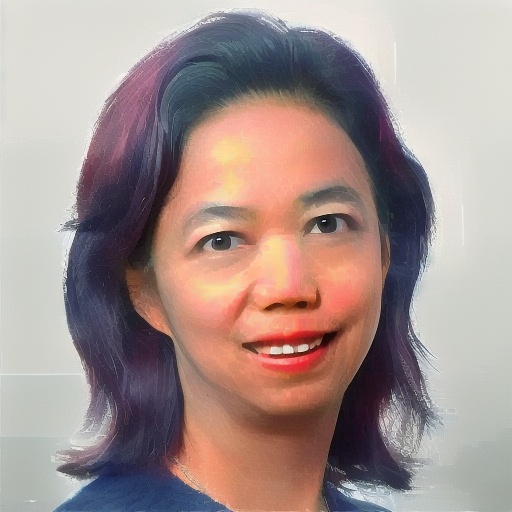}\\
    \includegraphics[width=1\textwidth]{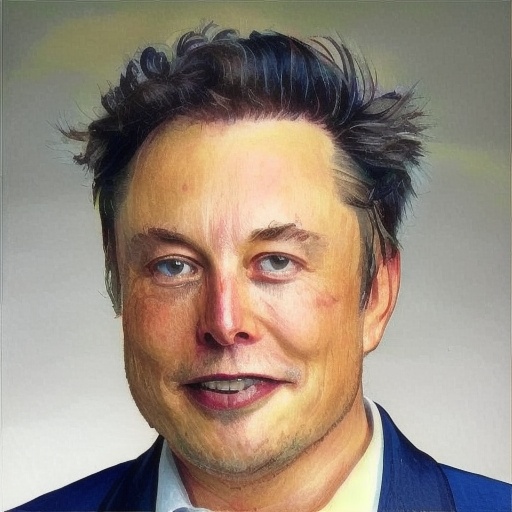}
\end{minipage}
\begin{minipage}[b]{0.112\textwidth}
    \makebox[1\textwidth]{\raisebox{0.5em}{\vphantom{X} $\beta=0.2$}}\\
    \includegraphics[width=1\textwidth]{Figures/MagicStyle/styles_alpha0.8/andry-dedouze-train-final/chenglong.jpeg}\\
    \includegraphics[width=1\textwidth]{Figures/MagicStyle/styles_alpha0.8/mannai_sunset/Feifei.jpg}\\
    \includegraphics[width=1\textwidth]{Figures/MagicStyle/styles_alpha0.8/vangogh_flower/Musk2.jpg}
\end{minipage}
\begin{minipage}[b]{0.112\textwidth}
    \makebox[1\textwidth]{\raisebox{0.5em}{\vphantom{X} $\beta=0.5$}}\\
    \includegraphics[width=1\textwidth]{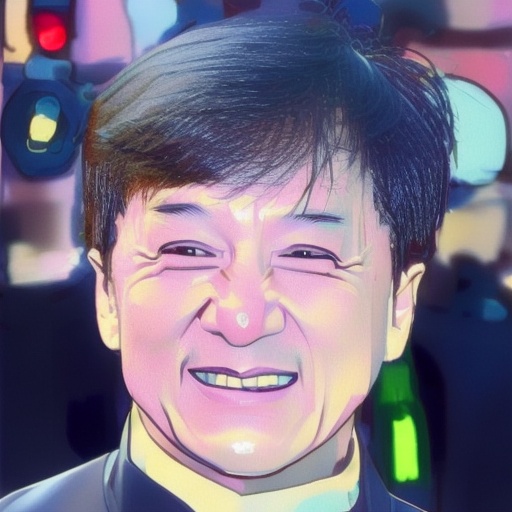}\\
    \includegraphics[width=1\textwidth]{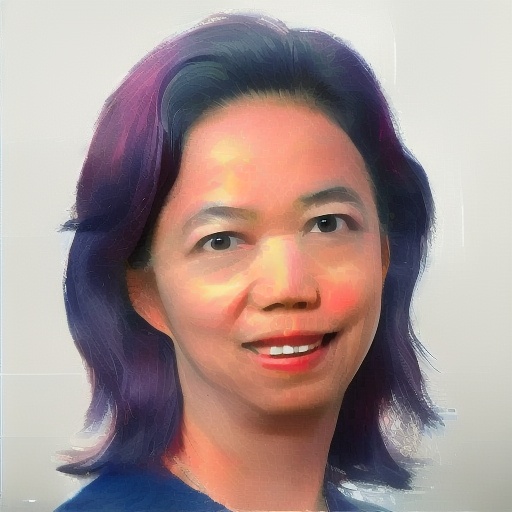}\\
    \includegraphics[width=1\textwidth]{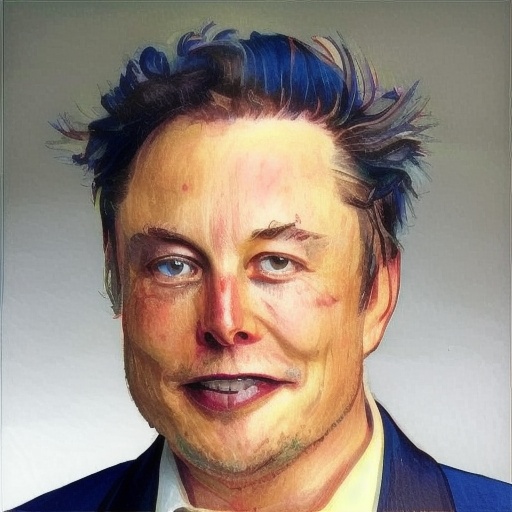}
\end{minipage}
\begin{minipage}[b]{0.112\textwidth}
    \makebox[1\textwidth]{\raisebox{0.5em}{\vphantom{X} $\beta=0.8$}}\\
    \includegraphics[width=1\textwidth]{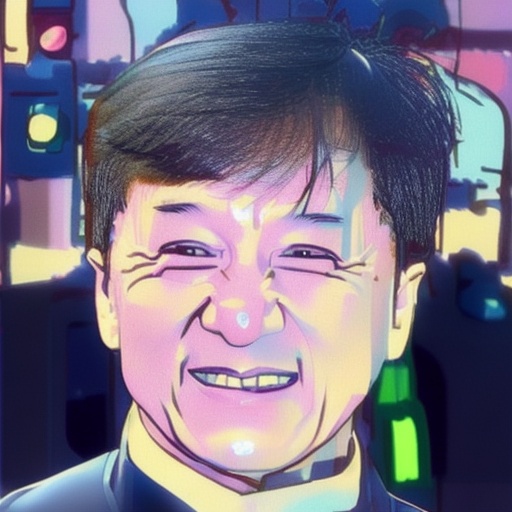}\\
    \includegraphics[width=1\textwidth]{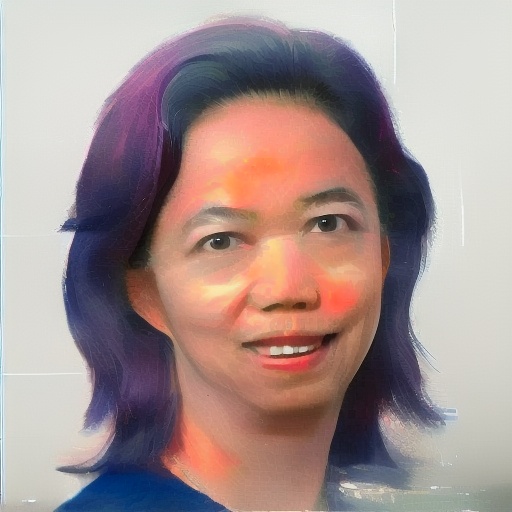}\\
    \includegraphics[width=1\textwidth]{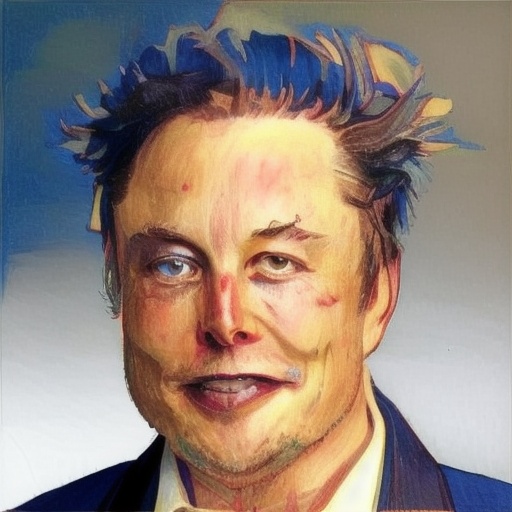}
\end{minipage}
\begin{minipage}[b]{0.112\textwidth}
    \makebox[1\textwidth]{\raisebox{0.5em}{\vphantom{X} $\beta=1$}}\\
    \includegraphics[width=1\textwidth]{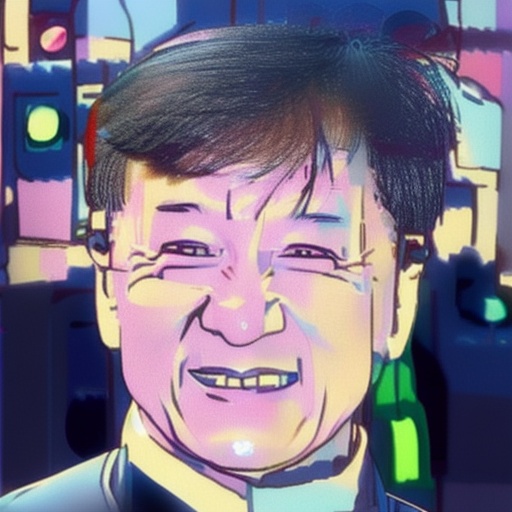}\\
    \includegraphics[width=1\textwidth]{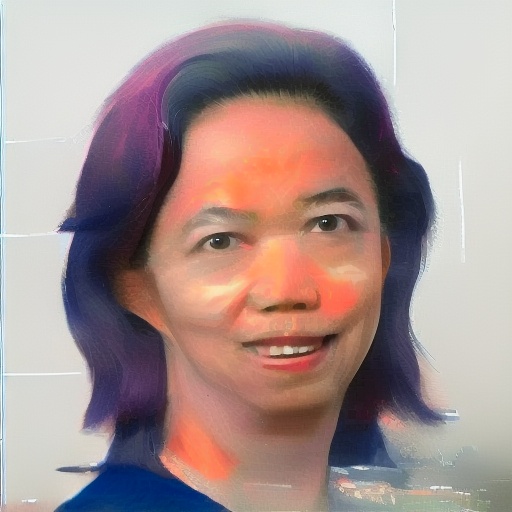}\\
    \includegraphics[width=1\textwidth]{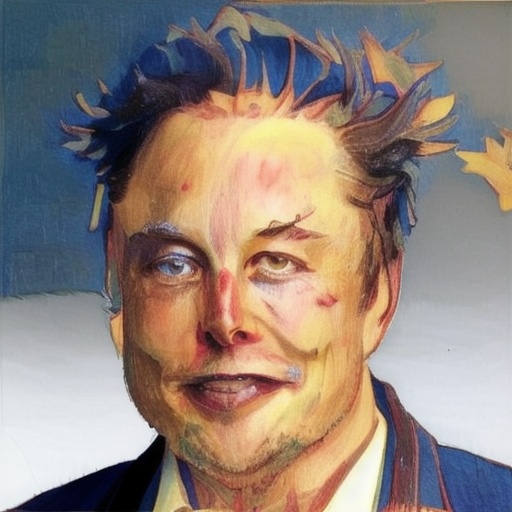}
\end{minipage}
\begin{minipage}[b]{0.112\textwidth}
    \makebox[1\textwidth]{\raisebox{0.5em}{\vphantom{X} wo. FFA}}\\
    \includegraphics[width=1\textwidth]{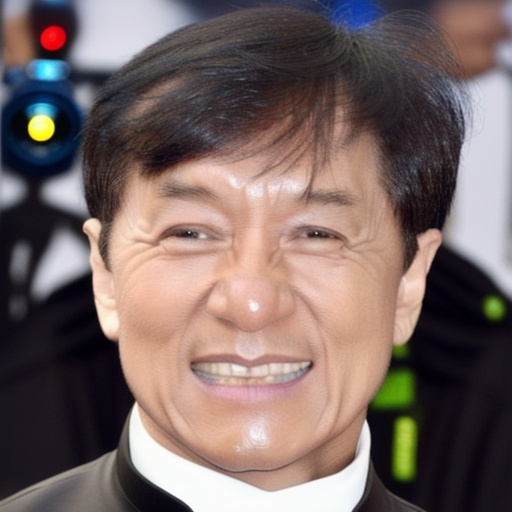}\\
    \includegraphics[width=1\textwidth]{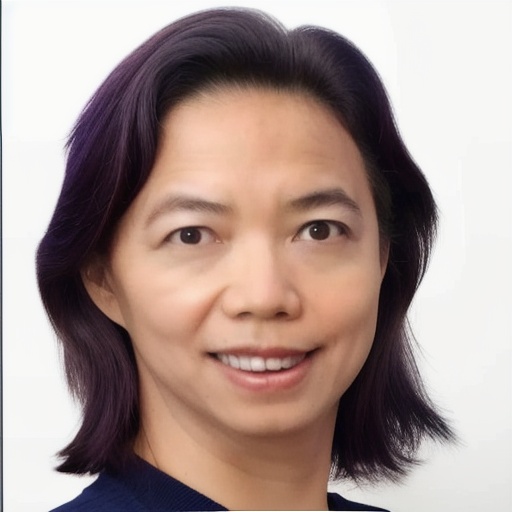}\\
    \includegraphics[width=1\textwidth]{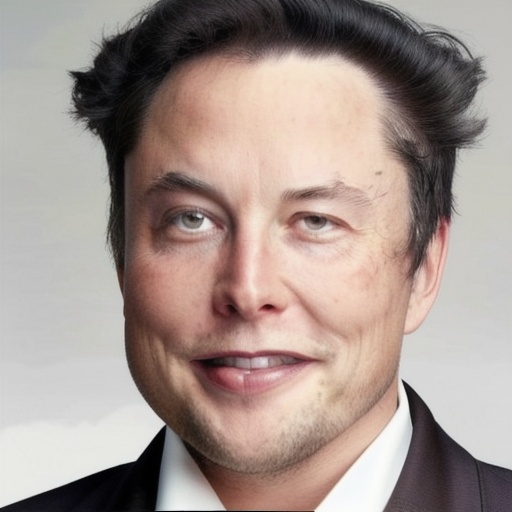}
\end{minipage}

\caption{Ablation results of Feature Fusion Attention(FFA). We adjust the weights of the content and style queries ($\alpha$ and $\beta$), while maintaining $\alpha+\beta=1$. By increasing $\beta$, we explore the impact of these two weights on image stylization in MagicStyle. Additionally, we replace FFA with standard Attention to validate the stylization effectiveness of FFA.}
\label{fig:ffa}
\end{figure*}

To further validate the effectiveness of Feature Fusion Attention (FFA), we conducted ablation experiments. As shown in Fig. \ref{fig:ffa}, we adjusted the coefficient $\beta$ from 0 to 1 and replaced FFA with a standard cross-attention module to observe the different image generation results. As $\beta$ increases, the degree of image stylization improves, but this comes at the cost of losing some identity details of the content image.
When FFA was replaced, the generated images failed to capture similar styles from the style image and instead remained consistent with the input content image. This result further validates the importance of FFA in the stylization process, indicating that FFA can effectively fuse content and style features, thereby enhancing the quality of the generated images.

\section{CONCLUSIONS}
The proposed MagicStyle effectively addresses the challenge of maintaining content image details while incorporating style features. Through the Content and Style DDIM Inversion (CSDI) and Feature Fusion Forward (FFF) phases, we demonstrated that our method can harmoniously blend texture and color information. The visualization experiments confirmed that MagicStyle successfully preserves the intricacies of the content image across various portraits. Comparative analyses with baseline models revealed that MagicStyle strikes a superior balance between content preservation and stylization, producing high-quality results.
Furthermore, our ablation studies highlighted the critical role of Feature Fusion Attention (FFA) in enhancing stylization quality. The experiments show that replacing FFA with a standard cross-attention module led to a significant loss of style features, underscoring its importance in our framework. Overall, MagicStyle represents a significant advancement in portrait stylization, offering a robust solution for integrating artistic styles while retaining essential content details.


\end{document}